%% file: main.tex
\pdfoutput=1
\documentclass[11pt]{article}
\usepackage[textwidth=400pt,centering]{geometry}
\usepackage{fullpage}
\usepackage[utf8]{inputenc} 
\usepackage[T1]{fontenc}    
\usepackage{url}            
\usepackage{nicefrac}       
\usepackage{xcolor}    
\usepackage{times}

\usepackage{csquotes}
\usepackage{makecell} 
\usepackage{microtype}
\usepackage{graphicx}
\usepackage{subcaption}
\usepackage{booktabs} 
\usepackage[hidelinks]{hyperref}

\input{math_commands.tex}

\usepackage{amsthm}
\usepackage[capitalize,noabbrev]{cleveref}
\theoremstyle{plain}
\newtheorem{theorem}{Theorem}

\newtheorem{lemma}[theorem]{Lemma}

\theoremstyle{definition}
\newtheorem{definition}[theorem]{Definition}

\theoremstyle{remark}
\newtheorem{remark}[theorem]{Remark}

\usepackage{titletoc}
\usepackage{calc}  
\usepackage{array} 

\usepackage[round]{natbib}
\usepackage[nottoc,notlot,notlof]{tocbibind}
\usepackage[vlined, linesnumbered, ruled]{algorithm2e}

\usepackage{pifont}
\usepackage{bbm}
\usepackage{multirow}
\usepackage{braket}
\usepackage{enumitem}
\definecolor{darkblue}{rgb}{0,0.08,0.8}
\newcommand\numberthis{\addtocounter{equation}{1}\tag{\theequation}}

\usepackage{verbatim}

\usepackage{tikz}
\usetikzlibrary{arrows.meta,positioning}

\tikzset{
  vone/.style={
    circle,
    draw,
    fill=white,
    minimum size=5.5mm,
    inner sep=0pt,
    font=\scriptsize
  },
  vtwo/.style={
    circle,
    draw,
    fill=gray!30,
    minimum size=5.5mm,
    inner sep=0pt,
    font=\scriptsize
  },
  obs/.style={->,>=Stealth,thin},
  bi/.style={<->,>=Stealth,thin}
}

\newcommand{\ul}{\underline{\lambda}}
\newcommand{\uL}{\underline{L}}

\newcommand{\inb}{V^{\mathrm{in}}}
\newcommand{\onb}{V^{\mathrm{out}}}

\title{Nearly Linear-Time Follow-the-Perturbed-Leader \\ for Weakly Observable Graph Bandits}
\author{Jongyeong Lee$^{1}$ \\ \fontsize{10}{12}\selectfont jongyeong@kist.re.kr \and Hyundam Yoo$^{2}$ \\ \fontsize{10}{12}\selectfont closer716@gmail.com}
\date{\fontsize{10}{12}\selectfont
$^1$ Korea Institute of Science and Technology 
$^2$ Independent researcher
}

\begin{document}
\maketitle

\begin{abstract}
  We study online learning with directed weakly observable feedback graphs and propose a Follow-the-Perturbed-Leader (FTPL) policy that achieves best-of-both-worlds guarantee with nearly linear expected time complexity.
  To preserve the simple structure of FTPL, we develop a novel graph-based loss estimator that combines forced exploration, conditional geometric resampling and adaptive learning rates with surrogate probabilities.
  Our policy achieves $\tilde{\gO}(\delta_*^{1/3} T^{2/3})$ regret for adversarial regimes as well as $\tilde{\gO}(\delta_* \log T / \Delta_{\min}^2)$ regret for stochastic regimes, where $\delta_*$ and~$\Delta$, respectively, denote the fractional weak domination number and the suboptimality gap.
  These results show that refined guarantees for problems with minimax regret of $\Theta(T^{2/3})$ can be obtained through appropriately designed adaptive learning rates and loss estimators without sacrificing the simplicity of FTPL.
\end{abstract}

\section{Introduction}
In online learning, a learner interacts with an environment over a sequence of rounds by making decisions and receiving feedback, where their performance is often evaluated by its cumulative loss~\citep{cesa2006prediction}.
This general framework encompasses many important sequential decision-making problems, such as partial monitoring~\citep{cesa2006regret, bartok2014partial} and online classification~\citep{littlestone1988learning, ben2009agnostic}.
In this paper, we consider a setting where the feedback structure is represented by a directed graph $G=(V,E)$, where the edge $(i,j)\in E$ indicates that selecting arm $i$ reveals feedback about arm $j$~\citep{mannor2011bandits, alon2015online}.
This problem generalizes both the multi-armed bandits (MAB), whose graph contains only self-loops~\citep{lai1985asymptotically, auer1995gambling}, and prediction with expert advice, whose graph is fully connected~\citep{littlestone1994weighted, freund1997decision}.

Since sublinear regret is impossible for unobservable graphs \citep{alon2015online}, it is natural to consider observable graphs, where the feedback of every arm can be obtained by playing at least one arm.
A feedback graph is called strongly observable if the feedback of every arm is revealed either by a self-loop or by edges from \emph{any} other arms \citep{erez2021towards,rouyer2022near}.
An observable graph that is not strongly observable is called weakly observable, where at least one arm without a self-loop can be observed only by playing a particular \emph{subset} of the other arms \citep{chen2021understanding}.
Learning with weakly observable graphs therefore requires additional explorations to observe such arms.
Consequently, the minimax regret is $\Theta(T^{2/3})$ for weakly observable graphs, whereas that for strongly observable graphs is $\Theta(\sqrt{T})$, up to graph-dependent factors~\citep{alon2015online,chen2021understanding}.


\begin{table*}[t]
\caption{Regret comparison for horizon $T$, graph order $K$, weak domination number $\delta$, fractional weak domination numbers $\delta_*$, fractional domination number $\delta'_*$, and minimum suboptimality gap $\Delta_{\min}$. Note that $\delta_*\leq \min(\delta, \delta_*')$ holds~\citep{chen2021understanding}. AwSC is the abbreviation of the adversarial regime with self-bounding constraints, which encompasses general intermediate regimes between stochastic and fully adversarial environments. MS-type denotes the bound in AwSC that has a form similar to the refined bound in \citet{masoudian2021improved}.} \label{tab: regret}
\begin{center}
\begin{tabular}{llll}
\textbf{Reference} & \textbf{Stochastic}  &\textbf{Adversarial} & \textbf{AwSC} \\
\midrule
\makecell[l]{\citet{kong2022simultaneously}\\ (Switching w/ Exp3-type)} & $\displaystyle \frac{\delta^2 \log^{3/2} T }{\Delta_{\min}^3}$ & $\displaystyle \delta^{1/3}\sqrt{\log T} (KT)^{2/3}$ & N/A\\
\midrule
\makecell[l]{\citet{ito2022nearly} \\ (FTRL)} & $\displaystyle \frac{\delta \log T \log (KT)}{\Delta_{\min}^2}$ & $\displaystyle(\delta \log T \log (KT))^{1/3} T^{2/3}$  & \checkmark \\
\midrule
\makecell[l]{\citet{dann2023blackbox} \\ (Reduction w/ Exp3-type) } & $\displaystyle \frac{\delta \log K \log T}{\Delta_{\min}^2}$ & $\displaystyle(\delta \log K)^{1/3} T^{2/3}$ & \checkmark \\
\midrule
\makecell[l]{\citet{tsuchiya2024simple}\\ (FTRL)} & $\displaystyle \frac{\delta'_* \log K \log T}{\Delta_{\min}^2}$ & $\displaystyle(\delta'_* \log K)^{1/3} T^{2/3}$ & \checkmark (MS-type) \\ 
\midrule
\makecell[l]{Algorithm~\ref{alg: ftpl} (Ours)\\ (FTPL)} & $\displaystyle \frac{\delta_* \log^2 K \log T}{\Delta_{\min}^2}$ & $\displaystyle (\delta_* \log^3 K)^{1/3} T^{2/3}  $ & \checkmark (MS-type) \\
\midrule
\end{tabular}
\end{center}
\end{table*}
These minimax regret orders characterize the worst-case cost but can be too pessimistic when the losses are generated by a favorable stochastic environment.
Ideally, a single policy should exploit such structures to achieve logarithmic regret without sacrificing its robustness to adversarial losses.
This motivates the study of best-of-both-worlds (BOBW) policies, which achieve (near-)optimal adversarial regret and polylogarithmic stochastic regret without knowing the environment in advance~\citep{bubeck2012best}.

For weakly observable graph bandits, \citet{kong2022simultaneously} obtained the first BOBW result using an elimination-based policy together with a test that switches to Exp3-type policy when adversarial behavior is detected.
\citet{ito2022nearly} instead developed a direct Follow-the-Regularized-Leader (FTRL) policy with a hybrid regularizer and adaptive learning rates, improving the adversarial regret to the near-optimal order.
\citet{dann2023blackbox} proposed a hierarchical reduction policy that combines an Exp3-type policy with a doubling-trick reduction that discards past observations, attaining optimal dependence on the horizon $T$ in both regimes.
More recently, \citet{tsuchiya2024simple} introduced refined adaptive learning rates, obtaining improved BOBW guarantees with a direct FTRL policy.
Nevertheless, these methods rely on regime detection, reduction-based procedures, or per-round convex optimization, motivating simple and computationally efficient BOBW policies for weakly observable graphs.

\paragraph{Contributions.}
We propose a simple and computationally efficient BOBW policy for weakly observable graph bandits based on Follow-the-Perturbed-Leader (FTPL) framework.
Our policy combines the standard FTPL rule using i.i.d.~Pareto perturbations and forced exploration using the fractional weakly domination number $\delta_*$.
To address weakly observable graphs, we employ adaptive exploration and learning rates and introduce a novel graph-based loss estimator.
We prove that our policy achieves $\gO( \delta_*^{1/3} T^{2/3}\log K)$ adversarial regret and gap-dependent logarithmic stochastic regret, as well as a refined corruption-dependent bound under self-bounding constraints~\citep{masoudian2021improved}, referred to as MS-type bounds by \citet{tsuchiya2024simple}.
Moreover, its expected per-round computational cost is $\gO(K\log K)$, which is nearly linear in the graph order $|V|=K$ and independent of the graph size $|E|$. 
These results show that appropriately designed adaptive learning rates and loss estimators enable simple FTPL policy to obtain refined BOBW guarantees even for problems with minimax regret of order $\Theta(T^{2/3})$.
Table~\ref{tab: regret} summarizes the regret guarantees of our policy and existing policies.

\section{Preliminaries}\label{sec: prelim}\vspace{-0.3em}
Let $G=(V,E)$ be a given directed graph, with vertex set $V=[K]=\{1,\ldots, K\}$ and edge set $E\subseteq V \times V$.
For each arm $i\in V$, let 
\begin{align*}
  \inb(i)&=\{j \in V: (j,i)\in E\}, \\ \onb(i)&=\{j \in V: (i,j)\in E\}
\end{align*}
denote its in- and out-neighborhoods, respectively.
At each round $t \in [T]$, the environment chooses a loss function $\ell_t: V \to [0,1]$.
Without knowing $\ell_t$, the learner plays an arm $i_t\in V$, incurs loss $\ell_{t,i_t}$, and observes $\ell_{t,j}$ for all $j\in\onb(i_t)$.
Note that the learner does not observe its incurred loss if $i_t$ has no self-loop.
Their performance is evaluated by the (pseudo-)regret defined as
\begin{align*}
  \Reg_T(i) = \E\qty[\sum_{t=1}^T \ell_{t,i_t} - \sum_{t=1}^T \ell_{t,i}], \, R_T := \max_{i\in V} R_T(i).
\end{align*}
To address stochastic, adversarial, and intermediate regimes within a unified framework, we consider the adversarial regime with self-bounding constraints (AwSC) $(\Delta,C,T)$~\citep{zimmert2021tsallis}, where any policy satisfies
\begin{equation}\label{def: SCA}
  \Reg_T \geq \E\qty[\sum_{t=1}^T \Delta_{i_t}- C].
\end{equation}
In this paper, we assume that there exists a unique optimal arm $i^* \in V$ satisfying $\Delta_{i^*}=0$, which is standard in BOBW analyses~\citep{ito2022nearly,dann2023blackbox}, and define $\Delta_{\min}:= \min_{i\in V \setminus \{i^*\}} \Delta_i$.
Following \citet{tsuchiya2024simple}, we call a regret bound under AwSC an MS-type bound if it has a form analogous to the refined bound in \citet{masoudian2021improved}, which maintains meaningful guarantee even in the extreme cases such as $C=\Theta(T)$ or $\Delta_{\min}=\Theta(1/\sqrt{T})$.

\subsection{Weakly observable graph}
We define weak domination with respect to all vertices without self-loops following \citet{chen2021understanding} and \citet{ito2022nearly}. 
This definition differs slightly from that of \citet{alon2015online}, as discussed in Appendix~A of \citet{chen2021understanding} and Appendix~D of \citet{ito2022nearly}.
\begin{definition}\label{def: weak domination}
  For a graph $G=(V,E)$, let $V_1\subseteq V$ be the set of vertices without self-loops.
  The fractional weak domination number of $\delta_*(G)$ is the optimal value of the following linear program 
  \begin{align*}
    &\mathrm{minimize} \quad \sum_{i \in V} x_i, \numberthis{\label{def: linear program}} \\
    &\mathrm{subject~to} \sum_{i \in \inb(j)} x_i \geq 1, \forall j \in V_1, \hspace{0.6em} 0 \leq x_i \leq 1, \forall i \in V.
  \end{align*}
  Replacing constraints $x_i\in [0,1]$ with $x_i\in \{0,1\}$ gives the corresponding integer program, whose optimal value is called the weak domination number $\delta(G)$.
\end{definition}
The fractional domination number $\delta'_*$ in \citet{tsuchiya2024simple} imposes the coverage constraints on $j\in V$, rather than $V_1$, implying $\delta_* \leq \delta'_*$.
Since (\ref{def: linear program}) is a fractional relaxation of the set cover problem defining $\delta$, we have $\delta_* \leq\delta$. 
Moreover, $\delta_*$ can be computed in polynomial time~\citep{cohen2021solving}, while computing $\delta$ is NP-hard~\citep{karp1972reducibility}.
For given $G$, we set $V_1$ to be the set of vertices without self-loops, and $V_2 = V\setminus V_1$ be the set of vertices with self-loops.

For a weakly observable graph with $|V|\geq 2$, \citet{alon2015online} characterized the minimax regret lower bound as $\tilde{\Theta}(\delta^{1/3}T^{2/3})$ for $T \geq |V|^3$, while \citet{chen2021understanding} established lower bounds in terms of $\delta_*$ and the integrality gap of an associated vertex-packing program. 

\subsection{Frameworks of BOBW policies}
One approach to BOBW guarantee is to combine policies designed for different regimes. 
Early methods employ statistical tests to detect non-stochastic behavior and switch to a policy for adversarial regimes~\citep{bubeck2012best,auer2016algorithm, lee2021achieving}.
In weakly observable graph bandits, \citet{kong2022simultaneously} combine an elimination-based stochastic policy with a test that transits to an Exp3 policy.
Such methods are tailored to the stochastic and adversarial regimes and do not directly provide refined guarantees under AwSC.
In contrast, \citet{dann2023blackbox} obtain refined guarantees by applying a general hierarchical reduction with doubling-trick procedures to an adaptive Exp3 policy.
Despite their different mechanisms, both methods obtain BOBW results through modular constructions.

A complementary line of research develops BOBW policies that incorporate adaptivity into a single framework without explicitly identifying the underlying regimes~\citep{seldin2014one}.
Most such policies are based on FTRL, or the online-mirror descent framework, where carefully designed regularizers and adaptive learning rates provide refined guarantees in various problems~\citep{rouyer2020tsallis, zimmert2021tsallis, huang2022adaptive, masoudian2022best, tsuchiya2023best, kuroki2024best}.
At each round, however, FTRL computes its selection distribution by solving a regularized optimization problem, which may not admit a simple closed-form solution when hybrid regularizers are used to obtain refined guarantees, as in those for weakly observable graph bandits~\citep{ito2022nearly,tsuchiya2024simple}.

In contrast, FTPL selects an arm by directly minimizing randomly perturbed cumulative loss estimates, which requires only linear time in $K$ and therefore becomes a computationally efficient alternative.
Given learning rates $\eta_t >0$ and the cumulative loss estimators $\hat{L}_t = \sum_{s=1}^{t-1}\hat{\ell}_s$ available before round $t$, FTPL draws perturbations from a predefined distribution $\mathcal{D}$, $r_t \sim\mathcal{D}$, and selects
\begin{equation*}
  i_t \in \argmin_{i\in V} \{\hat{L}_{t,i}-r_{t,i}/\eta_t \}.
\end{equation*}
Although $i_t$ can be sampled without computing the selection distribution, the corresponding arm-selection probability can be written by
\begin{align*}
  w_{t,i}  = \mathrm{Pr}(i_t=i|\hat{L}_t) = \int_{-\infty}^\infty \prod_{j\ne i} F(z+\eta_t \hat{\uL}_{t,j}) \dd F(z+\eta_t \hat{\uL}_{t,i}), \numberthis{\label{eq: wt}}
\end{align*}
where $\hat{\uL}_{t,i}=\hat{L}_{t,i} - \min_{j \in V}\hat{L}_{t,j}$ and $F$ is distribution function of $\gD$.
Thus, although $w_t$ is computable in principle, it generally lacks a simple closed-form expression and may require numerical integration, which makes BOBW analyses less straightforward.
Reflecting this difficulty, to the best of our knowledge, existing FTPL-based BOBW results are confined to problems with $\Theta(\sqrt{T})$ minimax regret, including multi-armed bandits~\citep{pmlr-v201-honda23a}, bandits with expert advices~\citep{pmlr-v336-lee26a}, combinatorial semi-bandits~\citep{zhan2026follow,chen2026further}, and decoupled bandits~\citep{kim2025follow}.






\subsection{Adaptive learning rates}
Both FTRL and FTPL are general leader-based frameworks with simple decision rules, whose problem-specific behavior is largely determined by three design choices: (i) the regularizer in FTRL or the perturbation distribution in FTPL, (ii) the learning rate $\eta_t$ and (iii) loss estimator $\hat{\ell}_t$ used to handle partial feedback. 
Many recent works on FTRL have focused on the first two components, introducing hybrid regularizers and $w_t$-dependent adaptive learning rates~\citep{zimmert2019beating, jin2023improved,tsuchiya2023stability, ito2024adaptive,pmlr-v291-nguyen25a}.
In particular, explicit computation of $w_t$ in FTRL enables both the use of standard importance-weighted (IW) estimators and refined learning rate updates in terms of $w_t$.
Therefore, the main algorithmic choices focus on the regularizer, which determines the property of $w_t$, and adaptive learning rates that exploit its data-dependent behavior.

In other words, the application of such approaches to FTPL is not straightforward since $w_t$ of FTPL is induced only implicitly by the perturbations.
For (iii), existing FTPL policies usually adopt resampling methods to estimate the inverse selection probability $w_{t,i_t}^{-1}$ of the selected arm and construct IW loss estimates~\citep{neu2016importance,chen2025geometric}.
Moreover, for (ii), recent studies replace the implicit $w_t$ with computable surrogate quantities~\citep{kim2025follow}.
In particular, \citet{pmlr-v336-lee26a} proposed probability-dependent adaptive learning rates using surrogate probability quantities, obtaining generalized BOBW guarantees for multi-armed bandits and bandits with expert advices. 

Motivated by these developments, we introduce two surrogate quantities.
Fix an arbitrary strict total order $\prec$ on $V$ for tie-breaking, and define the rank of loss estimates $\hat{L}_{t,i}$ in non-decreasing order by 
\begin{equation*}
  \sigma_{t,i} = 1 + |\{j \in V: \hat{L}_{t,j} < \hat{L}_{t,i}\}| +  |\{j \prec i: \hat{L}_{t,j} = \hat{L}_{t,i}\}|.
\end{equation*}
Let $j_t$ be the arm satisfying $\sigma_{t,j_t}=1$, which is unique by the order $\prec$.
Then, define surrogate quantities by
\begin{subequations}\label{def: p and q}
  \begin{align}
    p_{t,i} &= \min\qty(\frac{1}{(1+\eta_t \hat{\uL}_{t,i})^\alpha}, \frac{1}{\sigma_{t,i}}) ,\label{def: p}\\
    q_{t,i} &= \min\qty(\frac{1}{(1+\eta_{t-1} \hat{\uL}_{t,i})^\alpha}, \frac{1}{\sigma_{t,i}}), \label{def: q} 
  \end{align}
\end{subequations}
for $\alpha>0$ specified in Section~\ref{sec: policy}.
The quantity $p_{t,i}$ is computable once $\eta_t$ is fixed and provides an upper bound of $w_{t,i}$. 
In contrast, $q_{t,i}$ depends only on the previous learning rate $\eta_{t-1}$ and is therefore available before $\eta_t$ is chosen, serving as a computable proxy for $p_{t,i}$~\citep{pmlr-v336-lee26a}.
Note that $\hat{\uL}_{t,i}= \hat{L}_{t,i} - \hat{L}_{t,j_t}$ by definition of $j_t$.

\section{Proposed Policy}\label{sec: policy}
The proposed policy augments standard FTPL with forced exploration based on the normalized optimal fractional weakly dominating vector.
Specifically, let $x^*$ be an optimal solution to linear program in (\ref{def: linear program}).
Then, at each round $t$, given an exploration probability $\gamma_t \in (0,1/2]$ specified in Section~\ref{sec: params}, we draw $B_t\sim \mathrm{Ber}(\gamma_t)$ and perform forced explorations if $B_t=1$, i.e., $i_t \sim w^0$ for
\begin{equation}\label{def: w0}
  w_i^0 =  x_i^* / \delta_*(G).
\end{equation}
as employed in previous FTRL-based approaches~\citep{chen2021understanding, tsuchiya2024simple}.
If $B_t=0$, we draw i.i.d.~Pareto perturbations with shape $\alpha>0$ and play $i_t$ by the FTPL rule, i.e.,
\begin{equation*}
  i_t \in \argmin_{i \in [K]} \{\hat{L}_{t,i} - r_{t,i}/\eta_t\}, \, r_{t,i} \sim \mathrm{Par}(\alpha).
\end{equation*}
Let $w_t$ be the arm-selection distribution induced by the FTPL rule conditional on $\hat{L}_t$, as written in (\ref{eq: wt}).
Then, the overall arm-selection distribution of $i_t$ can be written as $i_t \sim y_t$ for
\begin{equation}\label{def: total prob}
  y_{t} = (1-\gamma_t)w_{t} + \gamma_t w^0.
\end{equation} 
It remains to specify the exploration probabilities $\gamma_t$, learning rates $\eta_t$, and loss estimators $\hat{\ell}_t$, which will be presented in the remainder of this section.

\subsection{Choice of parameters}\label{sec: params}
Forced exploration is important especially for controlling the observation probabilities of arms without self-loops, all of which belong to $V_1$. 
We define the exploration probability $\gamma_t$ jointly by the learning rate $\eta_t$ and the surrogate quantities introduced in (\ref{def: p and q}). 
Let $\beta_t =\eta_{t}^{-1}$.
At the beginning of the round $t$, we set forced exploration probability to 
\begin{equation}\label{def: gammat}
  \gamma_t := \sqrt{\frac{z_t}{\beta_t}} + \frac{u_t+v_t}{\beta_t},
\end{equation}
where 
\begin{subequations}\label{def: params}
  \begin{align}
    z_t &:= \alpha \delta_*\bigg(\sum_{i \in V_1 \setminus \{j_t\}}2e\sqrt{2} q_{t,i}^{1+\frac{1}{\alpha}} + \I[j_t \in V_1] \sum_{i \ne j_t} 3\sqrt{2}q_{t,i}^{1+\frac{1}{\alpha}}\bigg), \label{def: zt}\\
    u_t &:= \alpha\bigg(\sum_{i \in V_2 \setminus \{j_t\}}4e\sqrt{2} q_{t,i}^{\frac{1}{\alpha}} + \I[j_t \in V_2] \sum_{i \ne j_t} 24\sqrt{2}q_{t,i}^{\frac{1}{\alpha}}\bigg),  \label{def: ut}\\
    v_t &:= c_v (\alpha+1)\delta_* \bar{s}_t^{1/\alpha} \I[j_t \in V_1],   \label{def: vt} \\
    \bar{s}_t &:= \min\qty(1, \sum_{i\ne j_t} \frac{1}{(1+\eta_{t-1}\hat{\uL}_{t,i})^\alpha}).  \label{def: barst}
  \end{align}
\end{subequations}
Here, $c_v\geq 1.5 \cdot 2^{\frac{1}{\alpha}}$ is a fixed numerical constant. 
These auxiliary quantities in (\ref{def: params}) are chosen to match the analytical bounds for our adaptive exploration $\gamma_t$ and learning rates $\eta_t$.
Although we provide explicit numerical coefficients to establish the clean formal regret guarantees, they affect only multiplicative constants. 
In practical implementations, therefore, these coefficients may be treated as tuning parameters or simply set to one, although the theoretical guarantees in this paper correspond to the displayed choices.

Building on the surrogate-probability based construction of \citet{pmlr-v336-lee26a}, we use the following graph-adapted update rule for the inverse learning rate $\beta_t$ for $t\geq2$: \vspace{-0.1em}
\begin{equation}\label{def: lr}
  \beta_t = \min\bigg(\beta_{t-1} + \frac{1}{h_t}\qty(\sqrt{\frac{z_t}{\beta_{t-1}}} + \frac{u_t+v_t}{\beta_{t-1}}), 2^{\frac{1}{\alpha}}\beta_{t-1}\bigg),
\end{equation}
where
\begin{equation}\label{def: ht}
  h_t = \frac{4\alpha}{\alpha-1} \sum_{i \ne j_t} q_{t,i}^{1-\frac{1}{\alpha}}.
\end{equation}
Following \citet{tsuchiya2024simple}, we choose $\beta_1$ so that $\gamma_t \in (0,1/2]$ for any $t\in \sN$.
In particular, 
\begin{equation}\label{def: beta1 rough} \vspace{-0.4em}
  \beta_1 \geq \Omega\qty(\alpha \delta_* \log K, \alpha K^{1-\frac{1}{\alpha}}+\alpha\delta_*),
\end{equation}
whose precise value is given in (\ref{def: beta1}) in the appendix.
Notice that both $\hat{L}_t$ and $\eta_{t-1}$ are available at the end of round $t-1$, so $q_{t,i}$ and the auxiliary quantities can be computed before updating $\beta_t$ and determining $\gamma_t$. 
In contrast, $p_t$ depends on the current learning rate $\eta_t=\beta_t^{-1}$ and therefore introduce a circular dependence if used in these updates.
Avoiding such circular dependence is the main reason to construct the adaptive parameters using $q_t$ rather than $p_t$.
 \vspace{-0.1em}
\subsection{Unbiased loss estimator} \vspace{-0.1em}
One standard unbiased loss estimator under partial feedback is the importance-weighted (IW) estimator, defined using the probability that each arm is observed. 
In graph bandits, it is given by
\begin{equation*}\vspace{-0.3em}
  \hat{\ell}_{t,i}^{\mathrm{IW}} = \frac{O_{t,i}\ell_{t,i}}{P_{t,i}},
\end{equation*}
where $O_{t,i}= \I[i\in \onb(i_t)]$ and $P_{t,i}= \E[O_{t,i}|\hat{L}_t]$ is the observation probability.
Under the notation $y_t$ defined in~(\ref{def: total prob}), it holds $P_{t,i} = \sum_{j \in \inb(i)}y_{t,j}$.
Therefore, computing IW estimates is straightforward in FTRL framework since FTRL explicitly computes $w_t$ through regularized optimization, from which the overall distribution $y_t$ and thus every $P_{t,i}$ are obtained directly~\citep{ito2022nearly,tsuchiya2024simple}.

In contrast, FTPL does not explicitly compute $w_{t}$.
While $w_t$ can be evaluated numerically, doing so would weaken the computational advantage of FTPL.
In the multi-armed bandits, where an arm is observed only by a self-loop, geometric resampling addresses this difficulty by estimating $w_{t,i_t}^{-1}$ only, which repeatedly draws fresh perturbations and applies the FTPL rule until the originally selected arm $i_t$ is selected again~\citep{neu2016importance,chen2025geometric}. 
In the graph bandits, however, the IW estimator depends on the aggregated observation probability $P_{t,i}$ rather than the individual selection probability $w_{t,i}$.

To address graph feedback appropriately, we construct different loss estimators for $V_1$ and $V_2$ separately.
For arms in $V_1$, we only use observations obtained during forced exploration rounds, corresponding to $B_t=1$, and discard those obtained when $B_t=0$.
For $V_2$, we use conditional geometric resampling (CGR) to estimate only $w_{t,i_t}^{-1}$ and allocate the resulting estimate across the observed arms in $V_2$ with additional edge-dependent weights $e_{t,i,i_t}$ for $i\in \onb(i_t)$.
This construction is designed to avoid estimating $P_{t,i}^{-1}$ separately for every observed arm.
Overall, our loss estimator is defined by
\begin{align}\label{def: loss estimator}
  \hat{\ell}_{t,i}=
\begin{dcases}
  \frac{\I[B_t =1]O_{t,i}}{\gamma_t \rho_i} \ell_{t,i}, & i \in V_1, \\
  \frac{\I[B_t=0]O_{t,i}}{1-\gamma_t}e_{t,i,i_t}Z_t\ell_{t,i}, &i \in V_2,
\end{dcases}
\end{align}
where $Z_t = \sigma_{t,i_t} M_t$ for the number of CGR steps $M_t$ and the probability of forced observation for $i\in V_1$
\begin{equation}\label{def: rhoi}
  \rho_i = \Pr_{j\sim w^0}(i \in \onb(j)) = \frac{\sum_{j \in \inb(i)}x_{j}^*}{\delta_*}.
\end{equation}
Here, $x_j^*$ denotes an optimal solution to (\ref{def: linear program}) and we set $M_t=0$, i.e., CGR is not called, whenever $i_t \in V_1$.
To distribute the CGR estimates across the observed arms in $V_2$, we introduce an edge weight $e_{t,i,j}$, which is non-zero for each $i\in V_2$ and $j\in\inb_2(i)$.
For every off-diagonal edge, we set $e_{t,i,j}=\I[i\in \onb_2(j)]p_{t,j}/K$, which is a computable lower bound on $w_{t,j}$, and assign the remaining weights to the self-loop by defining
\begin{align}\label{def: edge weight}
  e_{t,i,i}= 1-\sum_{j \in \inb_2(i)\setminus\{i\}} e_{t,i,j}.
\end{align}
This definition is valid because every $i\in V_2$ has a self-loop and hence $i\in\inb_2(i)$.
Note that $p_{t,j}$ is computable when we update loss estimators (after observation) as $\eta_t$ is available before the decision of $i_t$.
The normalization of the edge weights ensures unbiasedness, while the bounds, $p_{t,j}/K\leq w_{t,j}$, control the contributions from off-diagonal edges.
These properties yield second-moment bounds proportional to $1/w_{t,i}$ rather than $1/P_{t,i}$, which recovers the dependence obtained in existing FTPL analyses for \emph{multi-armed bandits}~\citep{pmlr-v201-honda23a,lee2025nips}.
In particular, we have the following results.
\begin{lemma}\label{lem: estimator}
    The loss estimators in (\ref{def: loss estimator}) are unbiased and it holds that
    \begin{align*}
        \E\qty[\hat{\ell}_{t,i}^2 \middle| \hat{L}_t] \leq 
        \begin{cases}
            \delta_* /\gamma_t & \text{if } i \in V_1, \\
            4/w_{t,i}, &\text{if } i \in V_2.
        \end{cases}
    \end{align*}
\end{lemma}
The proof of Lemma~\ref{lem: estimator} is given in Appendix~\ref{app: estimator}.
\vspace{-0.5em}
\paragraph{Intuition behind the loss estimator.}
Existing FTRL approaches also addressed $V_1$ and $V_2$ separately, mainly through the choice of regularizer. 
For example, \citet{ito2022nearly} combined forced exploration with the Shannon entropy regularizer for arms in $V_1$ and use Tsallis entropy regularizer for arms in $V_2$, where Tsallis entropy was employed for BOBW guarantee in the multi-armed bandits~\citep{zimmert2021tsallis}.
Then, the standard IW estimator was used for both sets.
In contrast, our FTPL policy uses the same Pareto perturbation distribution for all arms and incorporates such distinctions into the loss estimator. 
In particular, we use observations obtained through forced exploration for $V_1$ and an MAB-style estimator based on CGR for $V_2$.
Thus, rather than constructing an FTPL analogue of a hybrid regularizer through different perturbation distributions, we maintain the same perturbations and adapt the estimators to their feedback structures.

For the edge-dependent weights, one naive choice could be $e_{t,i,j} = w_{t,j}/P_{t,i}$, which gives the upper bound on the second moment in terms of $1/P_{t,i}$ rather than $1/w_{t,i}$.
Our more conservative but computable construction instead applies the common value $e_{t,i,j}=p_{t,j}/K$ to the off-diagonal columns and assigns the remaining mass to its diagonal entry, which gives a bound proportional to $1/w_{t,i}$ rather than $1/P_{t,i}$. 
Although our estimator does not fully exploit all available observations, it is sufficient to establish the desired guarantees by recovering the MAB-type dependence in previous FTPL analyses.

\subsection{Conditional geometric resampling}
CGR is a computationally refined geometric resampling that restricts the sample space of perturbations to values satisfying a necessary condition for replaying the originally selected arm $i_t$~\citep{chen2025geometric}.
In this paper, we adopt the simplest variant, CGR I, described in Algorithm~\ref{alg: cgr}, which samples a perturbation vector conditional on the event that $r_{t,i_t}'$ is the largest among $\{r_{t,j}': \sigma_{t,j}\leq\sigma_{t,i_t}\}$.
This event has probability $1/\sigma_{t,i_t}$ by i.i.d.~sampling and is necessary for selecting $i_t$, where such conditional samples can be generated by drawing fresh i.i.d.~perturbations and swapping $r_{t,i_t}'$ with the largest perturbation among these arms.
Therefore, under this conditional sampling procedure, $i_t$ is selected with probability $\sigma_{t,i_t}w_{t,i_t}$, and thus $Z_t = \sigma_{t,i_t}M_t$ becomes an unbiased estimator of $w_{t,i_t}^{-1}$ for the number of CGR trials $M_t$ until $i_t$ is selected again.

The complete policy is given in Algorithm~\ref{alg: ftpl}, which combines forced exploration with probability $\gamma_t$, FTPL with Pareto perturbations, and graph-dependent loss estimator.
Although its full specification may appear complicated at first glance, the arm-selection rule simply follows the FTPL rule augmented with forced exploration.
The additional algorithmic steps are for computing the adaptive parameters and estimating the loss vectors.
Note that CGR is performed only when $B_t=0$ and $i_t\in V_2$, and we set $M_t=0$ otherwise.
\begin{lemma}\label{lem: CGR complexity}
  Let $r'$ be the resampled perturbations by Algorithm~\ref{alg: cgr}.
  Then, the number of resampling steps $M_t$ satisfies
  \begin{equation*}
    \E_{r'}[M_t | \hat{L}_t] \leq \log |V_2| +1.
  \end{equation*}
\end{lemma}
Since each CGR trial requires $\gO(K)$ time to generate and swap perturbations, the expected per-round computational cost by Algorithm~\ref{alg: cgr} is $\gO(K\log|V_2|)$.
On the other hand, computing the ranks $\sigma_{t,i}$ by sorting requires $\gO(K\log K)$ time, while all other operations take $\gO(K)$ time. 
Therefore, Algorithm~\ref{alg: ftpl} has an overall expected per-round computational cost of $\gO(K\log K)$.

\begin{algorithm}[t]
\caption{Conditional Geometric Resampling}
\label{alg: cgr}
\DontPrintSemicolon
\SetKwInOut{Input}{Input}
\Input{$\alpha>1$, cumulative loss $\hat{L}_t$, learning rate $\eta_t$,
         played arm $i_t\in V_2$.}
Define $\gA_{t} := \{j\in V:\hat{L}_{t,j}\le \hat{L}_{t,i_t}\}$.

Set $M_{t}\gets 0$.

\Repeat{$i_t=\argmin_{k\in V} \{\hat{L}_{t,k}-\beta_t r_k''\}$.}{
$M_{t}\gets M_{t}+1$.

Sample $r'=(r'_1,\ldots,r'_K)$ i.i.d.~from $\mathrm{Par}(\alpha)$.

Compute $k^\star \gets \argmax_{k\in\gA_{t}} r'_k$

Set $r''\gets r'$ and swap $r_{i_t}''$ and $r_{k^\star}''$.
}
\Return $M_t$.
\end{algorithm}
\begin{algorithm}[t]
\caption{FTPL for weakly observable graphs}
\label{alg: ftpl}
\DontPrintSemicolon
\SetKwInOut{Input}{Input}
\Input{Graph $G=(V,E)$, $\alpha\geq 2$.}

Set $V_1, V_2$ based on the existence of self-loops.

Solve linear program (\ref{def: linear program}) and find $\delta_*$ and $x^*$.

Set $w^0$ as in (\ref{def: w0}), $\beta_1$ as in (\ref{def: beta1 rough}), and $\hat{L}_1 =\mathbf{0}$.

For $i\in V_1$, define $\rho_i=\sum_{j \in \inb(i)}x_j^*/\delta_*$ as in (\ref{def: rhoi}).

\For{$t=1, \ldots, T$}{
Compute $z_t, u_t, v_t, \bar{s}_t, h_t, \gamma_t$ as in (\ref{def: gammat}--\ref{def: ht}).

Draw $B_t \sim \mathrm{Ber}(\gamma_t)$.

\If(\tcp*[f]{forced exploration}){$B_t=1$}{
  Play $i_t \sim w^0$.

  Observe losses $\{ \ell_{t,i}: i \in \onb(i_t) \}$.

  Set $\hat{\ell}_{t,i}=\I[i \in \onb_1(i_t)]\frac{\ell_{t,i}}{\gamma_t \rho_i}$.
}
\Else(\tcp*[f]{FTPL})
{
  Sample $r_{t,1},\ldots, r_{t,K}$ i.i.d.~from $\mathrm{Par}(\alpha)$.

  Play $i_t = \argmin_{i \in [K]} \{\hat{L}_{t,i}-\beta_t r_{t,i}\}$.

  Observe losses $\{ \ell_{t,i}: i \in \onb(i_t) \}$.

  \lIf{$i_t \in V_1$}{set $\hat{\ell}_{t,i}=0$, $\forall i\in V$.}

  \Else(\tcp*[f]{CGR}){
  Run Algorithm~\ref{alg: cgr} and set $Z_t =\sigma_{t,i_t}M_t$.

  \lFor{$i\in \onb_2(i_t)\setminus \{i_t\}$}{$e_{t,i,i_t} = \frac{p_{t,i_t}}{K}$}

  Set $e_{t,i_t, i_t}=1-\sum_{j \in \inb_2(i_t)\setminus \{i_t\}} \frac{p_{t,j}}{K}$.

  Set $\hat{\ell}_{t,i}= \I[i \in \onb_2(i_t)] \frac{e_{t,i,i_t} Z_t \ell_{t,i}}{1-\gamma_t}$.
}
}
Update $\hat{L}_{t+1} = \hat{L}_t + \hat{\ell}_t$.

Update $\beta_{t+1}$ as in (\ref{def: lr}).
}
\end{algorithm}

\subsection{Theoretical guarantee}\label{sec: guarantee}
Here, we provide the regret guarantees of the proposed policy, whose proofs are given in the appendix.
The first result considers adversarial regime.
\begin{theorem}\label{thm: adv}
  In adversarial regime, Algorithm~\ref{alg: ftpl} with $\alpha \geq 2$ satisfies that
  \begin{align*}
    \Reg_T \leq \gO\qty(\alpha \delta_*^{1/3} K^{1/(3\alpha)}T^{2/3}) + \gO\qty(\alpha \sqrt{(\delta_*+K^{1-1/\alpha})K^{1/\alpha}T}).
  \end{align*}
  If $T\geq K^3$ and $\alpha = 2\log K\geq 2$, this implies that
  \begin{equation*}
    \Reg_T \leq \gO(\log K \cdot \delta_*^{1/3}T^{2/3}).\vspace{-1em}
  \end{equation*}
\end{theorem}
This result matches the minimax dependence on $\delta_*$ and $T$ for weakly observable graph bandits up to logarithmic factors in $K$.
Since $\delta_* \leq\delta$, it also achieves the refined dependence on $\delta$.
The next result provides a gap-dependent guarantee under self-bounding constraints, which includes stochastic regimes with adversarial corruption as a special case.
\begin{theorem}\label{thm: sca}
  In adversarial regime with self-bounding constraints $(\Delta, C,T)$, Algorithm~\ref{alg: ftpl} with $\alpha \geq 2$ satisfies
  \begin{align*}
    \Reg_T \leq \gO\qty(\frac{\alpha^2 \delta_* K^{1/\alpha} \log K\log T}{\Delta_{\min}^2}+C) +\gO\qty(\frac{\gC(\delta_*, \alpha, K)\log T}{\Delta_{\min}}),
  \end{align*}
  where $\gC(\delta_*, \alpha, K)$ is independent of $T$.
  In particular, when $\alpha = 2\log K$, we obtain
  \begin{equation*}
    \Reg_T \leq \gO\qty(\frac{\delta_* \log^2 K \log T}{\Delta_{\min}^2}+\frac{ K\log^2 K \log T}{\Delta_{\min}}+C).
  \end{equation*}
\end{theorem}
Theorem~\ref{thm: sca} states a simplified gap-dependent bound for clarity, whose dependence on $C$ and $\Delta_{\min}$ can be loose in some extreme cases like $C=\Theta(T)$. 
In Appendix~\ref{app: MS type}, we provide a sharper MS-type bound that couples these quantities, which gives meaningful guarantees even for such extreme cases. 

The bounds in Theorems~\ref{thm: adv} and~\ref{thm: sca} exhibit weaker logarithmic dependences on $K$ than corresponding bounds obtained by FTRL policy~\citep{tsuchiya2024simple}. 
These additional logarithmic factors are due to the use of the surrogate quantity $q_t$, unlike $w_t$, which need not lie in the probability simplex. 
Similar phenomena were also observed in multi-armed bandits and bandits with expert advices~\citep{pmlr-v336-lee26a}.
\vspace{-0.4em}
\section{Numerical experiments}\label{sec: exp}\vspace{-0.2em}
We empirically evaluate the finite-time regret and computational efficiency of our FTPL policy.
This section presents some representative results, where the details and complete results over various graph structures and loss generations are provided in Appendix~\ref{app: exp}.
In all figures, the lines denote the empirical means and shaded regions indicate pointwise $95$\% confidence intervals.

\begin{figure*}[ht]
    \centering
    \begin{subfigure}[t]{0.47\textwidth}
        \centering
        \includegraphics[width=\linewidth]
        {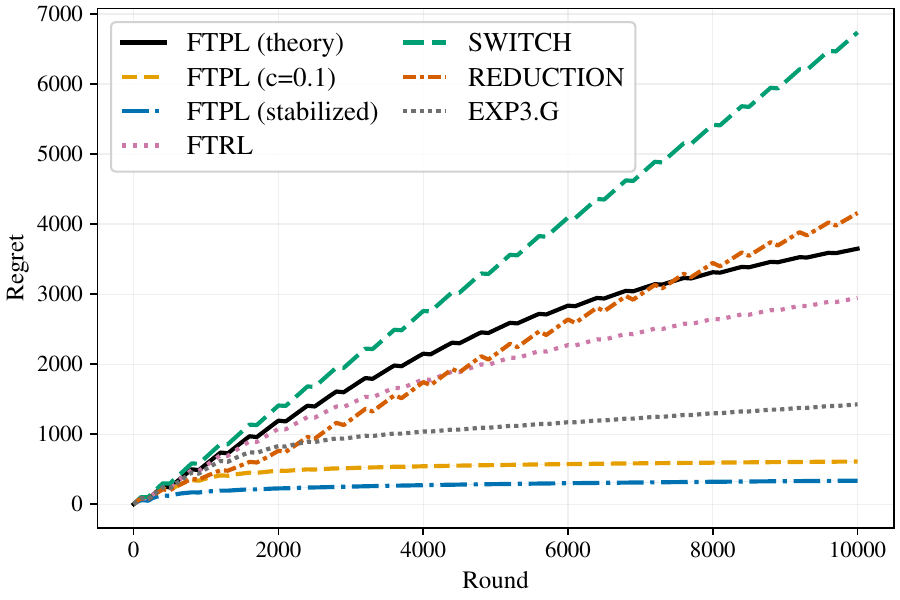}\vspace{-0.25em}
        \caption{Oblivious adversarial regime.}
        \label{fig: ring loop loopless ob ad}
    \end{subfigure}
    \hfill
    \begin{subfigure}[t]{0.47\textwidth}
        \centering
        \includegraphics[width=\linewidth]
        {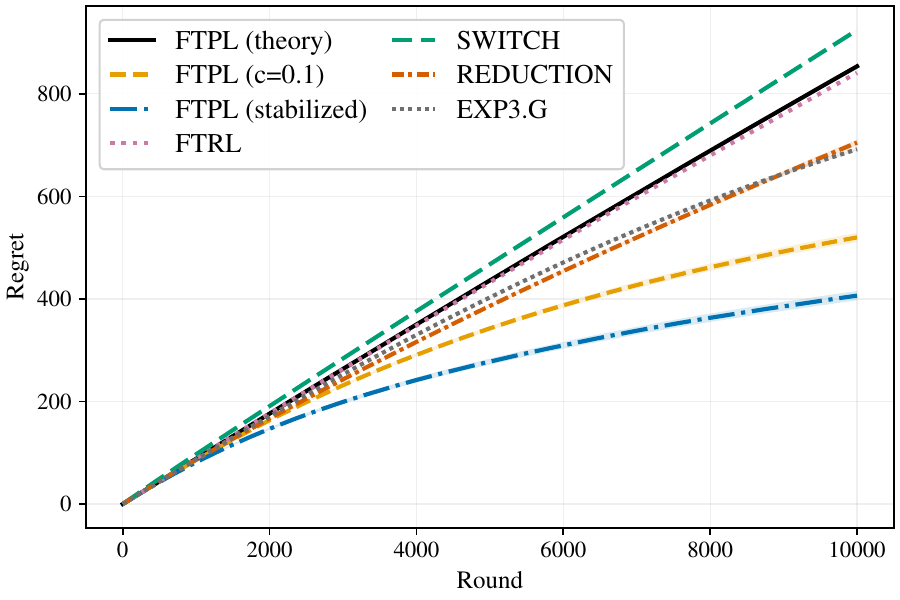}\vspace{-0.25em}
        \caption{Stochastic Bernoulli regime.}
        \label{fig: ring loop loopless st}
    \end{subfigure}
    \vspace{-0.25em}
    \caption{Regrets on the ring graph with three self-looped arms under different loss environments.} \vspace{-0.5em}
    \label{fig: ring loop loppless}
\end{figure*}

\textbf{Graph structure.} 
For the main comparison, we use a partially looped bidirected ring with $K=8$ arms.
Each arm reveals the losses of its two neighboring arms, while only three arms additionally reveal their own losses.
Thus, $|V_1|=5$ and $|V_2|=3$ in our notation.
In this graph, the three different domination numbers coincide, i.e., $\delta=\delta_*=\delta_*'=3$.
Hence, differences in empirical performance cannot be attributed to the particular domination number used by each policy.
In both environments, we choose the optimal arm from $V_1$ to exclude direct self-observation.
The complementary case in which the optimal arm belongs to $V_2$ is also given in the appendix.

\textbf{Environments.} 
In the adversarial regime, we use a deterministic oblivious sequence with $T=10000$ and blocks of length $100$.
In the first block of every four-block cycle, one suboptimal arm outperforms the intended best arm, while their roles are reversed in the remaining three blocks.
In particular, we set their cumulative losses to $7500$ and $2500$, respectively, making the intended best arm unique.
In the stochastic regime, losses are independently drawn from Bernoulli, where the unique optimal arm has mean loss $0.3$, while other remaining seven arms has mean $0.4$.

\textbf{Compared policies.} 
We compare our policy with (i) FTRL policy with the stability-penalty-bias matching adaptive learning rate of \citet{tsuchiya2024simple}, (ii) the testing and switching policy of \citet{kong2022simultaneously}, (iii) the hierarchical reduction of \citet{dann2023blackbox}, and (iv) the baseline EXP3.G~\citep{alon2015online}.
For brevity, we refer to these policies as \textsc{Ftrl}, \textsc{Switch}, \textsc{Reduction}, and  \textsc{Exp3.G}, respectively.

In addition, we consider three variants of our policy.
FTPL (theory) follows exactly the theoretical parameter choices specified in (\ref{def: params}--\ref{def: beta1 rough}).
FTPL ($c=0.1$) multiplies parameters in (\ref{def: params}) by $0.1$, i.e., except $h_t$ since $\beta_t$ is determined by the ratio of $h_t$ to other parameters.
Based on this configuration, FTPL (stabilized) further sets $\beta_1=4$ and clips the exploration rate $\gamma_t$ at $1/2$, which modifies both the initialization and the exploration rate.
All three FTPL variants use the same Pareto perturbation with $\alpha=\max\{2,2\log K\}$ and the same conditional geometric resampling estimator.

Note that rescaling by $0.1$ affects only the multiplicative constants and does not change the order of regret.
In contrast, clipping $\gamma_t$ may violate some arguments in the proof.
Hence, FTPL (stabilized) does not automatically satisfy our regret guarantees.

\textbf{Regret comparison.} 
Figure~\ref{fig: ring loop loppless} shows the results on the ring under the two environments, averaged over $600$ independent runs.
FTPL (theory) performs competitively with the baselines in both settings.
Consistent with the conservative nature of the theoretical parameter choices, the two practical FTPL variants substantially improve upon the theoretical configuration in these experiments.
In particular, setting $c=0.1$ improves finite-time performance without changing the main structure of the policy.
FTPL (stabilized), which combines a modified initialization with clipping, provides a further improvement.
Both practical variants attain lower regret than all the baselines in the adversarial and stochastic settings.
The same overall pattern persists across the additional graph structures and loss constructions given in the appendix.

\begin{figure}
   \centering
        \includegraphics[width=0.5\linewidth]
        {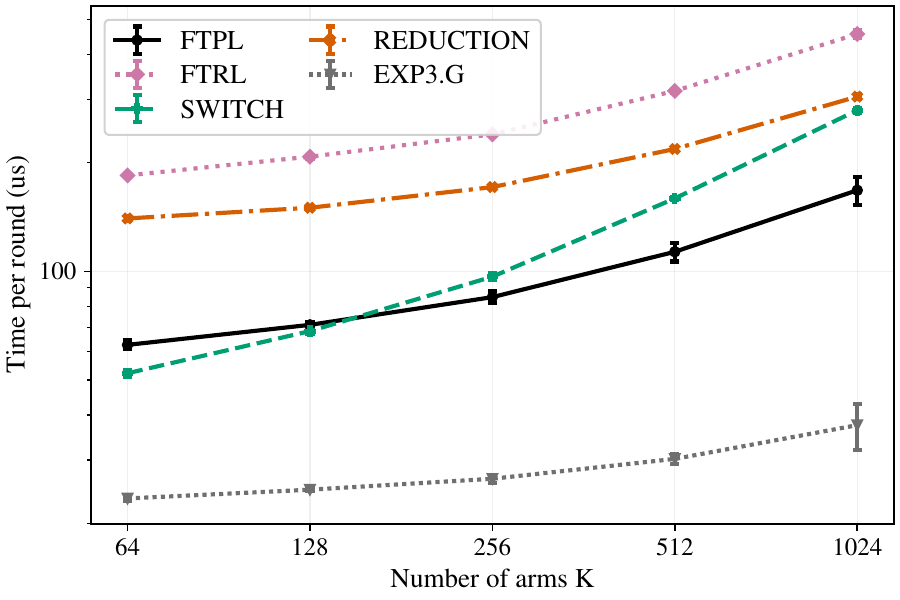}\vspace{-0.25em}
        \caption{Runtime for different $K$.}
        \label{fig: run time}
        \vspace{-1em}
\end{figure}

\textbf{Runtime.} 
We consider a graph where arm $1$ observes arms $1$ and $2$, arm $2$ has no out-neighbor, and each remaining arm observes only itself.
Thus, $|V_1|=1$ and $|V_2|=K-1$, making a computationally intensive setting in which FTPL may invoke CGR frequently.
Figure~\ref{fig: run time} shows the average runtime per round as the number of arms increases over $100$ independent runs, where the error bars indicate one standard deviation.
As expected, \textsc{Exp3.G} is the fastest policy due to its simple update rule.
Among the BOBW policies, our policy exhibits better runtime scaling and achieves the lowest runtime for moderate and large values of $K$.
\vspace{-0.3em}
\section{Conclusion}\vspace{-0.2em}
This paper studied online learning with weakly observable feedback graphs.
We proposed an FTPL policy that combines forced exploration with a graph-aware loss estimator based on conditional geometric resampling.
Combined with adaptive learning rates defined using surrogate probabilities, our policy achieves near-optimal regret in adversarial environments and polylogarithmic regret in stochastic environments.
Our guarantees also extend to adversarial environments with self-bounding constraints.
Moreover, the policy preserves the standard FTPL structure and can be implemented in nearly linear expected time per round, independently of the number of edges.
Our numerical results support the practical effectiveness of the proposed approach.
More broadly, our results may indicate the importance of designing loss estimators that exploit the feedback structure while preserving the computational simplicity of FTPL.
A natural direction is to extend these ideas to general partial monitoring problems or dueling bandits, where observations need not directly reveal individual losses.

\subsection*{AI use statement}
In this work, we used the generative AI tool ChatGPT-5.6 Sol model, to formalize definitions and mathematical claims, particularly those concerning weak observability and weak domination, suggest ingredients for proofs based on the ideas provided by the authors, assist in writing proofs, refine the experimental design, generate synthetic graph instances for experiments, particularly the specific domination number graph structure used in Appendix~\ref{app: exp}, implement methods in Python.
We did not use generative AI tools for developing the conceptual framework, interpretation of the results, writing the manuscript from the beginning, and translation or for cleaning or reformatting datasets.
Qualitative and thematic data analysis was not applicable to this work.

Additionally, we used the tool for brainstorming, searching for related literature, identifying relevant references, suggesting experimental settings, generating and editing software code, creating and modifying graph visualizations with tikz, revising the manuscript, improving readability, and suggesting the title of the paper.

We reviewed all AI-assisted work. 
All mathematical statements and proofs were checked and written line by line by the authors. 
All AI-assisted code was reviewed and tested against the stated algorithms, and the reported experimental results were generated and verified by the authors.
All final interpretations of the results were independently determined and verified by the authors. 
All references and literature summaries were checked against the original sources through a manual literature review.
The authors revised and approved all AI-assisted text.
We take responsibility for the final content of this work, including text, claims, and artifacts produced with the aid of generative AI.

\bibliographystyle{plainnat}
\bibliography{ref}

\newpage
\appendix
\phantomsection 
\addcontentsline{toc}{section}{Appendices} 
\section*{Appendix Contents} 
\startcontents[appendix]    
\printcontents[appendix]{l}{1}{} 
\newpage

\section{Proofs of Lemmas for Section~\ref{sec: policy}}
In this section, we provide the proofs of results directly related to proposed methods.
\subsection{Proof of Lemma~\ref{lem: estimator}}\label{app: estimator}
Here, we present the detailed proof of Lemma~\ref{lem: estimator}.
Before the proof, we first introduce a useful result on the relationship between $e_{t,i,j}$ and $w_{t,j}$, whose proof is deferred to Appendix~\ref{app: lemma ew}.
\begin{lemma}\label{lem: e and w}
  For any $i\in V$ and $t$, $p_{t,i}/K \leq w_{t,i}$ holds.
\end{lemma}
Notice that this lemma implies that $e_{t,i, j}\leq w_{t,j}, \, \forall j \in \inb_2(i) \setminus \{i\}$.

\begin{proof}[Proof of Lemma~\ref{lem: estimator}]
  By definition of $B_t$ and the construction of the proposed policy, the unbiasedness of $\hat{\ell}_{t,i}$ for $i\in V_1$ follows immediately.
  Since we construct non-zero loss estimator for $V_2$ only when it is observed by $i_t \in V_2$, for $i\in V_2$, we have
  \begin{align*}
    \E[\hat{\ell}_{t,i}|\hat{L}_t] = \sum_{j \in \inb_2(i)} \frac{(1-\gamma_t)w_{t,j}}{1-\gamma_t}e_{t,i,j} \E[Z_t|\hat{L}_t, i_t =j] \ell_{t,i} = \sum_{j \in \inb_2(i)} e_{t,i,j} \ell_{t,i} = \ell_{t,i},
  \end{align*}
  where the second equality follows from Lemma 2 of \citet{chen2025geometric}, which states $\E[Z_t|\hat{L}_t, i_t=j]=w_{t,j}^{-1}$, and the last equality follows from the definition of $e_{t,i,j}$ in (\ref{def: edge weight}).

  Next, for the conditional second moment of $i \in V_1$, we have
  \begin{align*}
    \E\qty[\hat{\ell}_{t,i}^2 \middle| \hat{L}_t] =  \E\qty[ \I[B_t =1, i\in \onb(i_t)] \frac{\ell_{t,i}^2}{\gamma_t^2 \rho_i^2} \middle| \hat{L}_t] = \frac{\ell_{t,i}^2}{\gamma_t \rho_i} = \frac{\delta_* \ell_{t,i}^2}{\sum_{j\in \inb(i)}x_j^* \gamma_t} \leq  \frac{\delta_*}{\gamma_t},
  \end{align*}
  where the last equality follows from the definition of $\rho_i$ in (\ref{def: rhoi}) and the last inequality follows from the fact that $\ell_{t,i} \in [0,1]$ and $\sum_{j \in \inb(i)} x_j^* \geq 1$ holds for any $i\in V_1$ by linear program in (\ref{def: linear program}).

  By the design of Algorithm~\ref{alg: cgr}, $M_t$ follows the geometric distribution with success probability $\sigma_{t,i_t}w_{t,i_t}$.
  Then, for $i\in V_2$, we have
  \begin{align*}
      \E\qty[\hat{\ell}_{t,i}^2 \middle| \hat{L}_t] &=  \E\qty[ \I[B_t =0, i\in \onb_2(i_t)] \frac{e_{t,i,i_t}^2 Z_t^2 \ell_{t,i}^2}{(1-\gamma_t)^2} \middle| \hat{L}_t] \\
      &= \sum_{j \in \inb_2(i)} (1-\gamma_t)w_{t,j} \qty(\frac{e_{t,i,j}\ell_{t,i}}{1-\gamma_t})^2 \E\qty[Z_t^2 \middle | \hat{L}_t, i_t=j] \tag{$\because \I[i \in \onb(j)] =  \I[j \in \inb(i)]$}\\
      &= \frac{\ell_{t,i}^2}{1-\gamma_t} \sum_{j \in \inb_2(i)} w_{t,j}e_{t,i,j}^2 \cdot \E\qty[Z_t^2 \middle | \hat{L}_t, i_t=j] \\
      &\leq \frac{\ell_{t,i}^2}{1-\gamma_t} \sum_{j \in \inb_2(i)} w_{t,j}e_{t,i,j}^2 \cdot \sigma_{t,j}^2\frac{2-\sigma_{t,j}w_{t,j}}{\sigma_{t,j}^2w_{t,j}^2} \numberthis{\label{eq: CGR 2nd moment}}\\
      &\leq \frac{2}{1-\gamma_t} \sum_{j \in \inb_2(i)}  \frac{e_{t,i,j}^2}{w_{t,j}}.
  \end{align*}
  Let $e_{t,i}^\Sigma = \sum_{j \in \inb_2(i) \setminus \{i\}} e_{t,i,j}$.
  Then, by the definition of $e_{t,i,j}$ in (\ref{def: edge weight}) and Lemma~\ref{lem: e and w}, we obtain
  \begin{align*}
    \sum_{j \in \inb_2(i)} \frac{e_{t,i,j}^2}{w_{t,j}} &= \frac{(1-e_{t,i}^\Sigma)^2}{w_{t,i}} + \sum_{j \in \inb_2(i)\setminus \{i\}} \frac{e_{t,i,j}^2}{w_{t,j}}  \tag{by definition (\ref{def: edge weight})} \\ 
    &\leq \frac{(1-e_{t,i}^\Sigma)^2}{w_{t,i}} + \sum_{j \in \inb_2(i)\setminus \{i\}} e_{t,i,j} \tag{by Lemma~\ref{lem: e and w}}\\
    &= \frac{(1-e_{t,i}^\Sigma)^2}{w_{t,i}}  + e_{t,i}^\Sigma \\
    &\leq \frac{(1-e_{t,i}^\Sigma)^2+ e_{t,i}^\Sigma}{w_{t,i}} \tag{$\because w_{t,i} \in (0,1)$} \\
    &\leq \frac{1-e_{t,i}^\Sigma+ (e_{t,i}^\Sigma)^2}{w_{t,i}} \leq \frac{1}{w_{t,i}},
  \end{align*}
  where the last inequality follows from $e_{t,i}^\Sigma \in [0,1)$.
  Consequently, 
  \begin{equation*}
    \E\qty[\hat{\ell}_{t,i}^2 \middle| \hat{L}_t] \leq \frac{2}{(1-\gamma_t)w_{t,i}} \leq \frac{4}{w_{t,i}},
  \end{equation*}
  where the last inequality follows from $\gamma_t \in (0,1/2]$ as defined in (\ref{def: gammat}), which concludes the proof.
\end{proof}

\subsection{Proof of Lemma~\ref{lem: CGR complexity}}
Here, we present the detailed proof of Lemma~\ref{lem: CGR complexity}.
\begin{proof}
  The result follows directly from the proof of Lemma~2 in \citet{chen2025geometric}, which analyzes the computational complexity of CGR~I, CGR with value swapping (Algorithm~\ref{alg: cgr}).
  When $r'$ denotes the resampled perturbations under value swapping, Lemma 2 of \citet{chen2025geometric} shows that
  \begin{equation*}
    \E_{r'}[M_t | \hat{L}_t, i_t] = \frac{1}{\sigma_{t,i_t} w_{t,i_t}}.
  \end{equation*}
  Therefore, the expected number of resampling steps in Algorithm~\ref{alg: cgr} satisfies that
  \begin{align*}
    \E_{r'}\qty[M_t | \hat{L}_t] &= \sum_{i \in V_2} \Pr[i_t = i | \hat{L}_t] \E_{r'}\qty[M_t | \hat{L}_t, i_t=i] \\
    &= \sum_{i\in V_2} w_{t,i} \frac{1}{\sigma_{t,i} w_{t,i}} \\
    &\leq 1 + \int_1^{|V_2|} \frac{1}{x} \dd x \\
    &= \log |V_2| + 1.
  \end{align*}
  Notice that we only consider the case of $i_t \in V_2$ rather than the summation over $V$.
  This is because Algorithm~\ref{alg: cgr} (CGR~I) runs only when $i_t \in V_2$ under Algorithm~\ref{alg: ftpl}.
\end{proof}

\subsection{Proof of Lemma~\ref{lem: e and w}}\label{app: lemma ew}
Here, we provide the proof of Lemma~\ref{lem: e and w}
\begin{proof}
  By the decision rule of FTPL, we have
  \begin{align*}
    w_{t,i} = \int_0^\infty f(z+\eta_t\hat{\uL}_{t,i}) \prod_{j \ne i} F(z+\eta_t\hat{\uL}_{t,j}) \dd z.
  \end{align*}
  Here, $f(z)=\alpha/(z+1)^{\alpha+1}$ for $z\geq 0$ by the choice of Pareto distribution with shape $\alpha>0$.
  Since $\hat{\uL}_{t,i} \geq 0$ for all $i\in V$ and $F$ is increasing, for $\ul_i = \eta_t \hat{\uL}_{t,i}$, we have
  \begin{align*}
    w_{t,i} &\geq \int_0^\infty f(z+\eta_t\hat{\uL}_{t,i}) F(z)^{K-1} \dd z \\
    &= \int_0^\infty \frac{\alpha}{(z+\ul_i+1)^{\alpha+1}} F(z)^{K-1} \dd z \\
    &= \frac{1}{(1+\ul_i)^\alpha} \int_0^\infty \frac{\alpha}{(1+x)^{\alpha+1}} F((1+\ul_i)x)^{K-1} \dd x \tag{by $z:= (1+\ul_i)x$} \\
    &\geq \frac{1}{(1+\ul_i)^\alpha} \int_0^\infty \frac{\alpha}{(1+x)^{\alpha+1}} F(x)^{K-1} \dd x \tag{$\because \ul_i \geq 0$} \\
    &= \frac{1}{K(1+\ul_i)^\alpha} \geq \frac{p_{t,i}}{K},
  \end{align*} 
  which concludes the proof.
\end{proof}

\section{Proof sketch}\label{sec: analysis}
In this section, we provide an outline of the analysis in Section~\ref{sec: guarantee}, where the detailed proofs are given in Appendices~\ref{app: analysis}--\ref{app: sca}.
At a high level, our analysis reduces the regret bound to a form analyzed by \citet{ito2022nearly} and \citet{tsuchiya2024simple}.
\begin{lemma}\label{lem: decomposition}
  For $z_t, u_t, v_t$ defined in (\ref{def: params}), Algorithm~\ref{alg: ftpl} with $\beta_t$ defined in (\ref{def: lr}) and $\alpha \geq 2$ satisfies 
  \begin{equation}\label{eq: regret to stab and pen}
    \Reg_T \leq \sum_{t=1}^T 3 \E\qty[\sqrt{\frac{z_t}{\beta_t}} + \frac{u_t + v_t}{\beta_t}] + \gO\qty( \max\qty(\delta_* \log K, K^{1-\frac{1}{\alpha}} + \delta_*)\alpha K^{\frac{1}{\alpha}}).
  \end{equation}
\end{lemma}
To control the first two leading terms in (\ref{eq: regret to stab and pen}), for any positive sequences, $a_{1:n}=\{a_1, \ldots, a_{n}\}$ and $b_{1:n}=\{b_1, \ldots, b_n\}$ of length $n \in \sN$, define
  \begin{equation}\label{def: G function}
    \begin{aligned}
      G_1(a_{1:n}, b_{1:n}) &:= \sum_{t=1}^n \frac{\sqrt{a_t}}{(\sum_{s=1}^t \sqrt{a_s}/b_s)^{1/3}}, \\
      G_2(a_{1:n}, b_{1:n}) &:= \sum_{t=1}^n \frac{a_t}{\sqrt{\sum_{s=1}^t a_s/b_s}},
    \end{aligned}
  \end{equation}
and $a_{\max}$ denote the maximum of given sequence $a_{1:n}$. 
The following lemma reduces leading terms in (\ref{eq: regret to stab and pen}) to the functionals $G_1,G_2$ in (\ref{def: G function}).
\begin{lemma}\label{lem: reg to G}
    Let $\gT := \{t \in [T]: \beta_{t} < 2^{1/\alpha}\beta_{t-1}\}$.
    Then, for $z_t, u_t, v_t, \beta_t$ defined in (\ref{def: params}--\ref{def: lr}), $G_1,G_2$ defined in (\ref{def: G function}), and $\alpha \geq 2$, we have
    \begin{align*}
        \sum_{t=1}^T \sqrt{\frac{z_t}{\beta_t}} &\leq 2^{\frac{1}{4}}G_1(\{z_{s}\}_{s\in \gT}, \{h_{s}\}_{s\in \gT}) + \alpha, \\[-0.1em]
        \sum_{t=1}^T \frac{u_t + v_t}{\beta_t} &\leq \sqrt{2}G_2(\{u_{s}+v_s\}_{s\in \gT},\{h_{s}\}_{s\in \gT}) + \frac{\alpha}{2}.
    \end{align*}
\end{lemma}
Hence, it remains to bound the $G_1,G_2$ terms, which can be done by applying Lemma~5 of \citet{tsuchiya2024simple} and Lemma~3 of \citet{ito2024adaptive}. 
In the adversarial regime, we obtain
\begin{align*}
  G_1(\{z_{s}\}_{s\in \gT}, \{h_{s}\}_{s\in \gT}) &\leq \frac{3}{2}(h_{\max}z_{\max})^{\frac{1}{3}} T^{\frac{2}{3}}, \\
  G_2(\{u_{s}+v_s\}_{s\in \gT},\{h_{s}\}_{s\in \gT})&\leq 2\sqrt{h_{\max}(u_{\max}+ v_{\max})T},
\end{align*}
which concludes the proof of Theorem~\ref{thm: adv} by substituting bounds on parameter.
The corresponding bounds under the AwSC regime are provided in Appendix~\ref{app: sca}.

\section{Proof of Lemmas for Appendix~\ref{sec: analysis}}\label{app: analysis}
In this section, we provide the detailed proofs for results given in Appendix~\ref{sec: analysis}.
Before the proof, we define $\phi(\lambda)$ denotes the arm-selection probability by FTPL with Pareto perturbations for given $\lambda\in \sR^K$, whose $i$-th element is written by
\begin{equation*}
  \phi_i(\lambda) = \int_0^\infty f(z+\ul_{i}) \prod_{j\ne i} F(z+\ul_j) \dd z.
\end{equation*} 
Notice that $w_{t,i}= \phi_i(\eta_t \hat{L}_t)$.
We use this notation throughout the remainder of the paper.

\subsection{Proof of Lemma~\ref{lem: decomposition}: regret decomposition}
Here, we provide the proofs for the regret decomposition.
We first introduce three auxiliary lemmas used in the proof of Lemma~\ref{lem: decomposition}, whose proofs are given in Appendix~\ref{app: auxiliary}.
\begin{lemma}\label{lem: stability all}
  Algorithm~\ref{alg: ftpl} satisfies that
  \begin{align*}
    \E\qty[\inp{\hat{\ell_t}}{\phi(\eta_t \hat{L}_t) - \phi(\eta_t \hat{L}_{t+1})}\middle | \hat{L}_t] \leq \frac{z_t}{\beta_t \gamma_t} + \frac{u_t}{\beta_t}.
  \end{align*}
\end{lemma}
\begin{lemma}\label{lem: max parameters}
  For $z_t, u_t, v_t$ in (\ref{def: params}) and $h_t$ in (\ref{def: ht}), it holds for any $t\in \sN$ that
  \begin{align*}
    z_t &\leq z_{\max} = (2e\sqrt{2}+3\sqrt{2})\alpha \delta_* \log K \\
    u_t &\leq u_{\max}= (4e\sqrt{2}+24\sqrt{2})\frac{\alpha^2}{\alpha-1}K^{1-\frac{1}{\alpha}}\\
    v_t &\leq v_{\max} = c_v (\alpha+1) \delta_*\\
    h_t &\leq h_{\max} = \frac{4\alpha^2}{\alpha-1} K^{\frac{1}{\alpha}}.
  \end{align*} 
\end{lemma}
Therefore, we choose
\begin{equation}\label{def: beta1}
  \beta_1 \geq \max\qty(16z_{\max}, 4(u_{\max}+v_{\max})) \implies \gamma_t \in (0,1/2].
\end{equation}
\begin{lemma}[Lemma 8 of \citet{pmlr-v336-lee26a}]\label{lem: p and q}
  For $p,q$ in (\ref{def: p and q}), $q_{t,i}\leq p_{t,i} \leq 2q_{t,i}$ holds for all $t\in \sN$ and $\alpha\geq 2$.
\end{lemma}

\begin{proof}[Proof of Lemma~\ref{lem: decomposition}]
First, we provide a general regret decomposition. We have
\begin{align*}
  \Reg_T &= \E\qty[\sum_{t=1}^T \inp{\ell_t}{y_t - e_{i^*}}] \\
  &= \E\qty[\sum_{t=1}^T \inp{\ell_t}{(1-\gamma_t)w_t - e_{i^*}}] +  \E\qty[\sum_{t=1}^T \inp{\ell_t}{\gamma_t w^0 }] \\
  &\leq \E\qty[\sum_{t=1}^T \inp{\ell_t}{w_t - e_{i^*}}] +  \E\qty[\sum_{t=1}^T \gamma_t] \tag{$\because \gamma_t \in (0, 1/2], \, \ell_{t,i} \in [0,1]$} \\
  &= \E\qty[\sum_{t=1}^T \inp{\hat{\ell}_t}{w_t - e_{i^*}}] +  \E\qty[\sum_{t=1}^T \gamma_t],
\end{align*}
where the last equality follows from the unbiasedness of the loss estimator established in Lemma~\ref{lem: estimator}.
Recall that $w_t$ is the probability vector induced by the FTPL rule at round $t$, with $w_{t,i}$ denoting the probability of arm $i$ being selected.
Therefore, we can directly apply the previous FTPL regret decomposition to the first term~\citep[Lemma 3.4]{kim2025follow}, which yields
\begin{align*}
  \E\qty[\sum_{t=1}^T \inp{\hat{\ell}_t}{w_t - e_{i^*}}]  \leq \sum_{t=1}^T \E\qty[\inp{\hat{\ell}_t}{\phi(\eta_t \hat{L}_t)-\phi(\eta_t \hat{L}_{t+1})}] + \sum_{t=1}^{T+1} (\beta_t - \beta_{t-1}) \E_{r_{t,\cdot} \sim \mathrm{Par}(\alpha)}[r_{t,i_t} - r_{t,i^*}].
\end{align*}
Here, we set $\beta_0=0$ for the analysis, which is introduced solely for notational convenience and does not appear in the actual algorithm.
Since the second term of RHS solely depends on the perturbation distributions, we can directly apply the previous results in \citet[Lemma 7]{pmlr-v336-lee26a}, which yields
\begin{align*}
  (\beta_t - \beta_{t-1}) \E_{r_{t,\cdot} \sim \mathrm{Par}(\alpha)}[r_{t,i_t} - r_{t,i^*}|\hat{L}_t] &=(\beta_t - \beta_{t-1}) \E_{r_{t,\cdot} \sim \mathrm{Par}(\alpha)}[r_{t,i_t} - r_{t,j_t}|\hat{L}_t]\\
  &\leq \sum_{i \ne j_t} \frac{2\alpha}{\alpha-1} p_{t,i}^{1-\frac{1}{\alpha}}.
\end{align*}
The first equality follows since $j_t$ is fixed given $\hat{L}_t$ and the perturbations $r_{t,i}$s are independently and identically distributed, so that $r_{t,j_t}$ and $r_{t,i^*}$ have the same expectation conditional on $\hat{L}_t$.
Therefore, by Lemma~\ref{lem: p and q} and $\beta_t$ in (\ref{def: lr}), we have
\begin{align*}
  \sum_{t=1}^{T+1} (\beta_t - \beta_{t-1}) \E_{r_{t,\cdot} \sim \mathrm{Par}(\alpha)}[r_{t,i_t} - r_{t,i^*}| \hat{L}_t] &\leq \sum_{t=1}^{T+1} (\beta_t - \beta_{t-1}) \sum_{i \ne j_t} \frac{2^{2-\frac{1}{\alpha}}\alpha}{\alpha-1} q_{t,i}^{1-\frac{1}{\alpha}}\tag{by Lemma~\ref{lem: p and q}}\\
  &=\sum_{t=1}^{T+1} 2^{-\frac{1}{\alpha}} (\beta_t - \beta_{t-1}) h_t \tag{by definition of $h_t$ in (\ref{def: ht})} \\
  &= \sum_{t=2}^{T+1} 2^{-\frac{1}{\alpha}} (\beta_t - \beta_{t-1}) h_t + \beta_1 h_1 \tag{$\because \beta_0=0$} \\
  &\leq \sum_{t=2}^{T+1} 2^{-\frac{1}{\alpha}} \sqrt{\frac{z_t}{\beta_{t-1}}} + 2^{-\frac{1}{\alpha}} \frac{u_t+v_t}{\beta_{t-1}} + \beta_1 h_1  \tag{by (\ref{def: lr})}\\
  &\leq \sum_{t=1}^{T} \sqrt{\frac{z_t}{\beta_{t}}} + \frac{u_t+v_t}{\beta_{t}} +\beta_1 h_{\max} + \sqrt{\frac{z_{\max}}{\beta_{1}}} + \frac{u_{\max}+v_{\max}}{\beta_1},
\end{align*}
where the last inequality holds since $\beta_t$ is monotonically increasing for $t$ and satisfies $\beta_t \leq 2^{1/\alpha}\beta_{t-1}$.
By Lemma~\ref{lem: max parameters} and the choice of $\beta_1$ in (\ref{def: beta1}), we have for $\alpha \geq 2$ that
\begin{align*}
  \beta_1 h_{\max} + \sqrt{\frac{z_{\max}}{\beta_{1}}} + \frac{u_{\max}+v_{\max}}{\beta_1} &\leq \beta_1 h_{\max} + \frac{1}{2} \\
  &\leq \gO\qty( \max\qty(\delta_* \log K, K^{1-\frac{1}{\alpha}} + \delta_*)\alpha K^{\frac{1}{\alpha}}).
\end{align*}
Therefore, Lemma~\ref{lem: stability all} implies that
\begin{align*}
  \Reg_T &\leq \E\qty[\sum_{t=1}^T \frac{z_t}{\beta_t\gamma_t} + \frac{u_t}{\beta_t} + \sqrt{\frac{z_t}{\beta_t}} + \frac{u_t+v_t}{\beta_t}+\gamma_t] + \gO\qty( \max\qty(\delta_* \log K, K^{1-\frac{1}{\alpha}} + \delta_*)\alpha K^{\frac{1}{\alpha}}) \\
  &\leq \E\qty[\sum_{t=1}^T \frac{z_t}{\beta_t\gamma_t} + \sqrt{\frac{z_t}{\beta_t}} + 2\frac{u_t+v_t}{\beta_t}+\gamma_t] + \gO\qty( \max\qty(\delta_* \log K, K^{1-\frac{1}{\alpha}} + \delta_*)\alpha K^{\frac{1}{\alpha}}).
\end{align*}
Here, by the definition of $\gamma_t$ in (\ref{def: gammat}), we have
\begin{align*}
  \frac{z_t}{\beta_t\gamma_t} +\gamma_t\leq 2\sqrt{\frac{z_t}{\beta_t}} + \frac{u_t+v_t}{\beta_t},
\end{align*}
which concludes the proof.
\end{proof}

\subsection{Proof of Lemma~\ref{lem: reg to G}}
Here, we provide the proof for deriving upper bounds on the auxiliary functions in the regret analysis.
\begin{proof}[Proof of Lemma~\ref{lem: reg to G}]
  Let $\gT := \{t \in [T]: \beta_{t} < 2^{1/\alpha}\beta_{t-1}\}$, so that $ \gT^c=  \{t \in [T]: \beta_{t} = 2^{1/\alpha}\beta_{t-1}\}$ by the definition of $\beta_t$ in (\ref{def: lr}).
  Then, it holds that
  \begin{align*}
      \sum_{t=1}^T \sqrt{\frac{z_t}{\beta_t}} &=  \sum_{t \in \gT^c} \sqrt{\frac{z_t}{\beta_t}} + \sum_{t \in \gT} \sqrt{\frac{z_t}{\beta_t}} \\
      &\leq  \sum_{t \in \gT^c} \sqrt{\frac{z_{\max}}{\beta_t}} + \sum_{t \in \gT} \sqrt{\frac{z_t}{\beta_t}} \\
      &\leq \sum_{s=0}^\infty \qty(\frac{1}{2})^{\frac{s}{2\alpha}} \sqrt{\frac{z_{\max}}{\beta_1}} + \sum_{t \in \gT} \sqrt{\frac{z_t}{\beta_t}} \\
      &= \frac{1}{1-2^{-1/(2\alpha)}}  \sqrt{\frac{z_{\max}}{\beta_1}} + \sum_{t \in \gT} \sqrt{\frac{z_t}{\beta_t}} \\
      &\leq 3.4\alpha \sqrt{\frac{z_{\max}}{\beta_1}} + \sum_{t \in \gT} \sqrt{\frac{z_t}{\beta_t}}.
  \end{align*}
  The last inequality follows from 
  \begin{align*}
      \frac{1}{1-2^{-1/(2\alpha)}} = \frac{1}{1-e^{-\log2/(2\alpha)}} \leq 1+ \frac{2\alpha}{\log 2} \leq \qty(\frac{1}{2}+\frac{2}{\log 2})\alpha \leq 3.4\alpha,
  \end{align*}
  for $\alpha\geq 2$ and $1+e^{-x} \geq \frac{x}{1+x}$ for $x>0$.
  Similarly, we can obtain
  \begin{align*}
      \sum_{t=1}^T \frac{u_t + v_t}{\beta_t}  \leq \qty(\frac{1}{2}+\frac{1}{\log 2})\alpha\frac{u_{\max} + v_{\max}}{\beta_1} + \sum_{t\in \gT} \frac{u_t + v_t}{\beta_t}.
  \end{align*}
  On $t \in \gT$, since $\beta_t$ is increasing with respect to $t$, we have
  \begin{align*}
      \beta_{t}^{\frac{3}{2}} - \beta_{t-1}^{\frac{3}{2}} &\geq \beta_{t}\sqrt{\beta_{t-1}} - \beta_{t-1}^{\frac{3}{2}} \geq \frac{\sqrt{z_t}}{h_t}+ \frac{u_t+v_t}{h_t\sqrt{\beta_{t-1}}} \geq \frac{\sqrt{z_t}}{h_t}, \\
      \beta_{t}^2 - \beta_{t-1}^2 &\geq \beta_{t}\beta_{t-1} - \beta_{t-1}^2 \geq \frac{2\sqrt{z_t \beta_{t-1}}}{h_t} + \frac{u_t+v_t}{h_t} \geq \frac{u_t+v_t}{h_t},
  \end{align*}
  which implies
  \begin{align*}
      \beta_{t+1}^{\frac{3}{2}} \geq \beta_1^{\frac{3}{2}} + \sum_{s \leq t, s \in \gT} \frac{\sqrt{z_s}}{h_s}, \text{ and } \beta_{t+1}^2 \geq \beta_1^2 + \sum_{s \leq t, s \in \gT} \frac{u_s+v_s}{h_s}.
  \end{align*}
  Therefore, by $\beta_{t+1} \leq 2^{\frac{1}{\alpha}}\beta_t$, we obtain for $\alpha \geq 2$ that
  \begin{align*}
    \sum_{t \in \gT} \sqrt{\frac{z_t}{\beta_t}}  &\leq 2^{\frac{1}{2\alpha}} \sum_{t \in \gT} \sqrt{\frac{z_t}{\beta_{t+1}}} \\
    &\leq 2^{\frac{1}{2\alpha}}  \sum_{t \in \gT} \frac{\sqrt{z_t}}{(\sum_{s \leq t, s\in \gT} \sqrt{z_s}/h_s)^{\frac{1}{3}}} \leq 2^{\frac{1}{4}}G_1(\{z_{s}\}_{s\in \gT}, \{h_{s}\}_{s\in \gT})
  \end{align*}
  and
  \begin{align*}
      \sum_{t \in \gT} \frac{u_t+v_t}{\beta_t} &\leq  2^{\frac{1}{\alpha}}\sum_{t \in \gT} \frac{u_t+v_t}{\beta_{t+1}}\\
      &\leq  2^{\frac{1}{\alpha}} \sum_{t\in \gT} \frac{u_t + v_t}{\sqrt{\sum_{s \leq t, s\in \gT}(u_s+v_s)/h_s}} \leq \sqrt{2}G_2(\{u_{s}+v_s\}_{s\in \gT},\{h_{s}\}_{s\in \gT}),
  \end{align*}
  which concludes the proof with the choice of $\beta_1$ in (\ref{def: beta1}).
\end{proof}

\section{Proof of Theorem~\ref{thm: adv}: Adversarial regime}
In this section, we prove the near-optimal guarantee in the adversarial regime.
We first introduce the previous results, which will be used also for the proof of Theorem~\ref{thm: sca}.
\begin{lemma}[Lemma 5 of \citet{tsuchiya2024simple} and Lemma 3 of \citet{ito2024adaptive}]\label{lem: G to bounds}
  For $G_1, G_2$ defined in (\ref{def: G function}) with positive sequences $a,b$ of size $n$ and any integer $J\geq 0$, it holds that
  \begin{align*}
      G_1(a_{1:n}, b_{1:n}) &\leq \frac{3}{2}\min\qty( \qty(\sqrt{2J}\sum_{t=1}^n \sqrt{a_tb_t})^{\frac{2}{3}} + \qty(\sqrt{2^{-J/2}a_{\max}b_{\max}})^{\frac{2}{3}}n^{\frac{2}{3}}, \qty(\sqrt{b_{\max}}\sum_{t=1}^n \sqrt{a_t})^{\frac{2}{3}} ), \\
      G_2(a_{1:n}, b_{1:n}) &\leq 2 \min\qty( \qty(2J\sum_{t=1}^n a_tb_t)^{\frac{1}{2}} + \qty(2^{-J}a_{\max}b_{\max})^{\frac{1}{2}}n^{\frac{1}{2}},  \qty(b_{\max}\sum_{t=1}^n a_t)^{\frac{1}{2}}).
  \end{align*}
\end{lemma}
\begin{proof}[Proof of Theorem~\ref{thm: adv}]
  By Lemma~\ref{lem: G to bounds}, we have
\begin{align*}
    G_1(\{z_{s}\}_{s\in \gT}, \{h_{s}\}_{s\in \gT}) \leq \frac{3}{2} \qty(\sqrt{h_{\max}}\sum_{t\in \gT}\sqrt{z_t})^{\frac{2}{3}}&\leq \frac{3}{2} \qty(\sqrt{h_{\max}}\sum_{t=1}^T \sqrt{z_t})^{\frac{2}{3}} \\
    &\leq  \frac{3}{2} \qty(\sqrt{h_{\max}z_{\max}}T)^{\frac{2}{3}} \\
    &= \frac{3}{2}(h_{\max}z_{\max})^{\frac{1}{3}} T^{\frac{2}{3}}
\end{align*}
and
\begin{align*}
    G_2(\{u_{s}+v_s\}_{s\in \gT},\{h_{s}\}_{s\in \gT}) &\leq 2\qty(h_{\max}\sum_{t\in \gT} (u_t+v_t))^{\frac{1}{2}} \\
    &\leq  2\qty(h_{\max}\sum_{t=1}^T (u_t+v_t))^{\frac{1}{2}} \\
    &\leq 2 (h_{\max}(u_{\max}+v_{\max}))^{\frac{1}{2}}T^{\frac{1}{2}}.
\end{align*}
Hence, by Lemmas~\ref{lem: decomposition},~\ref{lem: reg to G} and~\ref{lem: max parameters}, we obtain for $\alpha \geq 2$
\begin{align*}
    \Reg_T &\leq \gO\qty(\qty(\alpha^2 \delta_* K^{\frac{1}{\alpha}} \log K )^{\frac{1}{3}}T^{\frac{2}{3}}) + \gO\qty(\sqrt{\alpha K^{\frac{1}{\alpha}} \qty(\alpha K^{1-\frac{1}{\alpha}}+\alpha \delta_*)T}) \\
    &\hspace{3em}+ \gO(\alpha) + \gO\qty( \max\qty(\delta_* \log K, K^{1-\frac{1}{\alpha}} + \delta_*)\alpha K^{\frac{1}{\alpha}}).
\end{align*}
When $\alpha = 2\log K$ (so that $\alpha\geq 2$ for $K\geq 2$), we have
\begin{align*}
    \Reg_T \leq \gO\qty(\log K \delta_*^{1/3}T^{2/3}) + \gO(\log K\sqrt{(\delta_*+K)T}) + \gO(\log K) + \gO\qty(\log K \cdot \max\qty(\delta_*\log K, K)).
\end{align*}
Since $\delta_* \leq K$, if $T \geq K^3$, then
\begin{align*}
    \Reg_T \leq \gO(\log K \cdot \delta_*^{1/3}T^{2/3}),
\end{align*}
which concludes the proof.
\end{proof}

\section{Proof of Theorem~\ref{thm: sca}: Adversarial regime with self-bounding constraints}\label{app: sca}
In this section, we prove the near-optimal guarantee in the adversarial regime with self-bounding constraints ($\Delta, C, T)$, which, together with Theorem~\ref{thm: adv}, shows the BOBW guarantee of Algorithm~\ref{alg: ftpl}.

Define for any $\Delta \in \sR^K$ and $p \in \{ p \in [0,1]^K : \sum_{i=1}^K p_i = 1 \}$, 
\begin{align*}
    \Delta(p) = \inp{\Delta}{p}.
\end{align*}
By definition of $y_t = (1-\gamma_t)w_t + \gamma_t w^0$ and the choice of $\beta_1$ in (\ref{def: beta1}), which ensures $\gamma_t \leq 1/2$, we have
\begin{equation*}
  \Delta(w_t)\leq 2 \Delta(y_t), \, \forall t \in \sN.
\end{equation*}
Then, we first introduce an auxiliary lemma, whose proof is given in Appendix~\ref{app: G to w}.
\begin{lemma}\label{lem: G to w}
    For any $t \in \sN$ and $z_t, h_t$ defined in (\ref{def: zt}) and (\ref{def: ht}), we have
    \begin{align*}
        \sqrt{z_th_t}\leq  \frac{c_{zh}}{\Delta_{\min}} \Delta(w_t), \text{ and } (u_t+v_t)h_t \leq \frac{c_{uh}}{\Delta_{\min}} \Delta(w_t),
    \end{align*}
    where
    \begin{align*}
        c_{zh} &= \sqrt{4352 \alpha \delta_* K^{1/\alpha}}+ \frac{\sqrt{86\alpha^2 \delta_* K^{1/\alpha} \log K }}{0.19} = \Theta\qty(\sqrt{\alpha^2 \delta_* K^{1/\alpha} \log K}), \\
        c_{uh} &= 3200\alpha K + 192c_v \alpha \delta_* K^{\frac{1}{\alpha}} +\frac{8 \alpha K^{\frac{1}{\alpha}}(c_v(\alpha+1)\delta_* + 122\alpha K^{1-\frac{1}{\alpha}})}{0.19} = \Theta(\alpha K + \alpha \delta_* K^{\frac{1}{\alpha}}+ \delta_*\alpha^2 K^{1/\alpha}+\alpha^2 K).
    \end{align*}
\end{lemma}

\begin{proof}[Proof of Theorem~\ref{thm: sca}]
  By Lemma~\ref{lem: G to bounds} with 
  \begin{equation*}
    J_1= \max(1,\lceil 2\log_2 T \rceil), \qq{and} J_2 = \max(1, \lceil \log_2 T \rceil),
  \end{equation*}
  together with Lemma~\ref{lem: G to w}, we obtain
\begin{align*}
    G_1(\{z_{s}\}_{s\in \gT}, \{h_{s}\}_{s\in \gT}) &\leq \frac{3}{2^{2/3}} J_1^{1/3} \qty(\sum_{t=1}^T \sqrt{z_t h_t})^{2/3} + \frac{3}{2}2^{-J_1/3}(z_{\max}h_{\max})^{\frac{1}{3}}T^{\frac{2}{3}}  \\
    &\leq \frac{3c_{zh}^{2/3}J_{1}^{1/3}}{2^{2/3}\Delta_{\min}^{2/3}}\qty(\sum_{t=1}^T \Delta(w_t))^{2/3}  + \frac{3(z_{\max}h_{\max})^{\frac{1}{3}}}{2},
\end{align*}
and
\begin{align*}
    G_2(\{u_{s}+v_s\}_{s\in \gT},\{h_{s}\}_{s\in \gT}) &\leq 2\sqrt{2J_2 \sum_{t=1}^T (u_t+v_t)h_t} +\sqrt{2^{-J_2}(u_{\max}+v_{\max})h_{\max}T}\\
    &\leq  \frac{2\sqrt{2c_{uh}J_2}}{\sqrt{\Delta_{\min}}} \sqrt{\sum_{t=1}^T \Delta(w_t)}  + \sqrt{(u_{\max}+v_{\max})h_{\max}}.
\end{align*}
Here, using the relation between $\Delta(w_t)$ and $\Delta(y_t)$ and the self-bounding constraint as in (\ref{def: SCA}), we have
\begin{align}\label{eq: direct C}
    \E\qty[\sum_{t=1}^T \Delta(w_t)] \leq \E\qty[2\sum_{t=1}^T \Delta(y_t)] = \E\qty[2\sum_{t=1}^T \inp{\Delta}{e_{i_t}}] \leq 2(\Reg_T +C).
\end{align}
Combining this with the preceding bounds, the regret can be written in the form
\begin{align*}
  \Reg_T \leq a 2^{2/3} (\Reg_T+C)^{2/3} + b\sqrt{2}(\Reg_T + C)^{1/2} + c',
\end{align*}
where
\begin{equation*}
  a = 3 \qty(\frac{c_{zh}^{2}J_1}{(2\Delta_{\min})^{2}})^{\frac{1}{3}}, \hspace{2em} b= 2\sqrt{\frac{2c_{uh}J_2}{\Delta_{\min}}},
\end{equation*}
and $c'$ represents all the remaining $T$-independent terms.
Since it is equivalent to find an upper bound for $X=\Reg_T + C$ satisfying
\begin{equation*}
  X \leq a2^{2/3} X^{2/3} + b\sqrt{2} X^{1/2} + C + c',
\end{equation*}
solving this provides
\begin{equation}\label{eq: solving eq}
  X \leq 16a^3 + \frac{9}{2}b^2 + 3(C+c').
\end{equation}
For completeness, we provide the proof for this inequality in Appendix~\ref{app: solving eq}.
Therefore, we obtain
\begin{align*}
    \Reg_T \leq \gO\qty(a^3 +b^2 + C + c') &= \gO\qty(\frac{c_{zh}^2 \log T}{\Delta_{\min}^2} + \frac{c_{uh}\log T}{\Delta_{\min}}+C + c') \\
    &=\gO\qty(\frac{\alpha^2 \delta_* K^{1/\alpha} \log K \log T}{\Delta_{\min}^2} + \frac{(\alpha K + \alpha \delta_* K^{\frac{1}{\alpha}}+ \alpha^2 \delta_* K^{\frac{1}{\alpha}} + \alpha^2 K)\log T}{\Delta_{\min}}+C + c').
\end{align*}
Here, similarly to the adversarial regime, it holds that
\begin{align*}
  c' &\leq \gO\qty(\max(\delta_* \log K, K^{1-\frac{1}{\alpha}}+\delta_*)\alpha K^{\frac{1}{\alpha}}) + \gO\qty((z_{\max}h_{\max})^{\frac{1}{3}}+\sqrt{(u_{\max}+v_{\max})h_{\max}}) \\
  &=\gO\qty(\max(\delta_* \log K, K^{1-\frac{1}{\alpha}}+\delta_*)\alpha K^{\frac{1}{\alpha}}) + \gO\qty(\qty(\alpha^2 \delta_* K^{\frac{1}{\alpha}} \log K )^{\frac{1}{3}}+\alpha\sqrt{(K^{1-\frac{1}{\alpha}}+\delta_*)K^{\frac{1}{\alpha}}}) \\
  &=\gO\qty(\max(\delta_* \log K, K^{1-\frac{1}{\alpha}}+\delta_*)\alpha K^{\frac{1}{\alpha}}).
\end{align*}
Therefore, if $\alpha = \Theta(\log K)$, we have
\begin{align*}
   \Reg_T &\leq \gO\qty(\frac{\delta_* \log^2 K  \log T}{\Delta_{\min}^2} + \frac{(K\log K + \delta_* \log K+\delta_* \log^2 K + K\log^2 K)\log T}{\Delta_{\min}}+C + \log K \cdot \max\qty(\delta_*\log K, K)) \\
   &=  \gO\qty(\frac{\delta_* \log^2 K  \log T}{\Delta_{\min}^2} + \frac{K\log^2 K\log T}{\Delta_{\min}}+C + \log K \cdot \max\qty(\delta_*\log K, K)), \tag{$\because \delta_* \leq K$}
\end{align*}
which concludes the proof.
\end{proof}

\section{Improved regret bounds for adversarial regime with self-bounding constraints}\label{app: MS type}
In this section, we show that our proposed method can also achieve a refined regret bound in \citet{masoudian2021improved} as done in \citet{tsuchiya2024simple}. 
This proof is a slight modification of the proof of Theorem~\ref{thm: sca}, along the lines of Appendix~C of \citet{tsuchiya2024simple}.
\begin{theorem}
  In adversarial regime with self-bounding constraints $(\Delta, C, T)$, Algorithm~\ref{alg: ftpl} with $\alpha \geq 2$ satisfies
  \begin{align*}
    \Reg_T &\leq \gO\qty(\frac{\rho}{\Delta_{\min}^2}\log (\frac{\Delta_{\min}^2 T}{\rho (1+ C\Delta_{\min})}) + \qty(\frac{C^2 \rho}{\Delta_{\min}^2}\log (\frac{\Delta_{\min}^2 T}{\rho (1+ C\Delta_{\min})}))^{\frac{1}{3}}) + \gO(c'')
  \end{align*}
  if $T \geq \tau = (\rho/\Delta_{\min}^2 + C\rho/\Delta_{\min})$, where $\rho=\Theta(\max(c_{zh}^2, c_{uh}\Delta_{\min}))$ for $c_{zh}, c_{uh}$ defined in Lemma~\ref{lem: G to w} and $\gO(c'')=\gO(\max(\delta_* \log K, K^{1-\frac{1}{\alpha}}+\delta_*)\alpha K^{\frac{1}{\alpha}})$.
  If $T \leq \tau$, then 
  \begin{equation*}
    \Reg_T\leq \gO\qty((\alpha^2 \delta_* K^{\frac{1}{\alpha}}\log K)^{\frac{1}{3}}\tau^{2/3} + \sqrt{\alpha K^{\frac{1}{\alpha}}(\alpha K^{1-\frac{1}{\alpha}}+\alpha \delta_*)\tau}+ c'').
  \end{equation*}
\end{theorem}
\begin{proof}
  The refined result can be obtained by modifying the choice of $J_1$ and $J_2$ and the use of (\ref{eq: direct C}).
  First, using Lemma~\ref{lem: G to bounds} with
  \begin{equation*}
    J_{1,\epsilon}=  \lceil 2\log_2 (\epsilon T)\rceil, \qq{and} J_{2,\epsilon} = \lceil \log_2 (\epsilon T) \rceil,
  \end{equation*}
  for any $\epsilon \geq 1/T$, which gives
  \begin{align*}
     G_1(\{z_{s}\}_{s\in \gT}, \{h_{s}\}_{s\in \gT}) &\leq \frac{3c_{zh}^{2/3}J_{1,\epsilon}^{1/3}}{2^{2/3}\Delta_{\min}^{2/3}}\qty(\sum_{t=1}^T \Delta(w_t))^{2/3}  + \frac{3}{2} \qty(\frac{\sqrt{z_{\max}h_{\max}}}{\epsilon})^{\frac{2}{3}}, \\
     G_2(\{u_{s}+v_s\}_{s\in \gT},\{h_{s}\}_{s\in \gT}) &\leq \frac{2\sqrt{2c_{uh}J_{2,\epsilon}}}{\sqrt{\Delta_{\min}}} \sqrt{\sum_{t=1}^T \Delta(w_t)}  + \sqrt{\frac{(u_{\max}+v_{\max})h_{\max}}{\epsilon}}.
  \end{align*}
  Let
  \begin{equation*}
    R_w = \sum_{t=1}^T \Delta(w_t).
  \end{equation*}
  Then, by (\ref{eq: direct C}), for any $\kappa >0$, it holds that
  \begin{align*}
    \Reg_T &\leq (1+\kappa) \Reg_T - \kappa\qty( \frac{R_w}{2} - C) \\
    &\leq (1+\kappa) \qty(a R_w^{2/3}+ b \sqrt{R_w} + c') - \kappa\qty( \frac{R_w}{2} - C) \\
    &=  \qty((1+\kappa)a R_w^{2/3} - \frac{\kappa}{4}R_w) + \qty((1+\kappa)b \sqrt{R_w} - \frac{\kappa}{4}R_w) + \kappa C + (1+\kappa)c',
  \end{align*}
  where 
  \begin{equation*}
    a = 3 \qty(\frac{c_{zh}^{2}J_{1,\epsilon}}{(2\Delta_{\min})^{2}})^{\frac{1}{3}}, \hspace{2em} b= 2\sqrt{\frac{2c_{uh}J_{2,\epsilon}}{\Delta_{\min}}},
  \end{equation*}
  and $c'$ represents all the remaining $T$-independent terms.
  Since $Ax^2-Bx^3 \leq 4A^3/(27B^2)$ and $Ax-Bx^2 \leq A^2/(4B)$ hold for $A \geq 0, B>0$ and $x\geq 0$, we have
  \begin{align*}
    \Reg_T \leq \frac{64(1+\kappa)^3}{27\kappa^2} a^3 + \frac{(1+\kappa)^2}{\kappa} b^2 + \kappa C + (1+\kappa)c'.
  \end{align*}
  Take
  \begin{equation*}
    \kappa = \min\qty(1, \max\qty(\frac{a}{C^{\frac{1}{3}}} , \frac{b}{\sqrt{C}} )) \leq 1.
  \end{equation*}
  If $\kappa =1$, where either $a^3 \geq C$ or $b^2 \geq C$ is true, we have $C \leq a^3+b^2$ and thus
  \begin{equation*}
    \Reg_T \leq \gO\qty(a^3 + b^2 + c').
  \end{equation*}
  If $\kappa < 1$, where both $a^3 < C$ and $b^2 < C$ are true, we have 
  \begin{align*}
    \Reg_T &\leq \gO\qty(\qty(a^3C^2)^{1/3} + b\sqrt{C} + \kappa C + c') \\
    &\leq\gO\qty(\qty(a^3C^2)^{1/3} + b\sqrt{C} + C (\frac{a}{C^{1/3}}+\frac{b}{\sqrt{C}}) + c') \\
    &= \gO\qty(\qty(a^3C^2)^{1/3} + b\sqrt{C}+ c').
  \end{align*}
  In sum, we obtain
  \begin{align*}
    \Reg_T &\leq \gO\qty(a^3 + b^2+ \qty(a^3C^2)^{1/3} + b\sqrt{C}+ c') \\
    &\leq \gO\qty(\frac{c_{zh}^2}{\Delta_{\min}^2}\log (\epsilon T) + \frac{c_{uh}}{\Delta_{\min}}\log (\epsilon T) + \qty(\frac{C^2 c_{zh}^2}{\Delta_{\min}^2}\log (\epsilon T))^{\frac{1}{3}}+ \sqrt{\frac{C c_{uh}}{\Delta_{\min}}\log (\epsilon T)}+ c').
  \end{align*}
  Then, define $\rho=\max(c_{zh}^2, c_{uh}\Delta_{\min})$, so that
  \begin{equation*}
    \frac{c_{zh}^2}{\Delta_{\min}^2}\log (\epsilon T) + \frac{c_{uh}}{\Delta_{\min}}\log (\epsilon T) \leq \frac{2\rho}{\Delta_{\min}^2} \log (\epsilon T)
  \end{equation*}
  and thus
    \begin{align*}
    \Reg_T &\leq \gO\qty(\frac{\rho}{\Delta_{\min}^2}\log (\epsilon T) + \qty(\frac{C^2 \rho}{\Delta_{\min}^2}\log (\epsilon T))^{\frac{1}{3}} + c')\\ 
    &= \gO\qty(\frac{\rho}{\Delta_{\min}^2}\log (\epsilon T) + \qty(\frac{C^2 \rho}{\Delta_{\min}^2}\log (\epsilon T))^{\frac{1}{3}} + \qty(\frac{\sqrt{z_{\max}h_{\max}}}{\epsilon})^{\frac{2}{3}} + \sqrt{\frac{(u_{\max}+v_{\max})h_{\max}}{\epsilon}} + c'').
  \end{align*}
  Here, we use
  \begin{align*}
    c' &\leq \gO\qty(\max(\delta_* \log K, K^{1-\frac{1}{\alpha}}+\delta_*)\alpha K^{\frac{1}{\alpha}}) + \gO\qty((z_{\max}h_{\max}/\epsilon^2)^{\frac{1}{3}}+\sqrt{(u_{\max}+v_{\max})h_{\max}/\epsilon}),
  \end{align*}
  with $c''= \gO\qty(\max(\delta_* \log K, K^{1-\frac{1}{\alpha}}+\delta_*)\alpha K^{\frac{1}{\alpha}})$.
  Then, for $T\geq \tau = \rho/\Delta_{\min}^2 + C\rho/\Delta_{\min}$, setting
  \begin{equation*}
    \epsilon = 1/\tau  = \frac{1}{\rho/\Delta_{\min}^2 + C\rho/\Delta_{\min}} \geq \frac{1}{T},
  \end{equation*}
  which gives
  \begin{align*}
    \Reg_T &\leq \gO\qty(\frac{\rho}{\Delta_{\min}^2}\log (\frac{\Delta_{\min}^2 T}{\rho (1+ C\Delta_{\min})}) + \qty(\frac{C^2 \rho}{\Delta_{\min}^2}\log (\frac{\Delta_{\min}^2 T}{\rho (1+ C\Delta_{\min})}))^{\frac{1}{3}}) \\
    &\quad + \gO\qty( \qty((z_{\max}h_{\max})^{\frac{1}{3}} + \sqrt{(u_{\max}+v_{\max})h_{\max}})\tau^{2/3} + c'').
  \end{align*}
  By applying Lemma~\ref{lem: max parameters} and following the same steps as in Appendix~\ref{app: sca} concludes the proof.
  When $T<\tau$, we have $J_{1,\epsilon}=J_{2,\epsilon}=0$, so the leading term becomes $c'$. 
  Since the dependence on $\tau$ is absorbed into $\epsilon$, the regret bound is of order $\tau^{2/3}$ in such cases.
\end{proof}

\section{Proof of auxiliary lemmas}\label{app: auxiliary}
In this section, we provide proofs of the auxiliary lemmas used to establish the main results.
Let $\lambda \in [0,\infty)^K$ be a non-negative vector and let $\sigma_i(\lambda)$ denote the rank of $\lambda$ in non-decreasing order with ties broken according to a fixed arbitrary strict order $\prec$ on $V$, i.e.,
\begin{equation*}
  \sigma_{i}(\lambda) = 1 + |\{j \in V: \lambda_j < \lambda_i \}| \\ +  |\{j \prec i: \lambda_j = \lambda_i\}|.
\end{equation*}
Note that $\sigma_{t,i}=\sigma_i(\hat{L}_t)$.
Then, let $j^*(\lambda)$ denote the unique arm satisfying $\sigma_{j^*}(\lambda) = 1$.
For each $i \ne j^*(\lambda)$, define 
\begin{equation*}
  p_i(\lambda;\alpha) := \min\qty(\frac{1}{(1+\lambda_i)^\alpha}, \frac{1}{\sigma_i(\lambda)}).
\end{equation*}
From now on, when a general $\lambda \in \sR^K$ (instead of $\hat{L}_t$) and $\alpha\geq 2$ are considered, we omit the dependence on $\lambda$ and $\alpha$ from the notation whenever it is clear from the context.

\subsection{Proof of Lemma~\ref{lem: stability all}}
We first introduce two key auxiliary lemmas to prove Lemma~\ref{lem: stability all}, which decomposes the regret terms for $j_t$.
\begin{lemma}[$V_1$ leader decomposition]\label{lem: V1 leader}
    If $j_t \in V_1$, then
    \begin{equation*}
        \E\qty[\hat{\ell}_{t,j_t} \qty(\phi_{j_t}(\eta_t \hat{L}_t) - \phi_{j_t}(\eta_t \hat{L}_{t+1})) \middle | \hat{L}_t] \leq \frac{3\alpha \delta_*}{\beta_t \gamma_t} \sum_{i \ne j_t} p_{t,i}^{1+1/\alpha}.
    \end{equation*}
\end{lemma}
\begin{lemma}[$V_2$ leader decomposition]\label{lem: V2 leader}
    If $j_t \in V_2$, then
    \begin{equation*}
        \E\qty[\hat{\ell}_{t,j_t} \qty(\phi_{j_t}(\eta_t \hat{L}_t) - \phi_{j_t}(\eta_t \hat{L}_{t+1})) \middle | \hat{L}_t] \leq \frac{24\alpha}{\beta_t} \sum_{i \ne j_t} p_{t,i}^{1/\alpha}.
    \end{equation*}
\end{lemma}
\begin{proof}[Proof of Lemma~\ref{lem: stability all}]
  Since $\phi_i(\lambda)$ is increasing w.r.t. $\lambda_j$ for $i \ne j$, we have
  \begin{align*}
      \E\qty[\hat{\ell}_{t,i} \qty(\phi_i(\eta_t \hat{L}_t) - \phi_i(\eta_t \hat{L}_{t+1})) \middle | \hat{L}_t] &\leq \E\qty[\hat{\ell}_{t,i} \qty(\phi_i(\eta_t \hat{L}_t) - \phi_i(\eta_t (\hat{L}_{t}+\hat{\ell}_{t,i}e_i))) \middle | \hat{L}_t]\\
      &\leq\E\qty[ -\eta_t \hat{\ell}_{t,i}^2 \phi_i'(\eta_t\hat{L}_t) \middle | \hat{L}_t] \tag{by Appendix C.3.2.~of \citet{lee2024follow}} \\
      &= \E\qty[ -\eta_t \hat{\ell}_{t,i}^2 w_{t,i} \frac{\phi_i'(\eta_t\hat{L}_t)}{w_{t,i}} \middle | \hat{L}_t]  \\
      &\leq \E\qty[-\eta_t \hat{\ell}_{t,i}^2 w_{t,i} e\alpha  p_{t,i}^{\frac{1}{\alpha}} \middle | \hat{L}_t] \tag{by Lemma 7 of \citet{pmlr-v336-lee26a}} \\
      &\leq \begin{dcases}
        \frac{e\alpha}{\beta_t} \frac{\delta_*}{\gamma_t} w_{t,i} p_{t,i}^{\frac{1}{\alpha}} ,& \, i \in V_1, \\
        \frac{4e\alpha}{\beta_t} p_{t,i}^{\frac{1}{\alpha}}, &\, i \in V_2.
      \end{dcases} \tag{by Lemma~\ref{lem: estimator}}
  \end{align*}
  Since $w_{t,i}\leq p_{t,i}$ holds by definition (or see, \citet[Appendix C.1]{pmlr-v336-lee26a}), we obtain
  \begin{align*}
      \E\qty[\hat{\ell}_{t,i} \qty(\phi_i(\eta_t \hat{L}_t) - \phi_i(\eta_t \hat{L}_{t+1})) \middle | \hat{L}_t] &\leq \begin{dcases}
         \frac{ e\alpha  \delta_*  }{\beta_t \gamma_t} p_{t,i}^{1+1/\alpha} \, &\text{if } i \in V_1 \\
          \frac{4e\alpha }{\beta_t} p_{t,i}^{1/\alpha}, \, &\text{if } i \in V_2.
      \end{dcases}
  \end{align*}
  Hence, by Lemmas~\ref{lem: p and q},~\ref{lem: V1 leader}, and~\ref{lem: V2 leader}, we obtain for $\alpha\geq 2$
  \begin{align*}
    \E\qty[\inp{\hat{\ell}_t}{\phi(\eta_t \hat{L_t})-\phi(\eta_t \hat{L}_{t+1})}\middle| \hat{L}_t] &= \sum_{i=1}^K \E\qty[ \hat{\ell}_{t,i} \qty(\phi_i(\eta_t \hat{L}_t) - \phi_i(\eta_t \hat{L}_{t+1})) \middle| \hat{L}_t] \\
    &\leq \sum_{i \in V_1 \setminus\{j_t\}} \frac{2e\sqrt{2}\alpha \delta_*}{\beta_t \gamma_t} q_{t,i}^{1+\frac{1}{\alpha}} + \sum_{i \in V_2 \setminus \{j_t\}} \frac{4e\sqrt{2}\alpha}{\beta_t} q_{t,i}^{\frac{1}{\alpha}} \\
    &\hspace{1em}+ \I[j_t \in V_1] \frac{3\sqrt{2}\alpha \delta_*}{\beta_t \gamma_t} \sum_{i \ne j_t} q_{t,i}^{1+\frac{1}{\alpha}} + \I[j_t \in V_2] \frac{24\sqrt{2}\alpha}{\beta_t} \sum_{i \ne j_t} q_{t,i}^{\frac{1}{\alpha}}.
  \end{align*}
  Therefore, the definitions of $z_t, u_t$ in (\ref{def: zt}--\ref{def: ut}) conclude the proof.
\end{proof}

\subsection{Proof of Lemma~\ref{lem: max parameters}}
\begin{proof}
  By definition, $q_{t,i}\leq 1/\sigma_{t,i}$ holds for any $t \in \sN, i\in V$.
  Hence, for any $t \in \sN$ and $K\geq 2$, we have
  \begin{align*}
    z_{t} \leq \alpha \delta_*\qty(\sum_{i \ne j_t}(2e\sqrt{2}+3\sqrt{2})\sigma_{t,i}^{-1-\frac{1}{\alpha}}) = (2e\sqrt{2}+3\sqrt{2})\alpha \delta_* \sum_{i =2}^{K}i^{-1-\frac{1}{\alpha}} \leq (2e\sqrt{2}+3\sqrt{2})\alpha^2 (1-K^{-1/\alpha})\delta_*.
  \end{align*}
  For the first equality, recall the definition of $j_t$ where $\sigma_{t,j_t}=1$.
  Here, since $1-e^{-x} \leq x$, we have
  \begin{align*}
    \alpha (1-K^{-1/\alpha}) = \alpha\qty(1- e^{-\frac{\log K}{\alpha}}) \leq \log K,
  \end{align*}
  which implies
  \begin{equation*}
    z_{t} \leq (2e\sqrt{2}+3\sqrt{2})\alpha \delta_* \log K, \, \forall t \in \sN.
  \end{equation*}
  Similarly, we obtain
  \begin{align*}
    u_{t} &\leq \alpha \qty(\sum_{i \ne j_t}(4e\sqrt{2}+24\sqrt{2})\sigma_{t,i}^{-\frac{1}{\alpha}}) = (4e\sqrt{2}+24\sqrt{2})\alpha \sum_{i=2}^{K} i^{-\frac{1}{\alpha}} \leq (4e\sqrt{2}+24\sqrt{2})\frac{\alpha^2}{\alpha-1}K^{1-\frac{1}{\alpha}} \\
    v_t &\leq c_v (\alpha+1) \delta_* \tag{$\because \bar{s}_t \leq 1$} \\
    h_t & \leq \frac{4\alpha}{\alpha-1}\sum_{i\ne j_t} \sigma_{t,i}^{-1+\frac{1}{\alpha}} = \frac{4\alpha}{\alpha-1}\sum_{i =2}^K i^{-1+\frac{1}{\alpha}} \leq \frac{4\alpha^2}{\alpha-1} K^{\frac{1}{\alpha}},
  \end{align*}
  which concludes the proof.
\end{proof}

\subsection{Proof of Lemma~\ref{lem: G to w}}\label{app: G to w}
Following the previous BOBW analyses for FTPL~\citep{kim2025follow, pmlr-v336-lee26a}, we consider the event
\begin{equation}\label{def: Dt}
    \gD_{t,\alpha} = \qty{\sum_{i\ne j_{t}}  \frac{1}{(2^{1/\alpha}+\eta_t \hat{\uL}_{t,i})^{\alpha}} \leq \frac{1}{2}}.
\end{equation}
Then, we introduce the previous results that are required to prove Lemma~\ref{lem: G to w}.
\begin{lemma}[Lemma 14 of \citet{pmlr-v336-lee26a} and Lemma 11 of \citet{kim2025follow}]\label{lem: w bounds on Dt}
  For any $\alpha>1$, on the event $\gD_{t,\alpha}$ defined in (\ref{def: Dt}), we have
  \begin{equation*}
      w_{t,j_t} \geq \frac{1}{4}, \text{ and } \forall i \ne j_t, \, w_{t,i} \geq \frac{1}{8(1+\eta_t\hat{\uL}_{t,i})^\alpha}.
  \end{equation*}
  Consequently with Lemma~\ref{lem: p and q}, $q_{t,i} \leq 2p_{t,i} \leq 16w_{t,i}$ for all $i \ne j_t$. 
  On the complementary event $\gD_{t,\alpha}^c$, it holds that
  \begin{align*}
      w_{t,j_t} \leq \frac{1+e^{-1/2}}{2} \implies \Delta(w_t) = \inp{\Delta}{w_t} \geq \frac{1-e^{-1/2}}{2}\Delta_{\min},
  \end{align*}
  where $\Delta_{\min} = \min_{i \ne i^*}\Delta_i$ for the self-bounding constraint $\Delta$.
\end{lemma}
\begin{proof}[Proof of Lemma~\ref{lem: G to w}]
  First, suppose the event $\gD_{t,\alpha}$ occurs, where we have $q_{t,i} \leq 16 w_{t,i}$ for $i\ne j_t$ by Lemma~\ref{lem: w bounds on Dt}.
  Hence, for $\alpha \geq 2$ we have 
  \begin{align*}
      \I[\gD_{t,\alpha}]z_t h_t &\leq 8.5 \delta_* \frac{\alpha^2}{\alpha-1}\qty(\sum_{i\ne j_{t}} q_{t,i}^{1+1/\alpha}) \qty(\sum_{i\ne j_{t}} q_{t,i}^{1-1/\alpha}) \\
      &\leq 4352 \delta_* \alpha\qty(\sum_{i\ne j_{t}} w_{t,i}^{1+1/\alpha}) \qty(\sum_{i\ne j_{t}} w_{t,i}^{1-1/\alpha})
  \end{align*}
  Since $w_{t,j_t}\geq w_{t,i}$ for any $i \in V$ and $t\in \sN$, removing
  $j_t$ term yields no larger summation than removing $i^*$ term.
  Therefore,
  \begin{align*}
      \sum_{i\ne j_{t}} w_{t,i}^{1+1/\alpha} \leq \sum_{i\ne i^*}w_{t,i}^{1+1/\alpha} &\leq \qty(\sum_{i\ne i^*}w_{t,i})^{1+1/\alpha}\\ 
      &= \Delta_{\min}^{-1-1/\alpha}\qty(\sum_{i\ne i^*}\Delta_{\min}w_{t,i})^{1+1/\alpha}  \leq \Delta_{\min}^{-1-1/\alpha}\qty(\sum_{i\ne i^*}\Delta_{i}w_{t,i})^{1+1/\alpha} \numberthis{\label{eq: 1+alpha w}}
  \end{align*}
  and by H\"older's inequality
  \begin{align*}
      \sum_{i\ne j_{t}} w_{t,i}^{1-1/\alpha} \leq \sum_{i\ne i^*}w_{t,i}^{1-1/\alpha} &\leq \qty(\sum_{i\ne i^*}\Delta_{i}w_{t,i})^{1-1/\alpha}\qty(\sum_{i\ne i^*} \Delta_i^{-\alpha+1})^{1/\alpha} \\
      &\leq (K \Delta_{\min}^{-\alpha+1})^{1/\alpha} \qty(\sum_{i\ne i^*}\Delta_{i}w_{t,i})^{1-1/\alpha} \numberthis{\label{eq: 1-alpha w}}
  \end{align*}
  Recall that $\Delta(w_t) = \inp{\Delta}{w_t} =\sum_{i\ne i^*}\Delta_i w_{t,i}$, which implies
  \begin{align*}
      \I[\gD_{t,\alpha}]z_t h_t \leq 4352\alpha \frac{\delta_* K^{1/\alpha}}{\Delta_{\min}^2} (\Delta(w_t))^2 \implies \I[\gD_{t,\alpha}] \sqrt{z_t h_t} \leq \frac{\sqrt{4352  \alpha \delta_* K^{1/\alpha}}}{\Delta_{\min}} \Delta(w_t).
  \end{align*}
  On the other hand, by $q_{t,i}\leq 16w_{t,i}$ again, for $\alpha \geq 2$, we have 
  \begin{align*}
      \I[\gD_{t,\alpha}] u_t h_t &\leq 200  \frac{\alpha^2}{\alpha-1} \qty(\sum_{i\ne j_{t}}q_{t,i}^{1/\alpha}) \qty(\sum_{i\ne j_{t}}q_{t,i}^{1-1/\alpha}) \\
      &\leq 3200  \alpha \qty(\sum_{i\ne j_{t}}w_{t,i}^{1/\alpha}) \qty(\sum_{i\ne j_{t}}w_{t,i}^{1-1/\alpha})
  \end{align*}
  Here, again by H\"older's inequality, we have
  \begin{align*}
      \sum_{i\ne j_{t}} w_{t,i}^{1/\alpha} \leq \sum_{i\ne i^*}w_{t,i}^{1/\alpha} &\leq \qty(\sum_{i\ne i^*}\Delta_{i}w_{t,i})^{1/\alpha}\qty(\sum_{i\ne i^*} \Delta_i^{-1/(\alpha-1)})^{1-1/\alpha} \\
      &\leq (K \Delta_{\min}^{-1/(\alpha-1)})^{1-1/\alpha} \qty(\sum_{i\ne i^*}\Delta_{i}w_{t,i})^{1/\alpha} 
      \numberthis{\label{eq: 1/alpha w}}
  \end{align*}
  Combining this with (\ref{eq: 1-alpha w}) gives
  \begin{align*}
      \I[\gD_{t,\alpha}] u_t h_t &\leq  3200 \alpha \frac{K}{\Delta_{\min}}\Delta(w_t).
  \end{align*}
  It remains to control the contribution of $v_t$. 
  Since $\eta_{t-1}\geq \eta_t$, Lemma~\ref{lem: w bounds on Dt} implies that, on $\gD_{t,\alpha}$,
  \begin{align*}
      (1+\eta_{t-1} \hat{\uL}_{t,i})^{-\alpha} \leq (1+\eta_t\hat{\uL}_{t,i})^{-\alpha} \leq 8w_{t,i},
  \end{align*}
  which implies $\bar{s}_t \leq 8 \sum_{i\ne j_t}w_{t,i} \leq 8\sum_{i\ne i^*}w_{t,i}$ for $\bar{s}_t$ defined in (\ref{def: barst}).
  Furthermore, using $q_{t,i}\leq 16w_{t,i}$ and H\"older's inequality, we obtain
  \begin{align*}
      \sum_{i\ne j_t} q_{t,i}^{1-1/\alpha} \leq 16^{1-1/\alpha} \sum_{i\ne j_t} w_{t,i}^{1-1/\alpha} &\leq 16^{1-1/\alpha} K^{1/\alpha} \qty(\sum_{i\ne j_t}w_{t,i})^{1-1/\alpha}
  \end{align*}
  Therefore, since $(\alpha+1)/(\alpha-1)\leq 3$ for $\alpha \geq 2$,
  \begin{align*}
        \I[\gD_{t,\alpha}] v_t h_t &\leq c_v(\alpha+1)\delta_* \qty(8\sum_{i\ne j_t}w_{t,i})^{1/\alpha} \cdot 16^{1-1/\alpha} \frac{4\alpha}{\alpha-1} K^{1/\alpha} \qty(\sum_{i\ne j_t}w_{t,i})^{1-1/\alpha} \\
        &\leq 192 c_v  \alpha \delta_*  K^{1/\alpha}  \sum_{i\ne j_t} w_{t,i} \leq 192 c_v  \alpha d  K^{1/\alpha} \sum_{i\ne i^*} w_{t,i} \leq \frac{192 c_v  \alpha \delta_* K^{1/\alpha} }{\Delta_{\min}}\Delta(w_t),
  \end{align*}
  which proves the desired bounds on the event $\gD_{t,\alpha}$.

  On the complementary event $\gD_{t,\alpha}^c$, we have $\Delta(w_t) \geq (1-e^{-1/2})\Delta_{\min}/2 \geq 0.19 \Delta_{\min}$ by Lemma~\ref{lem: w bounds on Dt}.
  Therefore, any uniform upper bound on the corresponding quantity can be converted into a bound proportional to $\Delta(w_t)$, which implies for $\alpha \geq 2$
  \begin{align*}
      \I[\gD_{t,\alpha}^c] \sqrt{z_t h_t} &\leq \sqrt{z_{\max} h_{\max}} \frac{\Delta(w_t)}{0.19\Delta_{\min}} \leq \frac{\alpha\sqrt{86 \delta_* K^{1/\alpha} \log K }}{0.19\Delta_{\min}} \Delta(w_t), \\
      \I[\gD_{t,\alpha}^c] (u_t+v_t) h_t &\leq (u_{\max}+v_{\max})h_{\max} \frac{\Delta(w_t)}{0.19\Delta_{\min}} \leq \frac{8\alpha K^{1/\alpha}(c_v(\alpha+1)\delta_* + 122\alpha K^{1-1/\alpha})}{0.19\Delta_{\min}}\Delta(w_t).
  \end{align*}
  For the last inequality for each line, we use Lemma~\ref{lem: max parameters}.
\end{proof}

\subsection{Proof of Lemma~\ref{lem: V1 leader}}\label{app: V1 leader}
For the proof, we need a following result, whose proof is given in Appendix~\ref{app: jt stab sub 1}.
\begin{lemma}\label{lem: jt stab sub lemma}
    If $x \in \qty[0, \frac{1+\ul_i}{1.5(\alpha+1)}]$ for all $i \ne j^*(\lambda)$, then
    \begin{equation}\label{eq: lem V1 leader}
        \phi_{j^*(\lambda)}(\lambda) - \phi_{j^*(\lambda)}(\lambda + xe_{j^*(\lambda)})\leq 3\alpha x \sum_{i \ne j^*} p_i(\ul;\alpha)^{1+\frac{1}{\alpha}}.
    \end{equation}
    Moreover, for any $x\geq 0$,
    \begin{equation}\label{eq: lem V2 leader}
        \phi_{j^*(\lambda)}(\lambda) - \phi_{j^*(\lambda)}(\lambda + xe_{j^*(\lambda)})\leq \alpha x \cdot \min\qty(4 \sum_{i \ne j^*} p_i(\ul;\alpha)^{\frac{1}{\alpha}}, \phi_{j^*(\lambda)}(\lambda)).
    \end{equation}
\end{lemma}
\begin{remark}
  In Lemma~\ref{lem: jt stab sub lemma}, we consider the range of $x \leq \frac{1+\ul_i}{1.5(\alpha+1)}$.
  Here, the choice of multiplicative constant $1.5$ is solely for analytical reason, to introduce upper bounds with simple multiplicative factor $3$ in (\ref{eq: lem V1 leader}).
  If we modify the choice $1.5$ to $1$, corresponding factor becomes $4$ rather than $3$.
  Moreover, our proof for (\ref{eq: lem V2 leader}) also depends on the choice of $1.5$, where factor $4$ in (\ref{eq: lem V2 leader}) becomes $3$ if we reduce it to $1$.  
\end{remark}
\begin{proof}[Proof of Lemma~\ref{lem: V1 leader}]
    Let $j= j_t \in V_1$ for notational simplicity.
    Since $\phi_i(\lambda)$ is increasing w.r.t.~$\lambda_k$ for $i \ne k$, we have
    \begin{align*}
       \E\qty[\hat{\ell}_{t,j} \qty(\phi_{j}(\eta_t \hat{L}_t) - \phi_{j}(\eta_t \hat{L}_{t+1})) \middle | \hat{L}_t] &\leq \E\qty[\hat{\ell}_{t,j} \qty(\phi_j(\eta_t \hat{L}_t) - \phi_j(\eta_t (\hat{L}_{t}+\hat{\ell}_{t,j}e_j))) \middle | \hat{L}_t].
    \end{align*}
    Since $j \in V_1$, by definition in (\ref{def: loss estimator}), $\hat{\ell}_{t,j} = \frac{\I[B_t=1]O_{t,j}\ell_{t,j}}{\gamma_t \rho_j}$ and 
    \begin{equation*}
        \gamma_t \geq \frac{v_t}{\beta_t} = \frac{c_v(\alpha+1)\delta_* \bar{s}_t^{1/\alpha}}{\beta_t}\I[j_t \in V_1].
    \end{equation*}
    Therefore, it holds that 
    \begin{align*}
        \eta_{t}\hat{\ell}_{t,j} \leq \frac{\ell_{t,j}}{\gamma_t \rho_j} \leq\frac{\delta_*}{\beta_t \gamma_t \cdot \sum_{i \in \inb(j)}x_i^*} \leq \frac{\delta_*}{\beta_t \gamma_t} \leq \frac{\bar{s}_t^{-1/\alpha}}{c_v (\alpha+1)},
    \end{align*}
    where the third inequality follows from the definition of linear program (\ref{def: linear program}) with $j \in V_1$.
    Here, define 
    \begin{equation}\label{def: st}
      s_t := \min\qty(1, \sum_{i\ne j_t}\frac{1}{(1+\eta_t\hat{\uL}_{t,i})^\alpha}).
    \end{equation}
    Since $\beta_{t-1}\leq \beta_t \leq 2^{\frac{1}{\alpha}}\beta_{t-1}$ for any $i\in V$, we have
    \begin{equation*}
      1 + \eta_t \hat{\uL}_{t,i} = 1 + \frac{\beta_{t-1}}{\beta_t}\eta_{t-1} \hat{\uL}_{t,i} \geq \frac{\beta_{t-1}}{\beta_t} (1+\eta_{t-1}\hat{\uL}_{t,i}),
    \end{equation*}
    which implies
    \begin{equation*}
      (1+\eta_t \hat{\uL}_{t,i})^{-\alpha} \leq 2 (1+\eta_{t-1}\hat{\uL}_{t,i})^{-\alpha} \implies \bar{s}_t \leq s_t \leq 2\bar{s}_t.
    \end{equation*}
    By definition of $s_t$, we obtain 
    \begin{align*}
        s_t^{-1/\alpha} = \frac{1}{\min\qty(1, \sum_{i \ne j} \frac{1}{(1+\eta_{t} \hat{\uL}_{t,i})^{\alpha}})^{1/\alpha}} &\leq  \frac{1}{\min\qty(1, \min_{i \ne j}\frac{1}{(1+\eta_{t} \hat{\uL}_{t,i})^{\alpha}})^{1/\alpha}} \\
        &= \frac{1}{\min_{i\ne j} \frac{1}{1+\eta_{t}\hat{\uL}_{t,i}}} \\
        & \leq \max_{i \ne j} 1+\eta_{t} \hat{\uL}_{t,i}.
    \end{align*}
    Therefore, by (\ref{eq: lem V1 leader}) of Lemma~\ref{lem: jt stab sub lemma}, it holds for $c_v \geq 1.5 \cdot 2^{\frac{1}{\alpha}}$ that
    \begin{align*}
        \phi_j(\eta_t \hat{L}_t) - \phi_j(\eta_t (\hat{L}_{t}+\hat{\ell}_{t,j}e_j))  &\leq 3\alpha \eta_t\hat{\ell}_{t,j} \sum_{i\ne j} p_i(\eta_{t}\hat{\uL}_t; \alpha)^{1+\frac{1}{\alpha}} \\
        &= 3\alpha \eta_t\hat{\ell}_{t,j} \sum_{i\ne j} p_{t,i}^{1+\frac{1}{\alpha}}.
    \end{align*}
    Hence, by Lemma~\ref{lem: estimator}, we have
    \begin{align*}
        \E\qty[\hat{\ell}_{t,j} \qty(\phi_j(\eta_t \hat{L}_t) - \phi_j(\eta_t (\hat{L}_{t}+\hat{\ell}_{t,j}e_j))) \middle | \hat{L}_t] &\leq \E\qty[3\alpha \eta_t \hat{\ell}_{t,j}^2 \sum_{i\ne j} p_{t,i}^{1+\frac{1}{\alpha}} \middle | \hat{L}_t] \\
        &\leq 3\alpha \eta_t \sum_{i\ne j} p_{t,i}^{1+\frac{1}{\alpha}} \cdot \frac{\delta_*}{\gamma_t} \\
        &= \frac{3\alpha \delta_*}{\gamma_t \beta_t}\sum_{i\ne j} p_{t,i}^{1+\frac{1}{\alpha}},
    \end{align*}
    which concludes the proof.
\end{proof}

\subsection{Proof of Lemma~\ref{lem: V2 leader}}\label{app: V2 leader}
For the proof, we need a following result, whose proof is given in Appendix~\ref{app: jt stab sub 2}.
\begin{lemma}\label{lem: link to p and barR}
    Given $\lambda \in \sR^K$, it holds that
    \begin{equation*}
        \sum_{i \ne j^*(\lambda)} p_i(\lambda;\alpha)^{1/\alpha} \geq \frac{1}{2} \min\qty(1,\sum_{i \ne j^*(\lambda)} \frac{1}{(1+\lambda_i)^\alpha}).
    \end{equation*}
\end{lemma}
\begin{proof}[Proof of Lemma~\ref{lem: V2 leader}]
  Let $j= j_t \in V_2$ for notational simplicity.
  Since $\phi_i(\lambda)$ is increasing w.r.t.~$\lambda_k$ for $i \ne k$, we have
  \begin{align*}
  \E\qty[\hat{\ell}_{t,j} \qty(\phi_{j}(\eta_t \hat{L}_t) - \phi_{j}(\eta_t \hat{L}_{t+1})) \middle | \hat{L}_t] &\leq \E\qty[\hat{\ell}_{t,j} \qty(\phi_{j}(\eta_t \hat{L}_t) - \phi_{j}(\eta_t (\hat{L}_{t}+\hat{\ell}_{t,j}e_{j}))) \middle | \hat{L}_t].
  \end{align*}
  Define $s_t^{+} = \sum_{i \ne j} (1+\eta_t \hat{\uL}_{t,i})^{-\alpha}$, so that $s_t = \min(1,s_t^{+})$ for $s_t$ defined in (\ref{def: st}).
  \paragraph{Case 1.}
  First, consider the case where $s_t^{+} \geq 1/3$.
  In this case, Lemma~\ref{lem: link to p and barR} shows that
  \begin{align*}
      \sum_{i \ne j}p_{t,i}^{1/\alpha} \geq \frac{1}{2} \min(1, s_t^{+}) \geq \frac{1}{6}.
  \end{align*}
  On the other hand, by (\ref{eq: lem V2 leader}) of Lemma~\ref{lem: jt stab sub lemma}, we have
  \begin{align*}
      \phi_{j}(\eta_t \hat{L}_t) - \phi_{j}(\eta_t (\hat{L}_{t}+\hat{\ell}_{t,j}e_{j})) \leq \alpha \eta_t \hat{\ell}_{t,j} \phi_{j}(\eta_t \hat{L}_t).
  \end{align*}
  Therefore, similarly to the proof of Lemma~\ref{lem: V1 leader}, we have
  \begin{align*}
      \E\qty[\hat{\ell}_{t,j} \qty(\phi_{j}(\eta_t \hat{L}_t) - \phi_{j}(\eta_t \hat{L}_{t+1})) \middle | \hat{L}_t] &\leq \alpha \eta_t \E\qty[w_{t,j} \hat{\ell}_{t,j}^2\middle | \hat{L}_t] \\
      &\leq 4\alpha \eta_t. \tag{by Lemma~\ref{lem: estimator}}
  \end{align*}
  Hence, we obtain
  \begin{align*}
        \E\qty[\hat{\ell}_{t,j} \qty(\phi_{j}(\eta_t \hat{L}_t) - \phi_{j}(\eta_t \hat{L}_{t+1})) \middle | \hat{L}_t] \leq 4\alpha \eta_t \leq 24\alpha \eta_t  \sum_{i \ne j}p_{t,i}^{1/\alpha} .
  \end{align*}
  \paragraph{Case 2.}
  Next, suppose $s_t^{+} < 1/3$.
  Since $w_{t,i} \leq \frac{1}{(1+\eta_t \hat{\uL}_{t,i})^{\alpha}}$ holds for any $t\in \sN$ and $i\in V$, we obtain
  \begin{align*}
      \sum_{i\ne j} w_{t,i}\leq s_t^{+} < \frac{1}{3} \implies w_{t,j} = 1- \sum_{i\ne j}w_{t,i} \geq \frac{2}{3}.
  \end{align*}
  Again by (\ref{eq: lem V2 leader}) of Lemma~\ref{lem: jt stab sub lemma}, we obtain
  \begin{align*}
      \phi_{j}(\eta_t \hat{L}_t) - \phi_{j}(\eta_t (\hat{L}_{t}+\hat{\ell}_{t,j}e_{j})) \leq 4\alpha \eta_t \hat{\ell}_{t,j} \sum_{i\ne j} p_{t,i}^{1/\alpha},
  \end{align*}
  which implies
  \begin{align*}
      \E\qty[\hat{\ell}_{t,j} \qty(\phi_{j}(\eta_t \hat{L}_t) - \phi_{j}(\eta_t \hat{L}_{t+1})) \middle | \hat{L}_t] &\leq 4\alpha \eta_t \sum_{i\ne j} p_{t,i}^{1/\alpha} \cdot \E\qty[\hat{\ell}_{t,j}^2 \middle|\hat{L}_t] \\
      &\leq 16\alpha \eta_t \sum_{i\ne j} p_{t,i}^{1/\alpha} \frac{1}{w_{t,j}} \tag{by Lemma~\ref{lem: estimator}} \\
      &\leq 24\alpha \eta_t \sum_{i\ne j} p_{t,i}^{1/\alpha}.
  \end{align*}
  This result concludes the proof.
\end{proof}

\subsection{Proof of Lemma~\ref{lem: jt stab sub lemma}}\label{app: jt stab sub 1}
Here, we establish the main technical lemma used in the regret analysis.
\begin{proof}
  Fix $j^* = j^*(\lambda)$.
  \paragraph{Proof of (\ref{eq: lem V1 leader}).}
    By definition, we have
    \begin{equation}\label{eq: phi jstar integral}
        \phi_{j^*}(\lambda+xe_{j^*}) = \int_0^{\infty} f(z) \prod_{i \ne j^*} F(z+\ul_i-x ) \dd z.
    \end{equation}
    For $s\in [0,x]$, define $a_i(s; \ul) = \ul_i -s$.
    Then, by differentiating $\phi_{j^*}(\lambda+xe_{j^*})$ with respect to $x$, we obtain that
    \begin{equation*}
        \phi_{j^*}(\lambda) -  \phi_{j^*}(\lambda +x e_{j^*})  = \int_0^x \sum_{i \ne j^*} I_i(s) \dd s,
    \end{equation*}
    where
    \begin{equation}\label{def: I for stab}
        I_i(s) = \int_{\max(0, -a_i(s))}^\infty f(z)f(z+a_i(s)) \prod_{k\ne i, j^*} F(z+a_k(s))\dd z.
    \end{equation}
    Since $f(z) \leq \alpha$ and $F(z)\leq 1$ for any $z\geq 0$, it is obvious that $I_i(s) \leq \alpha$ for any $i \ne j^*$ and $s\in [0,x]$.
    Here, by assumption $x \in \qty[0,\frac{1+\ul_i}{1.5(\alpha+1)}]$ for all $i \ne j^*$, it holds that
    \begin{equation*}
        1+\ul_i - s \geq \qty(1-\frac{1}{1.5(\alpha+1)}) (1+\ul_i),
    \end{equation*}
    which implies for $\alpha \geq 2$ that
    \begin{align*}
        (1+\ul_i -s)^{-\alpha-1} \leq (1+\ul_i)^{-\alpha-1} \qty(1-\frac{1}{1.5(\alpha+1)})^{-(\alpha+1)} < 3(1+\ul_i)^{-\alpha-1}.
    \end{align*}
    \paragraph{Case 1.} If $a_i(s) \geq 0$, then for all $z \geq 0$, we have
    \begin{equation*}
        f(z+a_i(s)) =  \frac{\alpha}{(z+1+a_i(s))^{\alpha+1}}\leq \frac{\alpha}{(1+a_i(s))^{\alpha+1}} \leq  \frac{3\alpha}{(1+\ul_i)^{\alpha+1}}.
    \end{equation*}
    Since $F(z+a_k(s)) \leq 1$ for all $k$ and $\int_0^\infty f(z) = F(\infty) = 1$, this implies that
    \begin{equation*}
        I_i(s) \leq \frac{3\alpha}{(1+\ul_i)^{\alpha+1}}, \text{ if } a_i(s) \geq 0.
    \end{equation*}
    \paragraph{Case 2.} When $a_i(s) < 0$, we have that
    \begin{equation*}
        \ul_i  < s \leq x \leq \frac{1+\ul_i}{1.5(\alpha+1)} \implies \ul_i \leq \frac{2}{3\alpha+1}.
    \end{equation*}
    Hence, when $a_i(s) <0$, we have for $\alpha \geq 2$ that
    \begin{equation*}
        \frac{1}{(1+\ul_i)^{\alpha+1}} \geq \qty(1+\frac{2}{3\alpha+1})^{-\alpha-1} > \frac{1}{3}.
    \end{equation*}
    This implies that
    \begin{align}\label{eq: I bound 1}
        I_i(s) \leq \alpha \leq \alpha \frac{3}{(1+\ul_i)^{\alpha+1}}.
    \end{align}
    Therefore, (\ref{eq: I bound 1}) holds regardless of the sign of $a_i(s)$.

    Next, it remains to show that $I_i(s)$ is also bounded above by the function of $\sigma_i$.
    By definition of $\sigma_i$, there are $\max(\sigma_i-2, 0)$ arms $k\ne i,j^*$ such that $\ul_k \leq \ul_i$.
    For such $k$, it holds that
    \begin{equation*}
        a_k(s) \leq a_i(s), \, \text{and } F(z+a_k(s)) \leq F(z+a_i(s)). 
    \end{equation*}
    Hence, we have
    \begin{align*}
        I_i(s) &\leq \int_{\max(0, -a_i(s))}^\infty f(z)f(z+a_i(s))  F(z+a_i(s))^{\max(\sigma_i - 2,0)}\dd z \\
               &\leq \int_{\max(0, -a_i(s))}^\infty f(z)\dd z \cdot \sup_{y\geq 0} f(y)  F(y)^{\max(\sigma_i - 2,0)} \\
               &\leq \sup_{y\geq 0} f(y)  F(y)^{\max(\sigma_i - 2,0)}.
    \end{align*}
    When $\sigma_i =2$, it is obvious that $I_i(s) \leq \alpha$.
    Hence, for $\alpha \geq 2$, it holds that
    \begin{align*}
        I_i(s) \leq \alpha \leq \alpha 2^{3/2} \qty(\frac{1}{\sigma_i})^{1+1/\alpha} \leq 3 \alpha \qty(\frac{1}{\sigma_i})^{1+1/\alpha}.
    \end{align*}
    When $\sigma_i \geq 3$, by definition of $f$ and $F$, we have 
    \begin{align*}
       \sup_{y\geq 0} f(y)  F(y)^{\max(\sigma_i - 2,0)} &= \sup_{y\ge 0} \frac{\alpha}{(1+y)^{\alpha+1}} \qty(1-\frac{1}{(y+1)^\alpha})^{\max(\sigma_i-2,0)} \\
       &= \sup_{u \in [0,1)} \alpha u^{1+1/\alpha} (1-u)^{\sigma_i-2},
    \end{align*}
    where the maximum is attained at
    \begin{equation*}
        (1+y)^{-\alpha} = u = \frac{1 +1/\alpha}{\sigma_i -1 +1/\alpha}.
    \end{equation*}
    Since $(1+1/x)/(y-1+1/x)$ is decreasing with respect to $x \geq 2$ for $y\geq 3$, we have
    \begin{align*}
        \sup_{u \in [0,1)} \alpha u^{1+1/\alpha} (1-u)^{\sigma_i-2} &\leq \alpha \qty(\frac{1 +1/\alpha}{\sigma_i -1 +1/\alpha})^{1+\frac{1}{\alpha}} \\
        &= \alpha \qty(\frac{3/2}{\sigma_i+1/2})^{1+\frac{1}{\alpha}}. \tag{for $\alpha \geq 2$} \\
        &\leq \alpha (3/2)^{3/2} \qty(\frac{1}{\sigma_i})^{1+\frac{1}{\alpha}} \\
        &\leq 2\alpha  \qty(\frac{1}{\sigma_i})^{1+\frac{1}{\alpha}}.
    \end{align*}
    Therefore, for any $\sigma_i \geq 2$, we have
    \begin{equation}\label{eq: I bound 2}
        I_i(s) \leq 3\alpha  \qty(\frac{1}{\sigma_i})^{1+\frac{1}{\alpha}}.
    \end{equation}
    In sum, by (\ref{eq: I bound 1}) and (\ref{eq: I bound 2}), we obtain that
    \begin{align*}
        I_i(s) \leq 3\alpha \min\qty(\frac{1}{(1+\ul_i)^\alpha}, \frac{1}{\sigma_i(\lambda)})^{1+1/\alpha} \leq 3\alpha p_i(\ul;\alpha)^{1+\frac{1}{\alpha}},
    \end{align*}
    which concludes the proof for (\ref{eq: lem V1 leader}).

    \paragraph{Proof of (\ref{eq: lem V2 leader}).}
    In this case, it is clear when $4\alpha \sum_{i\ne j^*}p_{i}(\lambda;\alpha)^{1/\alpha}\geq 1$, since $\phi_{j^*} \in (0,1)$.
    Therefore, suppose $4\alpha \sum_{i\ne j^*}p_{i}(\lambda;\alpha)^{1/\alpha} <1 $ and the existence of $i \ne j^*$ with $\sigma_i\geq 2$ such that $x > \frac{1+\ul_i}{1.5(\alpha+1)}$.

    Then, for every non-leader arm $k $ with $1<\sigma_k \leq \sigma_i$, we have $\lambda_k \leq \lambda_i$.
    Therefore,
    \begin{equation*}
        p_k^{1/\alpha} = \min\qty(\frac{1}{1+\ul_k}, \frac{1}{\sigma_k^{1/\alpha}}) \geq \min\qty(\frac{1}{1+\ul_i}, \frac{1}{\sigma_i^{1/\alpha}}) = p_i^{1/\alpha}.
    \end{equation*}
    Since there are $\sigma_i-1$ such arms in $V\setminus \{j^*\}$ including $i$ itself, we have
    \begin{align*}
        \sum_{i\ne j^*}p_i(\lambda;\alpha)^{1/\alpha} &\geq (\sigma_i-1) \min\qty(\frac{1}{1+\ul_i}, \frac{1}{\sigma_i^{1/\alpha}}),
    \end{align*}
    which implies
    \begin{align*}
        \alpha x  \sum_{i\ne j^*}p_i(\lambda;\alpha)^{1/\alpha}  &> \alpha \frac{1+\ul_i}{1.5(\alpha+1)}(\sigma_i-1)\min\qty(\frac{1}{1+\ul_i}, \frac{1}{\sigma_i^{1/\alpha}}) \\
        &=  \frac{\alpha}{1.5(\alpha+1)}(\sigma_i-1)\min\qty(1, \frac{1+\ul_i}{\sigma_i^{1/\alpha}}).
    \end{align*}
    When $\min(1,(1+\ul_i)/\sigma_i^{1/\alpha})=1$, we have
    \begin{align*}
         \alpha x  \sum_{i\ne j^*}p_i(\lambda;\alpha)^{1/\alpha}  &> \frac{\alpha}{1.5(\alpha+1)}(\sigma_i-1) \\
         &\geq \frac{4}{9} > \frac{1}{4} \tag{$\because \alpha \geq 2, \sigma_i \geq 2$},
    \end{align*}
    which contradicts the assumption.
    Hence, assume $\min(1,(1+\ul_i)/\sigma_i^{1/\alpha})<1$.
    Then, we have
    \begin{align*}
        \alpha x  \sum_{i\ne j^*}p_i(\lambda;\alpha)^{1/\alpha}  &> \frac{\alpha(1+\ul_i)}{1.5(\alpha+1)}\frac{\sigma_i-1}{\sigma_i^{1/\alpha}} \\
        &\geq \frac{4}{9} \frac{\sigma_i-1}{\sigma_i^{1/2}} \tag{$\because \alpha \geq 2, \ul_i \geq 0$} \\
        & \geq \frac{2\sqrt{2}}{9} > \frac{1}{4}\tag{$\because \sigma_i \geq 2$},
    \end{align*}
    which also contradicts the assumption.
    Therefore, if there exists an arm $i\ne j^*$ such that $x> \frac{1+\ul_i}{1.5(\alpha+1)}$, then $ 4\alpha x  \sum_{i\ne j^*}p_i(\lambda;\alpha)^{1/\alpha} \geq 1$ should hold.

    Finally, it remains to show
    \begin{equation*}
        \phi_{j^*}(\lambda) - \phi_{j^*}(\lambda+xe_{j^*})\leq \alpha x \phi_{j^*}(\lambda).
    \end{equation*}
    Let $G(z) = \prod_{i \ne j^*} F(z+\ul_i)$.
    By (\ref{eq: phi jstar integral}), we have
    \begin{align*}
        \phi_{j^*}(\lambda+xe_{j^*}) &= \int_0^{\infty} f(z) \prod_{i \ne j^*} F(z+\ul_i-x ) \dd z \\
        &= \int_{-x}^{\infty} f(z+x) \prod_{i \ne j^*} F(z+\ul_i) \dd z  \\
        &= \int_{-x}^{\infty} -(1-F(z+x))' G(z) \dd z \\
        &= -(1-F(z+x))G(z) \eval_{z=-x}^{z=\infty}  + \int_{-x}^{\infty} (1-F(z+x)) \dd G(z) \\
        &= G(-x) +\int_{-x}^{\infty} (1-F(z+x)) \dd G(z).
    \end{align*}
    Similarly, we can obtain
    \begin{align*}
        \phi_{j^*}(\lambda) = G(0) + \int_0^\infty  (1-F(z)) \dd G(z).
    \end{align*}
    Then, by the definition of Pareto distribution, we have
    \begin{align*}
        \phi_{j^*}(\lambda) - \phi_{j^*}(\lambda+xe_{j^*}) &= G(0) - G(-x) - \int_{-x}^0  (1-F(z+x)) \dd G(z) + \int_0^\infty (F(z+x)-F(z))\dd G(z) \\
        &= \int_{-x}^0  F(z+x) \dd G(z) + \int_0^\infty (F(z+x)-F(z))\dd G(z) \\
        &\leq \int_{-x}^0  \alpha x \dd G(z) + \int_0^\infty (F(z+x)-F(z))\dd G(z) \tag{$\because f(\cdot)\leq \alpha, z+x \in [0,x]$} \\
        &= \int_{-x}^0  \alpha x \dd G(z) + \int_0^\infty \int_z^{z+x} f(y) \dd y \dd G(z) \\
        &= \int_{-x}^0  \alpha x \dd G(z) + \int_0^\infty \int_z^{z+x} \frac{\alpha}{1+y}(1-F(y)) \dd y \dd G(z) \tag{$\because$ formulation of Pareto} \\
        &\leq \int_{-x}^0  \alpha x \dd G(z) + \int_0^\infty \alpha x (1-F(z)) \dd G(z) \tag{$\because \frac{1}{1+y}\leq 1$, $F$ is monotonically increasing} \\
        &\leq \alpha x \qty(G(0) + \int_0^\infty (1-F(z)) \dd G(z))  \tag{$\because G(-x) \geq 0$}\\
        &= \alpha x \phi_{j^*}(\lambda),
    \end{align*}
    which concludes the proof.
\end{proof}
\subsection{Proof of Lemma~\ref{lem: link to p and barR}}\label{app: jt stab sub 2}
\begin{proof}
  Since $p_i \in (0,1]$, it suffices to show 
    \begin{equation}\label{eq: lem 7 subeq}
        \sum_{i \ne j^*(\lambda)} p_i(\lambda;\alpha) = \sum_{i \ne j^*(\lambda)} \min\qty(\frac{1}{(1+\lambda_i)^\alpha}, \frac{1}{\sigma_i(\lambda)}) \geq \frac{1}{2} \min\qty(1,\sum_{i \ne j^*(\lambda)} \frac{1}{(1+\lambda_i)^\alpha}).
    \end{equation}
    For notational simplicity, suppose $\lambda_1 \leq \lambda_2 \leq \cdots \leq \lambda_K$, i.e., $j^* =1$.
    \paragraph{Case 1.}
    First, assume $\sum_{i=2}^K (1+\lambda_i)^{-\alpha} \leq 1$.
    In this case, $\sum_{i=2}^K (1+\lambda_i)^{-\alpha}=\min(1,\sum_{i=2}^K (1+\lambda_i)^{-\alpha}).$
    Then, we claim that for every $i \geq 2$, $p_i(\lambda;\alpha) \geq (1+\lambda_i)^{-\alpha}/2$.

    Suppose not, i.e., assume $p_{i'}(\lambda;\alpha) < (1+\lambda_{i'})^{-\alpha}/2$ for some $i' \geq 2$.
    This implies that
    \begin{equation*}
        p_{i'}(\lambda;\alpha) = \min\qty(\frac{1}{(1+\lambda_{i'})^\alpha}, \frac{1}{i'}) < \frac{1}{2(1+\lambda_{i'})^\alpha}.
    \end{equation*}
    Therefore, $p_{i'}(\lambda;\alpha) < (1+\lambda_{i'})^{-\alpha}/2$ happen only if $i' > 2(1+\lambda_{i'})^\alpha$.
    Since $(1+x)^{\alpha}$ is increasing, we obtain
    \begin{equation*}
        (1+\lambda_2)^{-\alpha} \geq (1+\lambda_3)^{-\alpha} \geq \cdots \geq (1+\lambda_{i'})^{-\alpha} > \frac{2}{i'}.
    \end{equation*}
    Therefore, for any $i' \geq 2$,
    \begin{align*}
        \sum_{i=2}^K (1+\lambda_i)^{-\alpha} \geq \sum_{i=2}^{i'}(1+\lambda_i)^{-\alpha} > (i'-1)\frac{2}{i'} \geq 1,
    \end{align*}
    which contradicts the assumption.
    Therefore, for any $i \geq 2$, we have
    \begin{equation*}
      p_i(\lambda;\alpha) \geq (1+\lambda_i)^{-\alpha}/2
    \end{equation*}
    when $\sum_{i=2}^K (1+\lambda_i)^{-\alpha} \leq 1$, which implies
    \begin{align*}
        \sum_{i=2}^K p_i(\lambda;\alpha) \geq \frac{1}{2}\sum_{i=2}^K \frac{1}{(1+\lambda_i)^\alpha} \geq \frac{1}{2}\min\qty(1, \sum_{i =2}^K \frac{1}{(1+\lambda_i)^\alpha}).
    \end{align*}
    \paragraph{Case 2.}
    Next, assume $\sum_{i=2}^K (1+\lambda_i)^{-\alpha} > 1$.
    In this case, $\min\qty(1, \sum_{i =2}^K \frac{1}{(1+\lambda_i)^\alpha})=1$.
    Let $s_i^+ = \sum_{i=2}^K (1+\lambda_i)^{-\alpha}$ and define $a_i = \frac{(1+\lambda_i)^{-\alpha}}{s_i^+}$ so that $a_i<(1+\lambda_i)^{-\alpha}$. 
    Then, $\sum_{i=2}^K a_i = 1$ and $a_2 \geq a_3 \geq \cdots \geq a_K$.
    This means that we can apply the result of the case 1 for $a_i$ instead of $(1+\lambda_i)^{-\alpha}$, which provides
    \begin{align*}
        \min\qty(a_i, \frac{1}{i}) \geq \frac{a_i}{2}, \, \forall i \geq 2.
    \end{align*}
    Therefore, by $\sum_{i=2}^K a_i= 1$, we obtain
    \begin{align*}
        \sum_{i=2}^K p_i(\lambda;\alpha) \geq \sum_{i=2}^K \min\qty(a_i, \frac{1}{i}) \geq \frac{1}{2} = \frac{1}{2}\min\qty(1, \sum_{i =2}^K \frac{1}{(1+\lambda_i)^\alpha}),
    \end{align*}
    which concludes the proof.
\end{proof}
\subsection{Proof of inequality (\ref{eq: solving eq})}\label{app: solving eq}
\begin{proof}
  Here, we aim to show that 
  \begin{equation*}
    \{ x \geq 0 : x \leq a x^{2/3} + b\sqrt{x} + c \} \subseteq \qty{ x \geq 0: x \leq 4a^3 + \frac{9}{4}b^2 + 3c }
  \end{equation*}
  for $a>0, b>0, c>0$.
  Hence, it suffices to show
  \begin{align*}
    ax^{2/3} \leq \frac{x}{3} + \frac{4a^3}{3}, \qq{and} b\sqrt{x} \leq \frac{x}{3} + \frac{3b^2}{4}.
  \end{align*}
  By elementary calculation, for any $x\geq 0$, we obtain
  \begin{align*}
    0&\leq \frac{(x^{1/3}-2a)^2(x^{1/3}+a)}{3} = \frac{x-3ax^{2/3}+4a^3}{3} \\
    0&\leq \frac{(2\sqrt{x}-3b)^2}{12} = \frac{4x-12b\sqrt{x}+9b^2}{12},
  \end{align*}
  which concludes the proof.
\end{proof}

\section{Additional experiments}\label{app: exp}
In this section, we provide the complete experimental setup and the additional results omitted in the main text.
Whenever $V_2\neq\emptyset$, we report separately the cases in which the optimal arm belongs to $V_1$ and $V_2$.
We first recall the fractional domination number used by \citet{tsuchiya2024simple}, which is distinct from the fractional \textbf{weak} domination number considered in this paper.
\begin{definition}[Fractional domination number in \citet{tsuchiya2024simple}]\label{def: weak domination tsucchi}
  For a graph $G=(V,E)$, the fractional domination number $\delta_*'(G)$ is the optimal value of the following linear program 
  \begin{align*}
    &\mathrm{minimize} \quad \sum_{i \in V} x_i, \numberthis{\label{def: linear program tsuchhi}} \\
    &\mathrm{subject~to} \sum_{i \in \inb(j)} x_i \geq 1, \forall j \in V, \hspace{0.6em} 0 \leq x_i \leq 1, \forall i \in V.
  \end{align*}
  Replacing constraints $x_i\in [0,1]$ with $x_i\in \{0,1\}$ gives the corresponding integer program, whose optimal value is called the domination number $\delta'(G)$.
\end{definition}
The difference between (\ref{def: linear program tsuchhi}) and our definition in (\ref{def: linear program}) lies in the vertices subject to the coverage constraints.
Our definition imposes these constraints only for $j\in V_1$, whereas (\ref{def: linear program tsuchhi}) imposes them for every $j\in V$.
Consequently, our fractional weak domination number always satisfies $\delta_*\leq \delta_*' \leq \delta'$.
Our definition follows \citet{chen2021understanding}.
The code and instructions required to reproduce all experiments will be made publicly available on GitHub.

\subsection{Graph structures}
The regret experiments use $K=8$ arms and four feedback graphs, giving seven combinations of graph structure and optimal arm location.

\paragraph{Ring structure.}
The \textsc{Loopless Ring} is a bidirected ring where every arm reveals the losses of its two adjacent arms but not its own loss.
Thus, $V_1=V$ and $V_2=\emptyset$.

The \textsc{Partially Looped Ring}, used in Section~\ref{sec: exp}, is obtained by adding self-loops to three arms, so that $|V_1|=5$ and $|V_2|=3$.
For \textsc{Loopless Ring} and \textsc{Partially Looped Ring}, the four domination parameters, $\delta_*, \delta_*', \delta$, $\delta'$ coincide.

\paragraph{Bandit with apple tasting structure.}
The \textsc{Bandit Apple} graph combines an apple tasting graph consisting of arms $1$ and $2$ with bandit feedback graph of six arms, where arm $1$ reveals the losses of both arms $1$ and $2$, while arm $2$ has no outgoing edges.
Each arm in $\{3,\ldots,8\}$ has only a self-loop, which is the graph used in the runtime comparison in Section~\ref{sec: exp}.
In this case, $V_1 = \qty{2}$, $V_2 = \qty{1,3,4,\ldots,8}$, and $\delta=\delta_*=1$ and $\delta_*'=\delta'=7$.


\paragraph{Domination number structure.}
To examine the difference between the weak domination number and its fractional counterpart, we construct a graph $G_D=([8], E_D)$, where
\begin{align*}
  E_D  = \begin{pmatrix}
1 & 0 & 0 & 1 & 1 & 0 & 0 & 0 \\
0 & 1 & 0 & 0 & 1 & 1 & 0 & 0 \\
0 & 0 & 1 & 0 & 0 & 1 & 1 & 0 \\
0 & 0 & 0 & 0 & 0 & 0 & 1 & 1 \\
0 & 0 & 0 & 1 & 0 & 0 & 0 & 1 \\
1 & 0 & 0 & 0 & 0 & 0 & 0 & 0 \\
0 & 1 & 0 & 0 & 0 & 0 & 0 & 0 \\
0 & 0 & 1 & 0 & 0 & 0 & 0 & 0
\end{pmatrix},
\end{align*}
which satisfies $\delta_*(G_D)=5/2<\delta(G_D)=3$ and $\delta_*'(G_D)=\delta'(G_D)=4$.
Here, $(E_D)_{ij}=1$ means that playing arm $i$ reveals the loss of arm $j$.

For verification, it holds that $V_1 = \qty{4,5,6,7,8}$ and $V_2 =\qty{1,2,3}$, and our LP in (\ref{def: linear program}) requires
\begin{equation*}
  x_1 + x_5 \geq 1, \, x_1+x_2 \geq 1, \, x_2 + x_3 \geq 1, \, x_3+x_4 \geq 1, \text{ and } x_4+x_5 \geq 1.
\end{equation*}
Hence, the optimal solution is
\begin{equation*}
  x_1^* = x_2^* = x_3^* = x_4^* = x_5^* = \frac{1}{2}, \quad x_6^*=x_7^*=x_8^*=0,
\end{equation*}
which makes $\delta_*(G_D)=5/2$.
On the other hand, its integer solution can be obtained by $x_1^*=x_3^* = x_5^*=1$, i.e., $\delta(G_D)=3$.

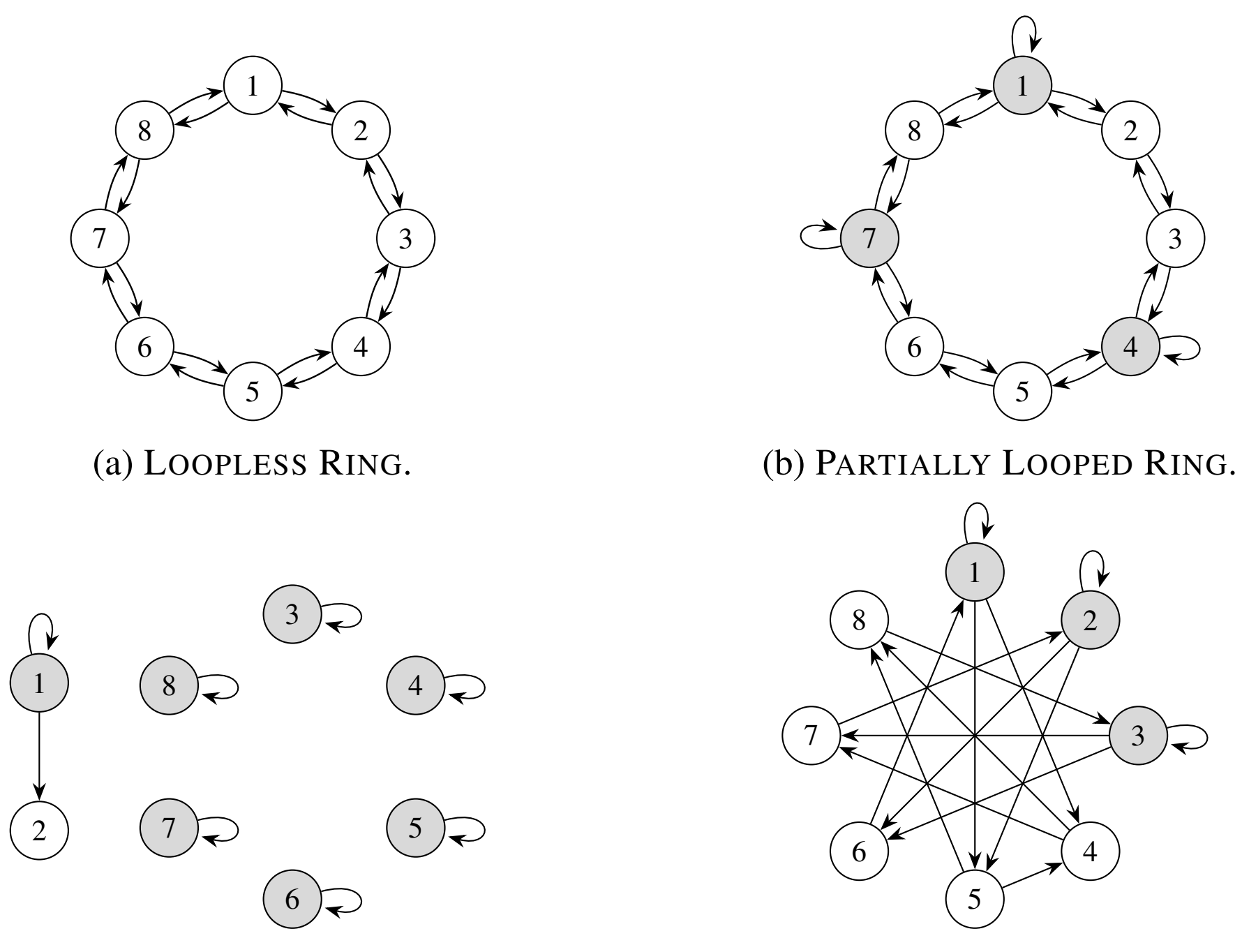
\begin{figure}[t]
\centering
\begin{subfigure}[t]{0.4\textwidth}
\centering
\begin{tikzpicture}[
    obs/.style={
        -{Stealth[length=1.6mm,width=1.2mm]},
        line width=0.45pt
    }
]
  \node[vone] (r1) at (90:1.45) {1};
  \node[vone] (r2) at (45:1.45) {2};
  \node[vone] (r3) at (0:1.45) {3};
  \node[vone] (r4) at (-45:1.45) {4};
  \node[vone] (r5) at (-90:1.45) {5};
  \node[vone] (r6) at (-135:1.45) {6};
  \node[vone] (r7) at (180:1.45) {7};
  \node[vone] (r8) at (135:1.45) {8};

  \draw[obs,bend left=12] (r1) to (r2);
  \draw[obs,bend left=12] (r2) to (r1);

  \draw[obs,bend left=12] (r2) to (r3);
  \draw[obs,bend left=12] (r3) to (r2);

  \draw[obs,bend left=12] (r3) to (r4);
  \draw[obs,bend left=12] (r4) to (r3);

  \draw[obs,bend left=12] (r4) to (r5);
  \draw[obs,bend left=12] (r5) to (r4);

  \draw[obs,bend left=12] (r5) to (r6);
  \draw[obs,bend left=12] (r6) to (r5);

  \draw[obs,bend left=12] (r6) to (r7);
  \draw[obs,bend left=12] (r7) to (r6);

  \draw[obs,bend left=12] (r7) to (r8);
  \draw[obs,bend left=12] (r8) to (r7);

  \draw[obs,bend left=12] (r8) to (r1);
  \draw[obs,bend left=12] (r1) to (r8);
\end{tikzpicture}
\caption{\textsc{Loopless Ring}.}
\label{fig-loopless-ring}
\end{subfigure}
\quad
\begin{subfigure}[t]{0.4\textwidth}
\centering
\begin{tikzpicture}
  \node[vtwo] (r1) at (90:1.45) {1};
  \node[vone] (r2) at (45:1.45) {2};
  \node[vone] (r3) at (0:1.45) {3};
  \node[vtwo] (r4) at (-45:1.45) {4};
  \node[vone] (r5) at (-90:1.45) {5};
  \node[vone] (r6) at (-135:1.45) {6};
  \node[vtwo] (r7) at (180:1.45) {7};
  \node[vone] (r8) at (135:1.45) {8};

  \draw[obs,bend left=12] (r1) to (r2);
  \draw[obs,bend left=12] (r2) to (r1);

  \draw[obs,bend left=12] (r2) to (r3);
  \draw[obs,bend left=12] (r3) to (r2);

  \draw[obs,bend left=12] (r3) to (r4);
  \draw[obs,bend left=12] (r4) to (r3);

  \draw[obs,bend left=12] (r4) to (r5);
  \draw[obs,bend left=12] (r5) to (r4);

  \draw[obs,bend left=12] (r5) to (r6);
  \draw[obs,bend left=12] (r6) to (r5);

  \draw[obs,bend left=12] (r6) to (r7);
  \draw[obs,bend left=12] (r7) to (r6);

  \draw[obs,bend left=12] (r7) to (r8);
  \draw[obs,bend left=12] (r8) to (r7);

  \draw[obs,bend left=12] (r8) to (r1);
  \draw[obs,bend left=12] (r1) to (r8);

  \path[obs] (r1) edge[loop above] (r1);
  \path[obs] (r4) edge[loop right] (r4);
  \path[obs] (r7) edge[loop left] (r7);
\end{tikzpicture}
\caption{\textsc{Partially Looped Ring}.}
\label{fig-partially-looped-ring}
\end{subfigure}
\\
\begin{subfigure}[t]{0.4\textwidth}
\centering
\begin{tikzpicture}
  \node[vtwo] (r1) at (-2.4,0.7) {1};
  \node[vone] (r2) at (-2.4,-0.7) {2};

  \node[vtwo] (r3) at (90:1.35) {3};
  \node[vtwo] (r4) at (30:1.35) {4};
  \node[vtwo] (r5) at (-30:1.35) {5};
  \node[vtwo] (r6) at (-90:1.35) {6};
  \node[vtwo] (r7) at (-150:1.35) {7};
  \node[vtwo] (r8) at (150:1.35) {8};

  \draw[obs] (r1) -- (r2);

  \path[obs] (r1) edge[loop above] (r1);
  \path[obs] (r3) edge[loop right] (r3);
  \path[obs] (r4) edge[loop right] (r4);
  \path[obs] (r5) edge[loop right] (r5);
  \path[obs] (r6) edge[loop right] (r6);
  \path[obs] (r7) edge[loop right]  (r7);
  \path[obs] (r8) edge[loop right]  (r8);
\end{tikzpicture}
\caption{\textsc{Bandit Apple}.}
\label{fig-bandit-apple}
\end{subfigure}
\quad
%
\begin{subfigure}[t]{0.4\textwidth}
\centering
\begin{tikzpicture}
  \node[vtwo] (d1) at (90:1.55) {1};
  \node[vtwo] (d2) at (45:1.55) {2};
  \node[vtwo] (d3) at (0:1.55) {3};
  \node[vone] (d4) at (-45:1.55) {4};
  \node[vone] (d5) at (-90:1.55) {5};
  \node[vone] (d6) at (-135:1.55) {6};
  \node[vone] (d7) at (180:1.55) {7};
  \node[vone] (d8) at (135:1.55) {8};

  \path[obs] (d1) edge[loop above] (d1);
  \path[obs] (d2) edge[loop above] (d2);
  \path[obs] (d3) edge[loop right] (d3);

  \draw[obs] (d1) -- (d4);
  \draw[obs] (d1) -- (d5);

  \draw[obs] (d2) -- (d5);
  \draw[obs] (d2) -- (d6);

  \draw[obs] (d3) -- (d6);
  \draw[obs] (d3) -- (d7);

  \draw[obs] (d4) -- (d7);
  \draw[obs] (d4) -- (d8);

  \draw[obs] (d5) -- (d4);
  \draw[obs] (d5) -- (d8);

  \draw[obs] (d6) -- (d1);
  \draw[obs] (d7) -- (d2);
  \draw[obs] (d8) -- (d3);
 
\end{tikzpicture}
\caption{\textsc{Domination} $G_D$.}
\label{fig-domination-gap}









\end{subfigure}

\caption{
Feedback graphs used in the experiments.
White vertices belong to $V_1$, while gray vertices belong to $V_2$.
A directed edge $i\to j$ means that playing arm $i$ reveals the loss of arm $j$.
}
\label{fig: graphs}
\end{figure}

The vertex partitions and the corresponding values of the weak domination number $\delta$, fractional weak domination number $\delta_*$, fractional domination number $\delta_*'$, and domination number $\delta'$ are summarized in Table~\ref{tab: graph}.
Each graph structure is illustrated in Figure~\ref{fig: graphs}.
\begin{table}[t]
\centering
\caption{Vertex partitions and domination parameters of the feedback graphs used in the experiments.}
\label{tab: graph}
\begin{tabular}{lcccccc}
\toprule
Graph structure & $V_1$& $V_2$& $\delta_*$ & $\delta$  & $\delta_*'$ & $\delta'$ \\
\midrule
\textsc{Loopless Ring} & $[8]$ & $\emptyset$ & $4$& $4$& $4$ & $4$ \\
\textsc{Partially Looped Ring} & $\{2,3,5,6,8\}$ & $\{1,4,7\}$ & $3$ & $3$ & $3$ & $3$\\
\textsc{Bandit Apple}& $\{2\}$& $\{1,3,\ldots,8\}$& $1$& $1$& $7$ & $7$\\
\textsc{Domination} $G_D$& $\{4,5,6,7,8\}$ & $\{1,2,3\}$ & $\frac{5}{2}$  & $3$  & $4$ & $4$ \\
\bottomrule
\end{tabular}
\end{table}

\subsection{Environments}
We evaluate all policies under three loss generation procedures.

\paragraph{Oblivious block losses.}
We use a deterministic oblivious sequence with horizon $T=10000$ and blocks of length $100$.
In the first block of every four block cycle, one suboptimal arm (decoy arm) has loss $0$ and the intended best arm has loss $1$.
Their losses are reversed in the remaining three blocks, while every other arm has loss $1$ throughout.
Since the horizon contains $25$ complete cycles, the cumulative losses of the decoy and intended best arms are $7500$ and $2500$, respectively.
Thus, the intended best arm is the unique best fixed arm, although the decoy appears optimal during the first block of every cycle.

\paragraph{Stochastically constrained adversarial Bernoulli losses.}
This setting follows the setting usually considered in BOBW literature~\citep{zimmert2021tsallis,pmlr-v201-honda23a,lee2024follow}.
In this setting, we use the construction with gap $\Delta=0.1$ and geometrically increasing phase lengths.
Starting from $H_1=1$, the phase lengths satisfy
\begin{equation*}
    H_{r+1}=\left\lceil 1.6 H_r\right\rceil.
\end{equation*}
During low-loss phases, losses are independently generated according to
\begin{equation*}
    \ell_{t,i^*}\sim\operatorname{Ber}(0)
    \qq{and}
    \ell_{t,i}\sim\operatorname{Ber}(\Delta)
    \quad
    \text{for all } i\neq i^*.
\end{equation*}
During high-loss phases, we instead use
\begin{equation*}
    \ell_{t,i^*}\sim\operatorname{Ber}(1-\Delta)
    \qq{and}
    \ell_{t,i}\sim\operatorname{Ber}(1)
    \quad
    \text{for all } i\neq i^*.
\end{equation*}
The phases alternate between these two cases.
Consequently, the absolute loss levels change across phases, while
\begin{equation*}
    \E[\ell_{t,i}-\ell_{t,i^*}]=\Delta
\end{equation*}
holds for every suboptimal arm and every round.

\paragraph{Stationary Bernoulli losses.}
For the stochastic environment, losses are independently generated from fixed Bernoulli distributions.
The unique optimal arm has mean loss $0.3$, while each of the remaining seven arms has mean loss $0.4$.

\subsection{Results}
Across the additional feedback graphs, both practical FTPL variants achieve lower regret than FTPL (theory). 
This pattern appears in all three environments and whether the optimal arm belongs to $V_1$ or $V_2$.
The graph structure affects the magnitude of regret but does not change the qualitative comparison.

\begin{figure*}[t]
    \centering
    \begin{subfigure}[t]{0.47\textwidth}
        \centering
        \includegraphics[width=\linewidth]
        {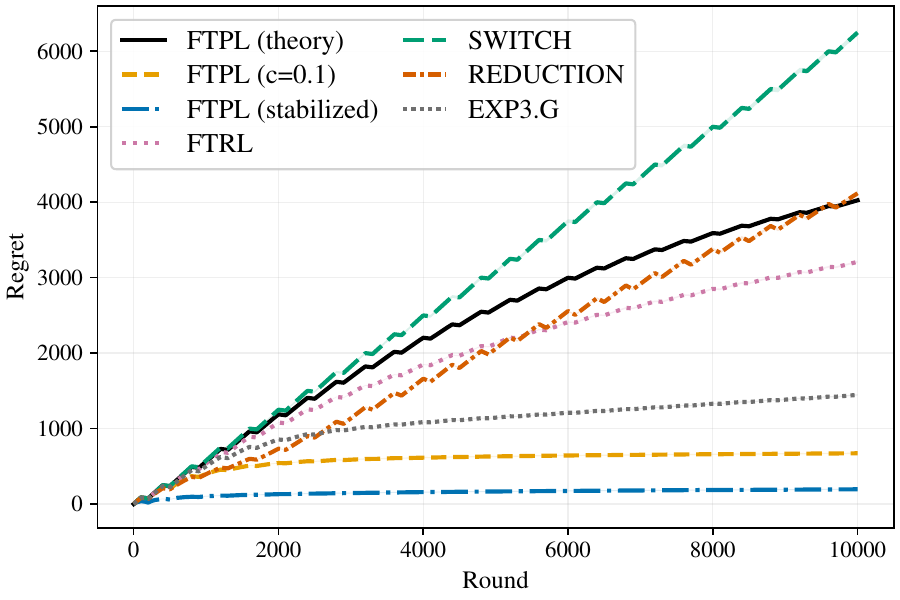}\vspace{-0.25em}
        \caption{\textsc{Loopless Ring}.}
    \end{subfigure}
    \\
    \begin{subfigure}[t]{0.47\textwidth}
        \centering
        \includegraphics[width=\linewidth]
        {figures/ring_loops_best_loopless__oblivious_blocks__with_exp3g__ci95.pdf}\vspace{-0.25em}
        \caption{\textsc{Partially Looped Ring} with the optimal arm in $V_1$.}
    \end{subfigure}
    \quad
      \begin{subfigure}[t]{0.47\textwidth}
        \centering
        \includegraphics[width=\linewidth]
        {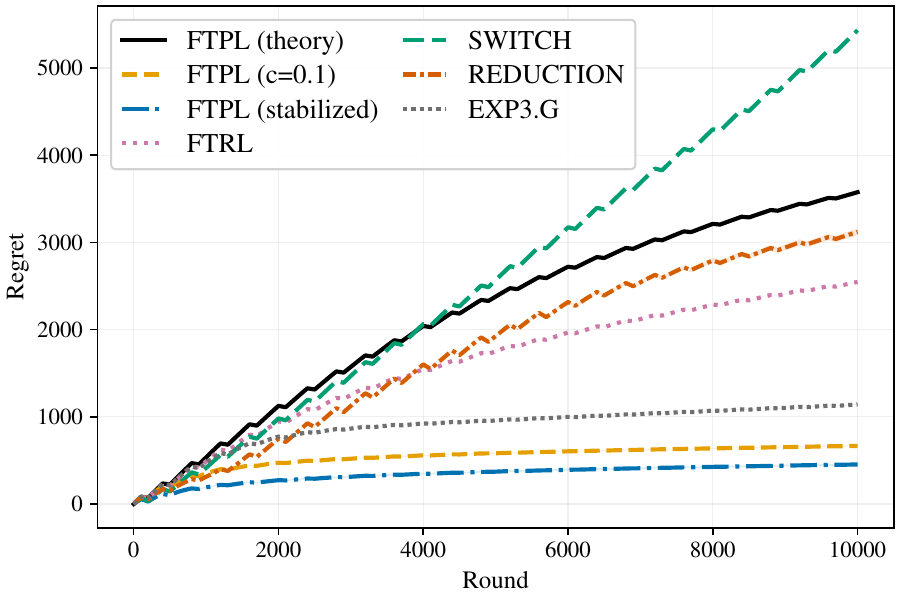}\vspace{-0.25em}
        \caption{\textsc{Partially Looped Ring} with the optimal arm in $V_2$.}
    \end{subfigure}
    \\
        \begin{subfigure}[t]{0.47\textwidth}
        \centering
        \includegraphics[width=\linewidth]
        {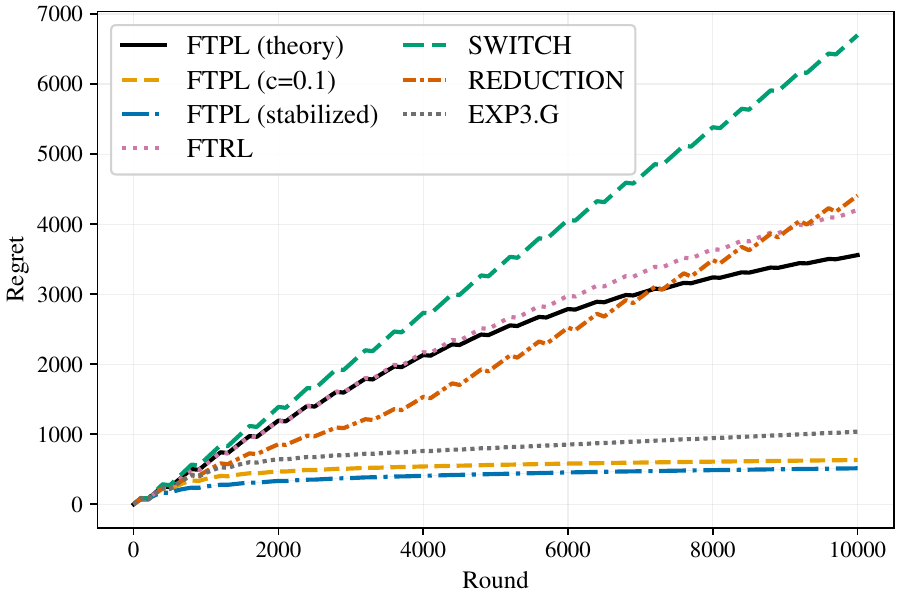}\vspace{-0.25em}
        \caption{\textsc{Bandit Apple} with the optimal arm in $V_1$.}
    \end{subfigure}
    \quad
      \begin{subfigure}[t]{0.47\textwidth}
        \centering
        \includegraphics[width=\linewidth]
        {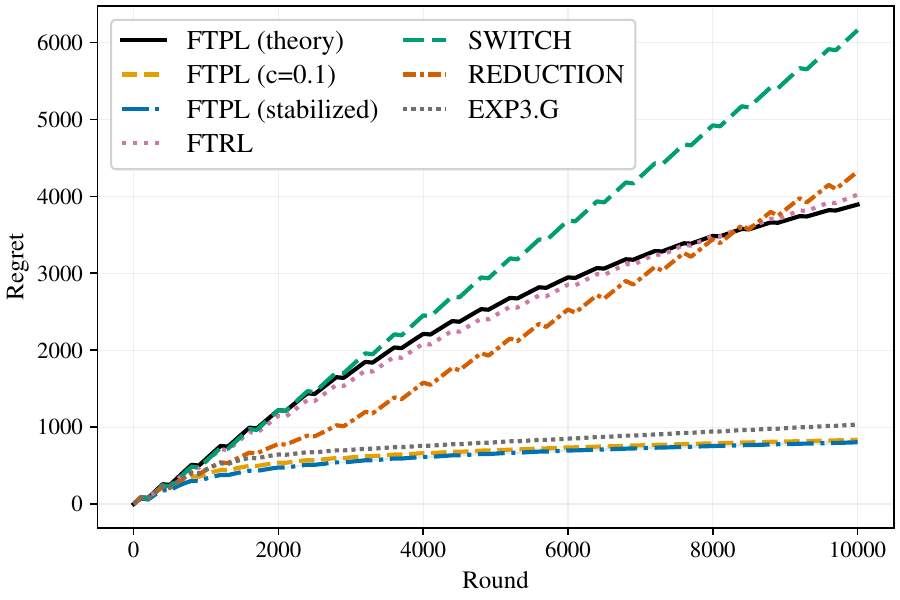}\vspace{-0.25em}
        \caption{\textsc{Bandit Apple} with the optimal arm in $V_2$.}
    \end{subfigure}
        \\
        \begin{subfigure}[t]{0.47\textwidth}
        \centering
        \includegraphics[width=\linewidth]
        {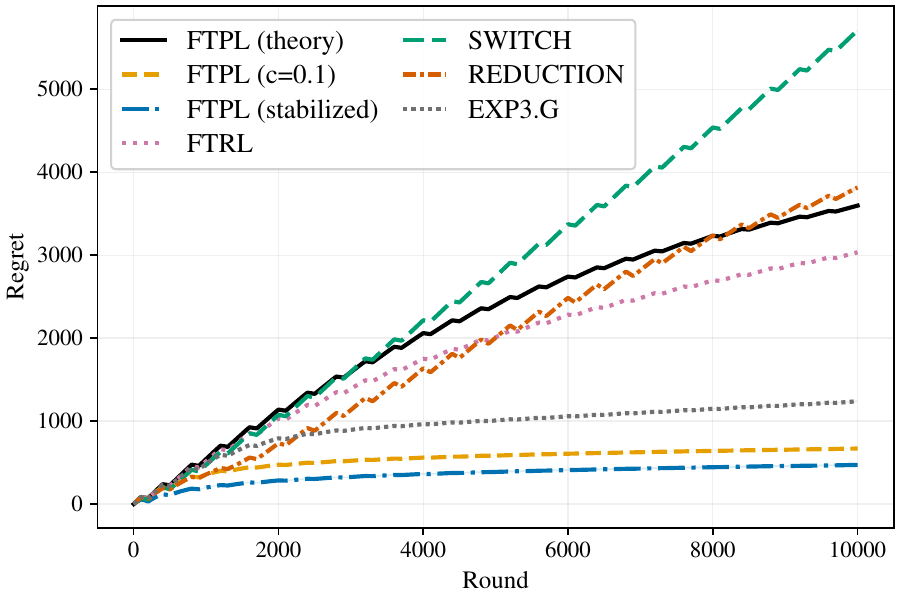}\vspace{-0.25em}
        \caption{\textsc{Domination} with the optimal arm in $V_1$.}
    \end{subfigure}
    \quad
      \begin{subfigure}[t]{0.47\textwidth}
        \centering
        \includegraphics[width=\linewidth]
        {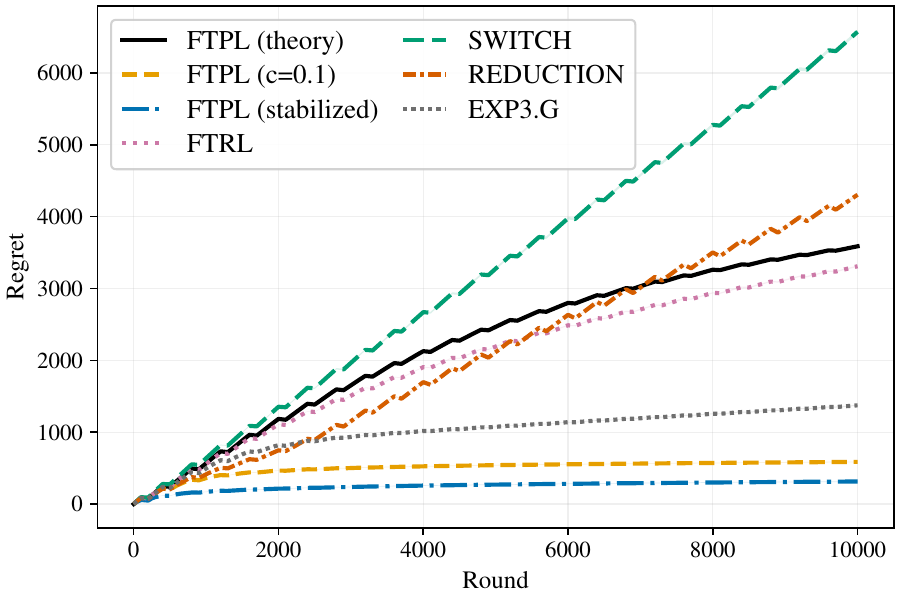}\vspace{-0.25em}
        \caption{\textsc{Domination} with the optimal arm in $V_2$.}
    \end{subfigure}
    \caption{Regret under the oblivious adversarial regime across different graph structures.}
    \label{fig: oblivious all}
\end{figure*}

\begin{figure*}[t]
    \centering
    \begin{subfigure}[t]{0.47\textwidth}
        \centering
        \includegraphics[width=\linewidth]
        {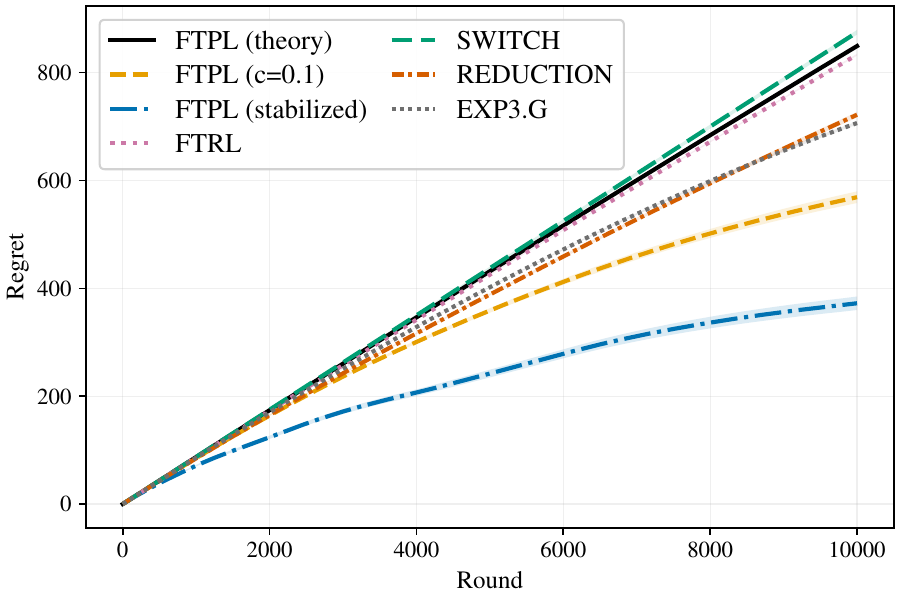}\vspace{-0.25em}
        \caption{\textsc{Loopless Ring}.}
    \end{subfigure}
    \\
    \begin{subfigure}[t]{0.47\textwidth}
        \centering
        \includegraphics[width=\linewidth]
        {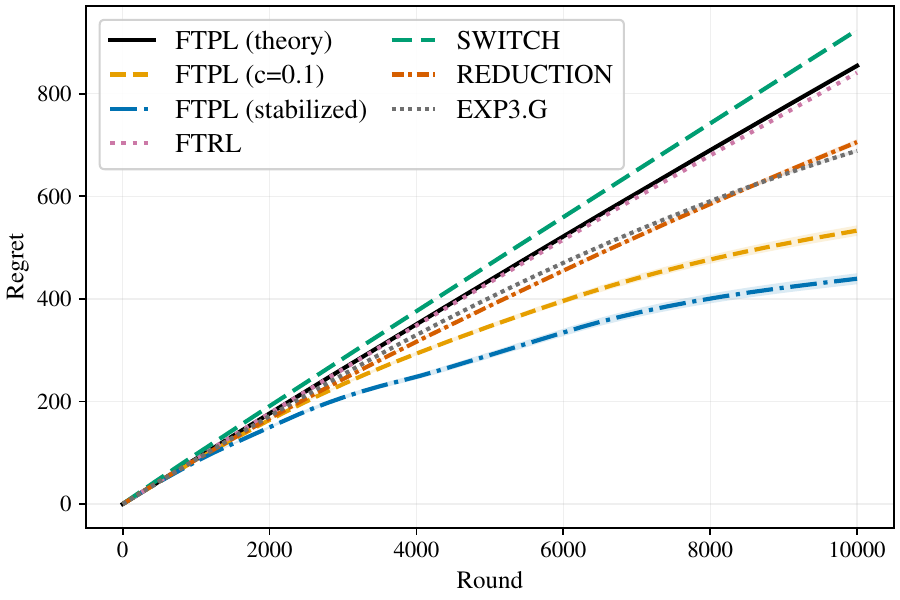}\vspace{-0.25em}
        \caption{\textsc{Partially Looped Ring} with the optimal arm in $V_1$.}
    \end{subfigure}
    \quad
      \begin{subfigure}[t]{0.47\textwidth}
        \centering
        \includegraphics[width=\linewidth]
        {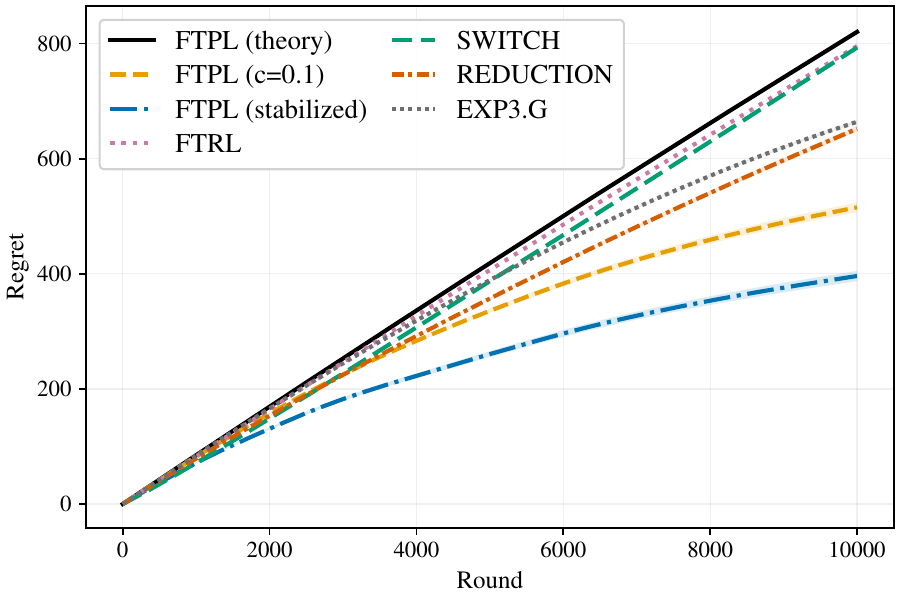}\vspace{-0.25em}
        \caption{\textsc{Partially Looped Ring} with the optimal arm in $V_2$.}
    \end{subfigure}
    \\
        \begin{subfigure}[t]{0.47\textwidth}
        \centering
        \includegraphics[width=\linewidth]
        {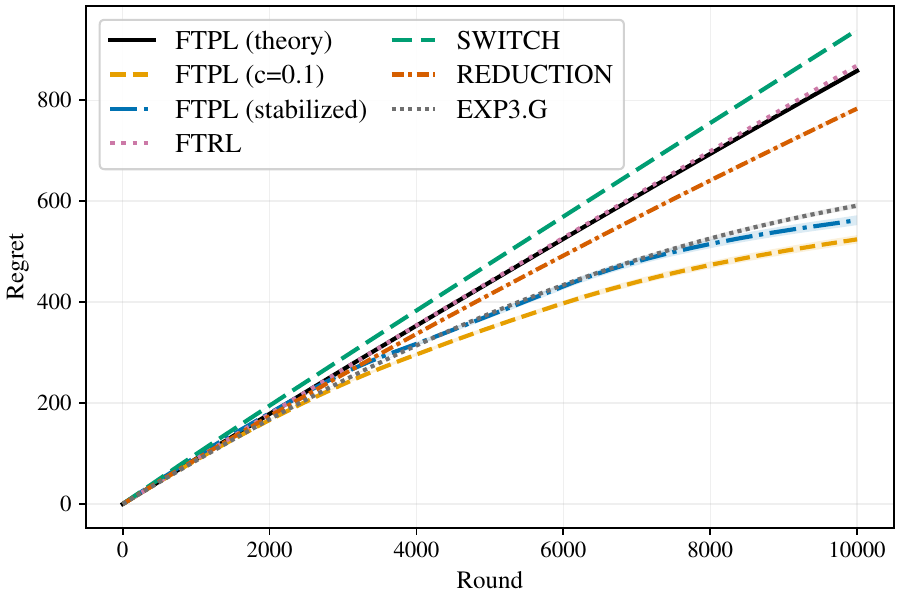}\vspace{-0.25em}
        \caption{\textsc{Bandit Apple} with the optimal arm in $V_1$.}
    \end{subfigure}
    \quad
      \begin{subfigure}[t]{0.47\textwidth}
        \centering
        \includegraphics[width=\linewidth]
        {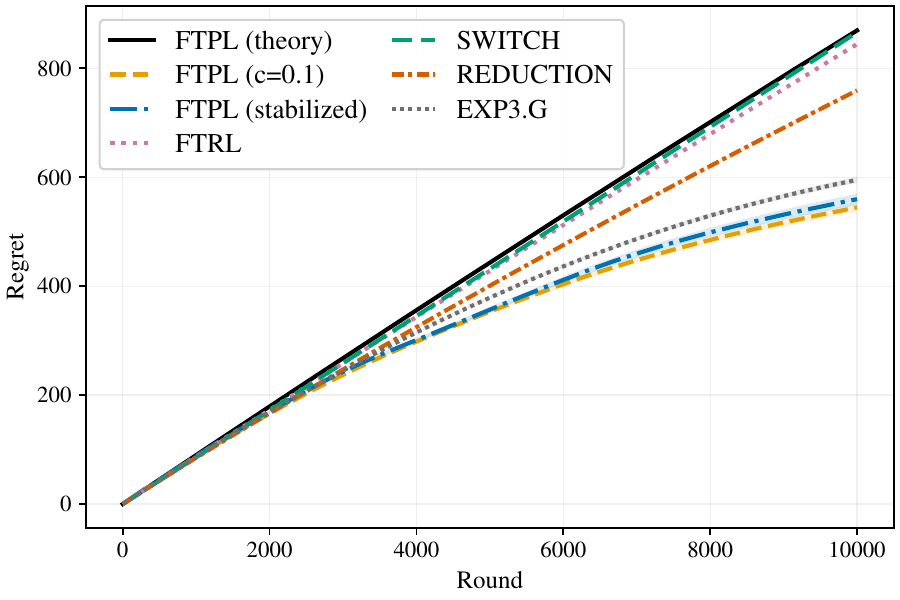}\vspace{-0.25em}
        \caption{\textsc{Bandit Apple} with the optimal arm in $V_2$.}
    \end{subfigure}
        \\
        \begin{subfigure}[t]{0.47\textwidth}
        \centering
        \includegraphics[width=\linewidth]
        {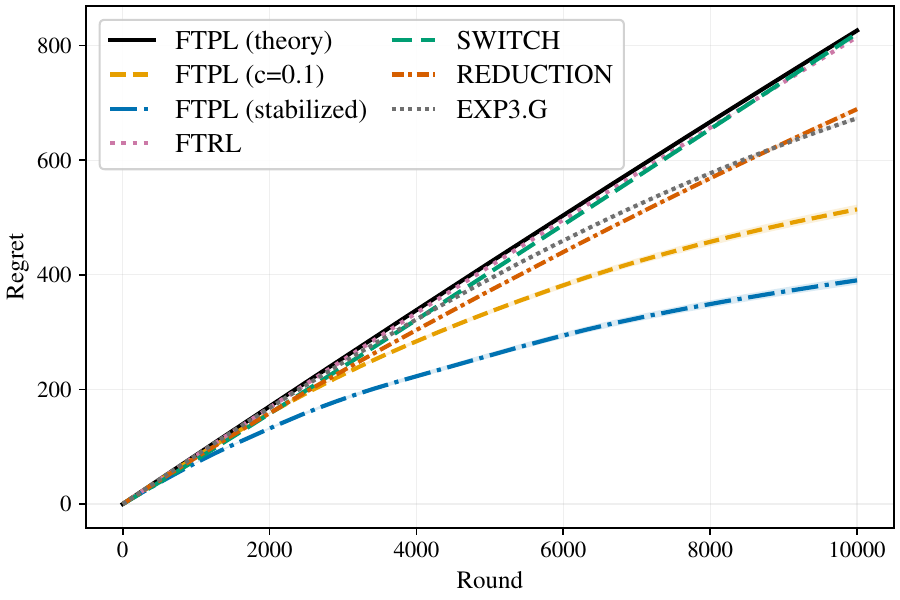}\vspace{-0.25em}
        \caption{\textsc{Domination} with the optimal arm in $V_1$.}
    \end{subfigure}
    \quad
      \begin{subfigure}[t]{0.47\textwidth}
        \centering
        \includegraphics[width=\linewidth]
        {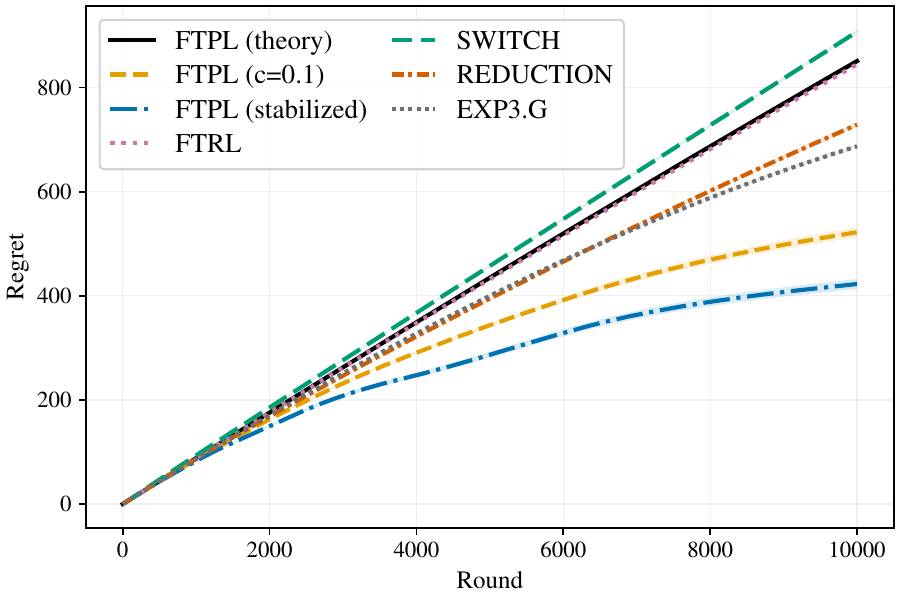}\vspace{-0.25em}
        \caption{\textsc{Domination} with the optimal arm in $V_2$.}
    \end{subfigure}
    \caption{Regret under the stochastically constrained adversarial regime across different graph structures.}
    \label{fig: sca all}
\end{figure*}

\begin{figure*}[t]
    \centering
    \begin{subfigure}[t]{0.47\textwidth}
        \centering
        \includegraphics[width=\linewidth]
        {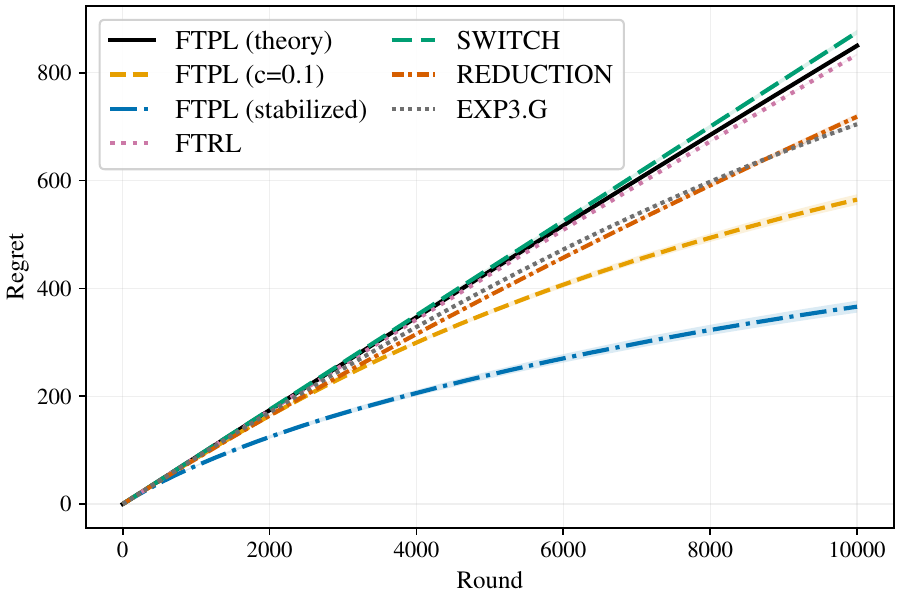}\vspace{-0.25em}
        \caption{\textsc{Loopless Ring}.}
    \end{subfigure}
    \\
    \begin{subfigure}[t]{0.47\textwidth}
        \centering
        \includegraphics[width=\linewidth]
        {figures/ring_loops_best_loopless__stationary_bernoulli__with_exp3g__ci95.pdf}\vspace{-0.25em}
        \caption{\textsc{Partially Looped Ring} with the optimal arm in $V_1$.}
    \end{subfigure}
    \quad
      \begin{subfigure}[t]{0.47\textwidth}
        \centering
        \includegraphics[width=\linewidth]
        {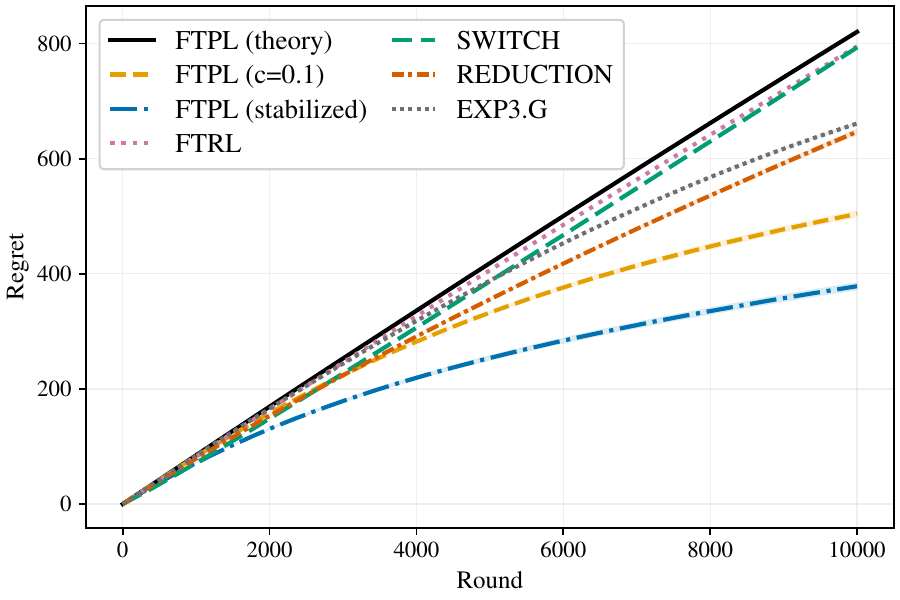}\vspace{-0.25em}
        \caption{\textsc{Partially Looped Ring} with the optimal arm in $V_2$.}
    \end{subfigure}
    \\
        \begin{subfigure}[t]{0.47\textwidth}
        \centering
        \includegraphics[width=\linewidth]
        {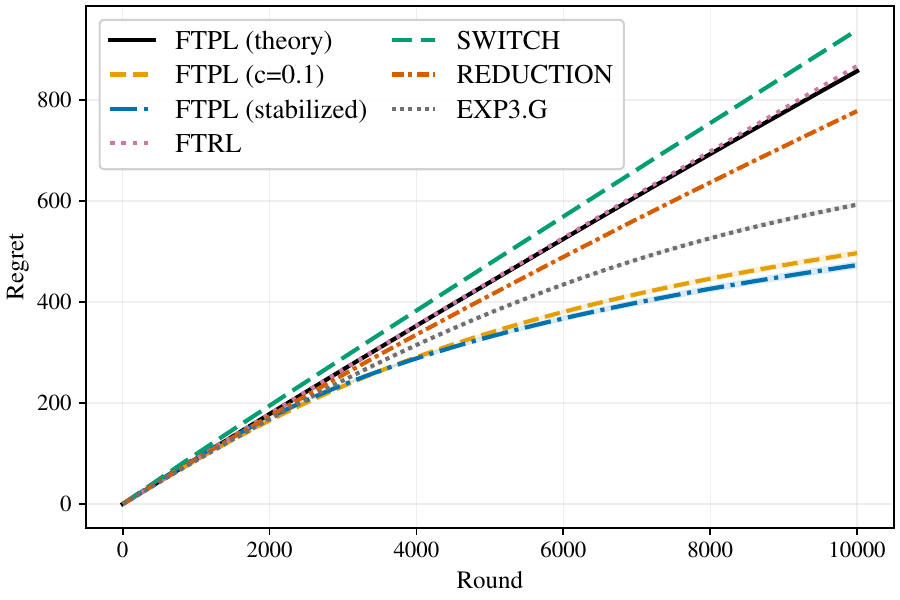}\vspace{-0.25em}
        \caption{\textsc{Bandit Apple} with the optimal arm in $V_1$.}
    \end{subfigure}
    \quad
      \begin{subfigure}[t]{0.47\textwidth}
        \centering
        \includegraphics[width=\linewidth]
        {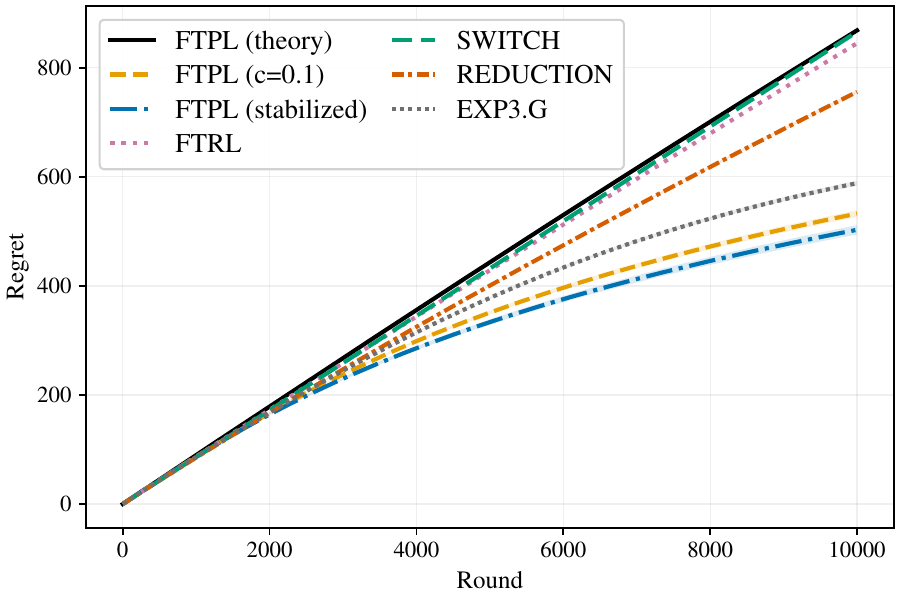}\vspace{-0.25em}
        \caption{\textsc{Bandit Apple} with the optimal arm in $V_2$.}
    \end{subfigure}
        \\
        \begin{subfigure}[t]{0.47\textwidth}
        \centering
        \includegraphics[width=\linewidth]
        {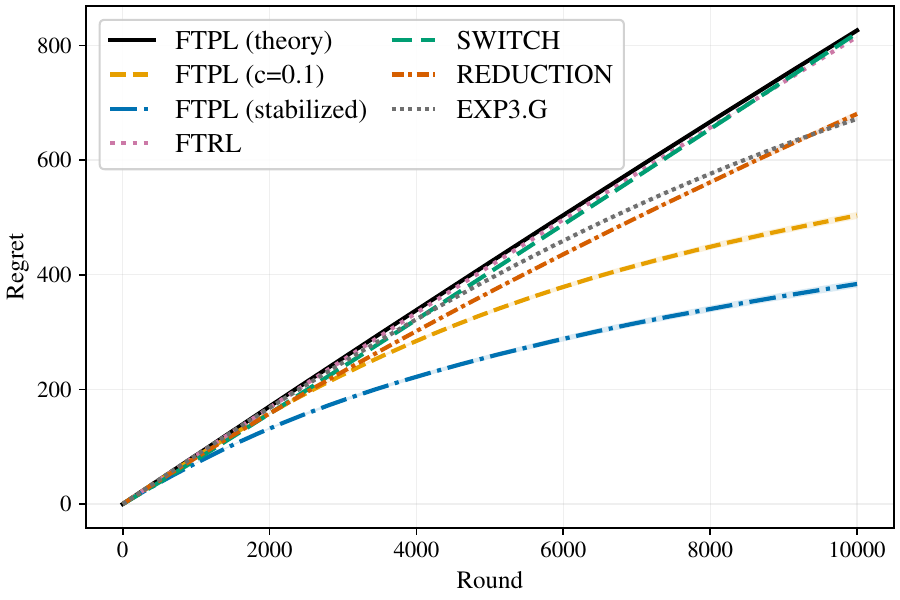}\vspace{-0.25em}
        \caption{\textsc{Domination} with the optimal arm in $V_1$.}
    \end{subfigure}
    \quad
      \begin{subfigure}[t]{0.47\textwidth}
        \centering
        \includegraphics[width=\linewidth]
        {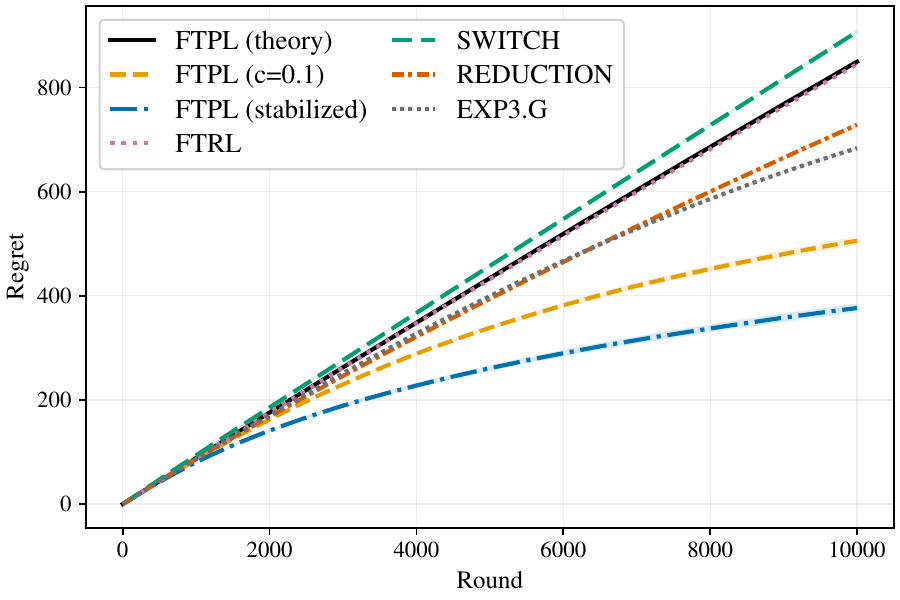}\vspace{-0.25em}
        \caption{\textsc{Domination} with the optimal arm in $V_2$.}
    \end{subfigure}
    \caption{Regret under the stochastic regime across different graph structures.}
    \label{fig: st all}
\end{figure*}
\end{document}

%% file: math_commands.tex
\usepackage{amsmath, bm, amssymb, mathtools}
\usepackage{physics}

\def\eqref#1{equation~\ref{#1}}

\def\1{\bm{1}}

\def\inp#1#2{\left\langle #1, #2 \right\rangle}

\DeclareMathAlphabet{\mathsfit}{\encodingdefault}{\sfdefault}{m}{sl}
\SetMathAlphabet{\mathsfit}{bold}{\encodingdefault}{\sfdefault}{bx}{n}

\def\gA{{\mathcal{A}}}

\def\gC{{\mathcal{C}}}
\def\gD{{\mathcal{D}}}

\def\gO{{\mathcal{O}}}

\def\gT{{\mathcal{T}}}

\def\sN{{\mathbb{N}}}

\def\sR{{\mathbb{R}}}

\newcommand{\E}{\mathbb{E}}

\newcommand{\Reg}{\mathrm{Reg}}

\newcommand{\I}{\mathbbm{1}}

\DeclareMathOperator*{\argmax}{arg\,max}
\DeclareMathOperator*{\argmin}{arg\,min}